%% file: main.tex
\pdfoutput=1
\documentclass[10pt,twocolumn,letterpaper]{article}

\usepackage{cvpr}              % To produce the CAMERA-READY version

\definecolor{cvprblue}{rgb}{0.21,0.49,0.74}
\usepackage[pagebackref,breaklinks,colorlinks,allcolors=cvprblue]{hyperref}

\usepackage{amsmath,amsfonts,bm}
\usepackage{amssymb}
\usepackage{graphicx}
\usepackage{booktabs}
\usepackage{multirow}
\usepackage{float}
\usepackage{cancel}
\usepackage{tikz}
\usepackage{stackengine}
\usepackage{colortbl}

\newcommand{\vv}{\mathbf{v}}

\input{math_commands.tex}

\def\paperID{2216} % *** Enter the Paper ID here
\def\confName{CVPR}
\def\confYear{2025}

\title{Cross-modal Information Flow in Multimodal Large Language Models}

\author{
    \textbf{Zhi Zhang}\textsuperscript{*},\hspace{0.2cm}
    \textbf{Srishti Yadav}\textsuperscript{*†},\hspace{0.2cm}
    \textbf{Fengze Han}\textsuperscript{‡},\hspace{0.2cm}
    \textbf{Ekaterina Shutova}\textsuperscript{*}\\
    \textsuperscript{*}ILLC, University of Amsterdam, Netherlands \\
    \textsuperscript{†}Dept. of Computer Science, University of Copenhagen, Denmark  \\
    \textsuperscript{‡}Dept. of Computer Engineering, Technical University of Munich, Germany \\
    {\tt\small zhangzhizz2626@gmail.com, srya@di.ku.dk, fengze.han@tum.de, e.shutova@uva.nl}
}

\begin{document}
\maketitle
\input{sec/0_abstract}    
\input{sec/1_intro}

\input{sec/2_related_work}

\input{sec/3_method}

\input{sec/4_experiment_setting}
\input{sec/5_where_information_flow}

\input{sec/6_how_information_integration}

\input{sec/7_how_final_decision}
\input{sec/8_conclusion}

\clearpage

{
\small
\bibliographystyle{ieeenat_fullname}
\bibliography{main,reference_import_from_citedrive,references_import_from_zotero}
}

\appendix
\input{sec/X_suppl}

% \clearpage
% \input{rebuttal}


\end{document}

%% file: math_commands.tex
\def\eqref#1{equation~\ref{#1}}
\def\1{\bm{1}}

\def\va{{\bm{a}}}

\def\vf{{\bm{f}}}

\def\vh{{\bm{h}}}

\def\vt{{\bm{t}}}

\def\vv{{\bm{v}}}

\def\mA{{\bm{A}}}

\def\mE{{\bm{E}}}

\def\mH{{\bm{H}}}
\def\mI{{\bm{I}}}

\def\mK{{\bm{K}}}

\def\mM{{\bm{M}}}

\def\mQ{{\bm{Q}}}

\def\mT{{\bm{T}}}

\def\mV{{\bm{V}}}
\def\mW{{\bm{W}}}

\DeclareMathAlphabet{\mathsfit}{\encodingdefault}{\sfdefault}{m}{sl}
\SetMathAlphabet{\mathsfit}{bold}{\encodingdefault}{\sfdefault}{bx}{n}

\def\sI{{\mathbb{I}}}

\def\sO{{\mathbb{O}}}

\def\sQ{{\mathbb{Q}}}

\def\sS{{\mathbb{S}}}
\def\sT{{\mathbb{T}}}

\def\sW{{\mathbb{W}}}

\def\emM{{M}}

%% file: sec/0_abstract.tex
\begin{abstract}
The recent advancements in auto-regressive multimodal large language models (MLLMs) have demonstrated promising progress for vision-language tasks. While there exists a variety of studies investigating the processing of linguistic information within large language models, little is currently known about the inner working mechanism of MLLMs and how linguistic and visual information interact within these models. In this study, we aim to fill this gap by examining the information flow between different modalities---language and vision---in MLLMs, focusing on visual question answering.
Specifically, given an image-question pair as input, we investigate where in the model and how the visual and linguistic information are combined to generate the final prediction. Conducting experiments with a series of models from the LLaVA series, we find that there are two distinct stages in the process of integration of the two modalities. In the lower layers, the model first transfers the more general visual features of the whole image into the representations of (linguistic) question tokens. In the middle layers, it once again transfers visual information about specific objects relevant to the question to the respective token positions of the question. Finally, in the higher layers, the resulting multimodal representation is propagated to the last position of the input sequence for the final prediction.
Overall, our findings provide a new and comprehensive perspective on the spatial and functional aspects of image and language processing in the MLLMs, thereby facilitating future research into multimodal information localization and editing. Our code and collected dataset are released here: \url{https://github.com/FightingFighting/cross-modal-information-flow-in-MLLM.git}.
\end{abstract}

%% file: sec/1_intro.tex
\section{Introduction}
\label{sec: intro}

\begin{figure}[t]
\centering
\includegraphics[width=0.85\linewidth]{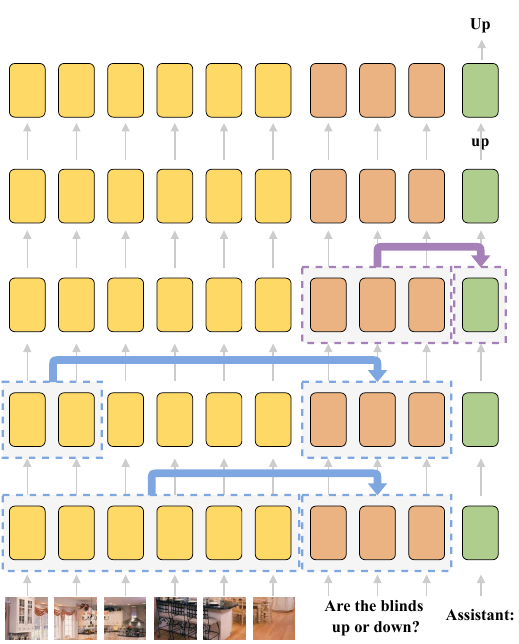}
% \rule{5cm}{6cm} 
\caption{
Illustration of the internal mechanism of MLLMs when solving multimodal tasks. From bottom to top layers, the model first propagates general visual information from the whole image into the linguistic hidden representation; next, selected visual information relevant to answering the question is transferred to the linguistic representation; finally, the integrated multimodal information within the hidden representation of the question flows to \textit{last position} facilitating the final prediction. In addition, the answers are initially generated in lowercase form and then converted to uppercase for the first letter.
}
\label{fig:information flow overview}
\end{figure}

Multimodal large language models (MLLMs) \citep{Liu2023llava1_5,liu2024llavanext,li_blip-2_2023,dai_instructblip_2023,bai_qwen-vl_2023} have demonstrated notable performance across a wide range of vision-language tasks, which is largely attributed to the combination of powerful auto-regressive large language models \citep{touvron_llama_2023, touvron_llama_nodate, zhang_opt_2022, Zheng2023Vicuna} and visual encoders \citep{radford2021learning, fang2023eva,dosovitskiy_image_2020}. Specifically, LLMs generate responses based on both visual and linguistic inputs where visual representations extracted from an image encoder precede the word embeddings in the input sequence. Despite the successful performance and wide applicability of MLLMs, there is still a lack of understanding of their internal working mechanisms at play when solving multimodal tasks. Acquiring deeper insights into these mechanisms could not only enhance the interpretability and transparency \cite{Mechanistic-2024-10-20,nanda2023progress} of these models but also pave the way for developing more efficient and robust models for multimodal interactions.

Some initial studies have begun to explore the internal states corresponding to external behaviors of MLLMs, focusing on specific aspects such as information storage in the model's parameters \cite{basu2024understanding}, reflecting undesirable content generation through logit distributions of the generated tokens \cite{zhao2024first}, the localization and evolution of object-related visual information~\cite{neo2024towards, schwettmann_multimodal_2023, palit_towards_nodate}, the localization of safety mechanism \cite{xu2024cross} and the reduction of redundant visual tokens \cite{zhang_redundancy_2024}. However, the information flow between the two modalities within MLLMs remains poorly-understood, thus prompting our main question: \emph{Where in the model and how is visual and linguistic information integrated within the auto-regressive MLLMs to generate the final prediction in vision-language tasks? }

%\vspace{+0.2cm}
%%\begin{quote}
%(\textbf{\emph{Q} })~~\emph{Where in the model and how are visual and linguistic information integrated within the auto-regressive MLLMs to generate the final prediction in vision-language tasks? 
%}
%%\end{quote}
%\vspace{+0.2cm}

To address this question, we investigate the interaction of different modalities by locating and analyzing the information flow \cite{elhage2021mathematical} between them, across different layers. Our focus is on the task of visual question answering (\textit{VQA}), a popular multimodal task, where the \textit{answer} is generated by MLLMs based on the input \textit{image} and the corresponding \textit{question}. Specifically, we aim to reverse engineer the information flow between the two modalities at inference time, 
by selectively inhibiting specific attention patterns between tokens corresponding to visual and linguistic inputs and by observing the resulting changes in the performance of the \textit{answer} prediction. 

 In modern auto-regressive MLLMs, which employ Transformer decoder-only architecture~\cite{vaswani_attention_2017}, the attention layer is the sole module enabling communication between hidden representations corresponding to different positions of the input. To inhibit cross-modal information flow, we therefore adopt an \textit{attention knockout} approach, proposed by \citet{geva_dissecting_2023}. %,suzuki1986introduction,wang2022interpretability}.
 We use it to block attention edges connecting different types of hidden representations (\textit{e.g. image} \text{and} \textit{question}) at specific transformer layers.
%Through a comprehensive analysis of MLLMs across various \textit{VQA} tasks, we outline our experimental procedure (also shown in \Cref{fig:information flow overview}) and findings below :

We apply this method to a range of MLLMs from the LLaVA series, including \textit{LLaVA-1.5-7b}, \textit{LLaVA-1.5-13b} \citep{Liu2023llava1_5}, \textit{LLaVA-v1.6-Vicuna-7b} \citep{liu2024llavanext} and \textit{Llama3-LLaVA-NEXT-8b} \cite{lmms-lab/llama3-llava-next-8b-2024-11-13} and a number of diverse question types in VQA, as shown in \Cref{tab: question type}. Our experiments focus on the following research questions: (1) How is the (more general) visual information from the whole image fused with the linguistic information in the question? (2) How is the more targeted visual information (i.e. specific image regions directly relevant to answering the question) integrated with linguistic information from the question? and (3) In what ways do the linguistic and visual components of the input contribute to the final \textit{answer} prediction? To answer these questions we conduct a series of experiments, blocking information flow between (1) the input positions corresponding to the whole image to the different parts of the question; (2) the input positions corresponding to image regions containing objects relevant to answering the question, to the question; (3) the input positions corresponding to the image and the question to the final prediction, across different layers of the MLLM. 

Our results reveal that in MLLMs, visual information undergoes a two-stage integration into the language representation within the lower-to-middle layers: first in a comprehensive manner, and subsequently in a more targeted fashion. This integrated multimodal representation is then propagated to the hidden representations in the subsequent layers, ultimately reaching the last position for generating an accurate response. The visualization of this mechanism is shown in \Cref{fig:information flow overview}. To the best of our knowledge, ours is the first paper to elucidate the information flow between the two modalities in auto-regressive MLLMs. It thus contributes to enhancing the transparency of these models and provides novel and valuable insights for their development.

%% file: sec/2_related_work.tex
\section{Related work}
\label{sec: related work}

\paragraph{MLLMs} multimodal large language models have demonstrated remarkable performance across a wide range of vision-language tasks, which is largely attributed to the development of the auto-regressive large language models. The representative MLLMs \citep{Liu2023llava1_5, liu_visual_2023, bai_qwen-vl_2023, liu2024llavanext, li_blip-2_2023, dai_instructblip_2023, li2024minigeminiminingpotentialmultimodality} consist of an image encoder \citep{radford2021learning, fang2023eva,dosovitskiy_image_2020} and a powerful decoder-only large language model \citep{touvron_llama_2023, touvron_llama_nodate, zhang_opt_2022, Zheng2023Vicuna}. The visual and linguistic information are integrated in original LLM. In this paper, we will investigate this inner working mechanism of multimodal information processing into these models.

\vspace{-0.3cm}
\paragraph{Interpretability of multimodal models}
The interpretability of multimodal models has attracted a great deal of attention in the research community. Works in 
\cite{cao_behind_2020,frank_vision-and-language_2021} treat the model as a black box, analyzing input---output relationships to interpret the behavior of models, such as comparing the importance of different modalities \cite{cao_behind_2020} and the different modalities' contribution to visual or textual tasks \cite{frank_vision-and-language_2021}.
The works from \cite{stan2024lvlm,aflalo2022vl,chefer_generic_2021,lyu2022dime} aim to explain predictions by tracing outputs to specific input contributions for a single sample, including through merging the attention scores \cite{stan2024lvlm,aflalo2022vl}, using gradient-based methods \cite{chefer_generic_2021} or model disentanglement \cite{lyu2022dime}.
Additionally, some works \cite{dahlgren-lindstrom-etal-2020-probing,hendricks-nematzadeh-2021-probing,salin_are_2022} adopt a top-down approach, probing learned representations to uncover high-level concepts, such as visual-semantics \cite{dahlgren-lindstrom-etal-2020-probing}, verb understanding \cite{hendricks-nematzadeh-2021-probing}, shape and size \cite{salin_are_2022}. In contrast, our work focuses on the model’s internal processing mechanisms when solving multimodal tasks.

%several works use different probing tasks to evaluate which modality is more important \cite{cao_behind_2020}.  %encoder

%LVLM-Interpret, an interactive tool for interpreting responses from large vision-language models. The tool offers a way to visualize how generated outputs relate to the input image through raw attention, relevancy maps, and causal interpretation. \cite{stan2024lvlm} % mllm

%We find that the hidden states at the specific transformer layers play a crucial role in the successful activation of the safety mechanism \cite{xu2024cross}

%In CLIP, \cite{goh2021multimodal} discovered multimodal neurons in the image encoder that respond to concepts across different modalities

\vspace{-0.3cm}
\paragraph{Mechanistic interpretability of MLLMs} 
Mechanistic interpretability \cite{nanda2023progress, Mechanistic-2024-10-20} is an emerging research area in NLP, aiming to reverse-engineer detailed computations within neural networks. While it has gained attraction in NLP, research in the multimodal domain remains limited. \citet{palit_towards_nodate} introduced a causal tracing tool for image-conditioned text generation on BLIP \cite{li_blip_nodate}, marking one of the few early efforts in this area. Several initial studies have started to explore the internal states of MLLMs by linking external behaviours to specific mechanisms, such as information storage in model parameters \cite{basu2024understanding}, undesirable content generation reflected in the logit distributions of the first generated token \cite{zhao2024first}, localization and evolution of object-related visual information~\cite{neo2024towards, schwettmann_multimodal_2023, palit_towards_nodate}, safety mechanism localization \cite{xu2024cross}, and reducing redundant visual tokens \cite{zhang_redundancy_2024}. However, research offering a comprehensive understanding of the internal mechanisms behind multimodal information integration in MLLMs is still lacking. This paper makes an important first step towards filling this gap.

%% file: sec/3_method.tex
\section{Tracing information flow in MLLMs}
\begin{figure}[t]
\centering
\includegraphics[width=0.8\linewidth]{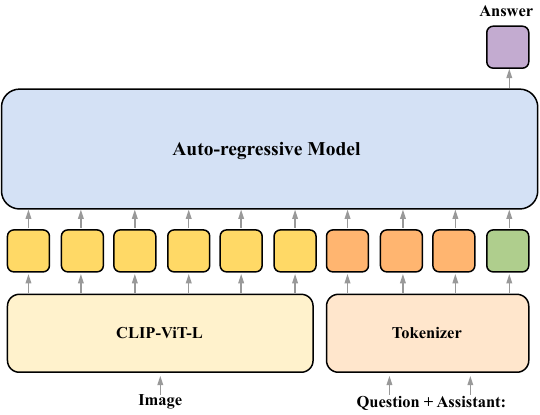}
% \rule{6cm}{4cm} 
\caption{The typical architecture of multimodal large language model. It consists of an image encoder and a decoder-only large language model in which the multimodal information is integrated. We omitted the projection matrix for the visual patch feature as it is nonessential for our analysis. 
}
\label{fig:mllm architecture}
\end{figure}

The focus of this paper is on auto-regressive multimodal large language models, which consist of an image encoder and a decoder-only language model, as shown in \Cref{fig:mllm architecture}. The image encoder transforms images into representations that the language model can take as input, while the language model integrates these visual cues with any provided text, generating responses one word at a time. Often, these components are initialized from a pre-trained image encoder (\textit{e.g.} \textit{CLIP-ViT-L-336px} \citep{radford2021learning} ) and a large language model (\textit{e.g. Llama 2} \citep{touvron_llama_2023}) respectively. Since the interaction between modalities only occurs in the decoder-only transformer, our analysis centers around it and we refer to it as MLLM for brevity unless otherwise specified.
 
\subsection{Background: MLLMs}
\label{sec: background}

\paragraph{Input} 
The input to an MLLM typically comprises image and text features, with the image features being initially extracted from an image encoder and the text being encoded through word embeddings.
Formally, an image $x$ is evenly split into fixed-size patches and encoded by an image encoder to obtain $N_{V}$ visual patch features $\mV=[\vv_i]_{i=1}^{N_{V}}$, $\vv_i \in \mathbb{R}^{d}$. Similarly, the text $t$, consisting of $N_T$ tokens, is embedded into representations through a lookup table of word embeddings, resulting in the text input $\mT=[\vt_i]_{i=1}^{N_T}$, $\vt_i \in \mathbb{R}^{d}$. By concatenation of $\mV$ and $\mT$, the multimodal input sequence $\mI = [\vv_1\dots\vv_{N_{V}}, \vt_1\dots\vt_{N_T}] \in \mathbb{R}^{N\times d}$, where $N=N_{V}+{N_T}$, is fed into MLLM. %We omitted the projection matrix for the visual patch feature as it is nonessential for our analysis. 
\vspace{-0.3cm}
\paragraph{Hidden representation} The input sequence is fed into the MLLM, where the hidden representation at each token position is encoded across $L$ transformer layers. Each layer primarily consists of two modules: a masked multi-head attention (MHAT) followed by a fully connected feed-forward network (FFN) \citep{vaswani_attention_2017}. For conciseness, we have excluded the bias terms and layer normalization, as they are not crucial for our analysis.
Formally, the hidden representation $\vh_i^\ell \in \mathbb{R}^{d}$ in the position $i$ of the input sequence at layer $\ell$  can be expressed as
\begin{equation}\label{eq:hidden}
\vh_i^\ell = \vh_i^{\ell-1} + \va_i^{\ell} + \vf_i^{\ell},
\end{equation}
where $\va_i^{\ell} \in \mathbb{R}^{d}$ and $\vf_i^{\ell} \in \mathbb{R}^{d}$ are the outputs of MHAT and FFN modules at layer  $\ell$, respectively. $\vh_i^0$ represents a vector in the input $\mI$ with position of $i$. All hidden representations at layer $\ell$ corresponding to the whole input $\mI$ can be denoted by $\mH^\ell=[\vh_i^\ell]_{i=1}^{N} \in \mathbb{R}^{N\times d}$.
\vspace{-0.3cm}
\paragraph{MHAT} The masked multi-head attention (MHAT) module in each transformer layer $\ell$ contains four projection matrixes: $\mW_Q^\ell, \mW_K^\ell, \mW_V^\ell, \mW_O^\ell \in \mathbb{R}^{d \times d}$. For the multi-head attention, the input $\mH^{\ell-1}$ is first projected to query,key and value: $\mQ^\ell=\mH^{\ell-1}\mW_Q^\ell$, $\mK^\ell=\mH^{\ell-1}\mW_K^\ell$, $\mV^\ell=\mH^{\ell-1}\mW_V^\ell$. Then the projected query, key and value matrices are evenly split along the columns to $H$ different heads: $\{\mQ^{\ell,j}\}_{j=1}^H, \{\mK^{\ell,j}\}_{j=1}^H, \{\mV^{\ell,j}\}_{j=1}^H \in \mathbb{R}^{N\times \frac{d}{H}}$, respectively. %Following works \cite{elhage2021mathematical, dar2022analyzing,geva_dissecting_2023}, after splitting $\mW_O^\ell$ into $\{\mW_O^{\ell,j}\}_{j=1}^H \in \mathbb{R}^{d\times \frac{d}{H}}$, the output of MHAT $\mA^\ell=[\va_i^\ell]_{i=1}^N \in \mathbb{R}^{N \times d}$ at layer $\ell$ can be represented as the sum of the output from different heads
After splitting $\mW_O^\ell$ into $\{\mW_O^{\ell,j}\}_{j=1}^H \in \mathbb{R}^{d\times \frac{d}{H}}$, we follow works in \cite{elhage2021mathematical, dar2022analyzing,geva_dissecting_2023} to represent the output of MHAT $\mA^\ell=[\va_i^\ell]_{i=1}^N \in \mathbb{R}^{N \times d}$ at layer $\ell$ as the sum of the output from different heads
\begin{align}
    &\mA^{\ell} 
    = \sum_{j=1}^H \mA^{\ell,j} \mV^{\ell,j} \mW_O^{\ell,j}\label{eq:attention}\\ 
    &\mA^{\ell,j} 
    = \text{softmax} \Bigg(\frac{\mQ^{\ell,j}(\mK^{\ell,j})^T}{\sqrt{d/H}} + \mM^{\ell,j}\Bigg)\label{eq:attention_weights}
\end{align}
where $\mM^{\ell,j}$ is a strictly upper triangular mask for $A^{\ell,j}$ for $j$-th head at layer $\ell$. For an auto-regressive transformer model, $\mM^{\ell,j}$ is used to guarantee that every position of the input sequence cannot attend to succeeding positions and attends to all preceding positions. Therefore, for the element $ \emM_{s,t}^{\ell, j}$ with the coordinate $(s,t)$ in $\mM^{\ell,j}$,
\begin{equation}\label{eq: mask}
    \emM_{s,t}^{\ell, j}=
        \begin{cases}
        -\infty{} & \text{if } t>s, \\
        0 & \text{otherwise}.
        \end{cases}
\end{equation}
\vspace{-0.3cm}
\paragraph{FFN} FFN computes the output representation through
\begin{equation}
\label{eq:ffn}
\vf_j^{\ell} = \mW_U^\ell\; \sigma\Big(\mW_B^\ell \big(\va_j^{\ell} + \vh_j^{\ell-1}\big)\Big)
\end{equation}
where $\mW_U^\ell\in \mathbb{R}^{d \times d_{\textit{\text{ff}}}}$ and $\mW_B^\ell\in \mathbb{R}^{d_{\textit{\text{ff}}} \times d}$ are projection matrices with inner-dimensionality $d_{\textit{\text{ff}}}$, and $\sigma$ is a nonlinear activation function. 
\vspace{-0.3cm}
\paragraph{Output} The hidden representation $\vh_N^L$ corresponding to the last position  $N$ of the input sequence at final layer $L$ is projected by an unembedding matrix $\mE \in \mathbb{R}^{|\mathcal{V}|\times d}$ and finally the probability distribution over all words in the vocabulary $\mathcal{V}$ is computed by
\begin{equation}\label{eq:output_distribution}
P_N = \text{softmax}\big(\mE\vh_N^L\big),
\end{equation}
where the word with the highest probability in $P_N$ is the final prediction.

\subsection{Attention knockout}
In this paper, we mainly investigate the interaction between different modalities by locating and analyzing the information flow between them. We adopt a reverse-engineering approach to trace the information flow. Specifically, by intentionally blocking specific connections between different components in the computation process, we trace the information flow within them by observing changes in the probability of final prediction.

In MLLMs, the attention module (MHAT) is the only module, which has the function of communication between different types of hidden representation corresponding to different positions in the input sequence. Therefore, we intentionally block the attention edges between hidden representations at different token positions (termed as \textit{attention knockout}) to trace the information flow between them.
We take inspiration from the work of \cite{geva_dissecting_2023}, where the authors use attention knockout to assess how the factual information is extracted from a single-modality LLM by evaluating the contribution of certain words in a sentence to last-position prediction. We extend this method to multimodal research by not only examining the contribution of each modality to the last-position prediction but also the transfer of information between different modalities. %This is because we discovered that certain modality specific information does not directly transmit critical information to the last position for the prediction (Please see \Cref{sec: indirect information flow to question}).

Intuitively, when blocking the attention edge connecting two hidden representations corresponding to different positions of the input sequence leads to a significant deterioration in model performance, it suggests that there exists functionally important information transfer between these two representations. 
Therefore, we locate the information flow between different hidden representations corresponding to different positions of the input sequence, such as visual inputs, linguistic inputs, and the last position in the input sequence (the position of answer prediction), by blocking the attention edge between them in the MHAT module and observing the resulting decline in performance as compared to the original model with an intact attention pattern.

Formally, in order to prevent information flow from the hidden representations $\vh^\ell_s$ with position $s$ in the source set $\sS$ (\textit{e.g.} all positions of visual tokens in the input sequence) to the hidden representations $\vh^\ell_t$ with position $t$ in the target set $\sT$ (\textit{e.g.} all positions of linguistic tokens in the input sequence) at a specific layer $\ell<L$, we set the corresponding element $M_{s,t}^{\ell,j}$ in $\mM^{\ell,j}$ to $-\infty$ and the updated \cref{eq: mask} is 
\begin{equation}
    \emM_{s,t}^{\ell, j}=
        \begin{cases}
        -\infty{} & \text{if (} t>s\text{) or (}s\text{ in }\sS \text{ and } t \text{ in } \sT \text{ )}, \\
        0 & \text{otherwise}.
        \end{cases}
        \label{eq: updated mask}
\end{equation}
This prevents the token position in the target set from attending to that in the source set when MLLM generates the predicted answer. %Therefore, this allows us to examine the information flow between different parts of the input sequence within MLLM layers, by observing the change in the performance of final-position prediction. To simplify the presentation, we will use the information flow between different positions at a specific layer to represent the information flow between the hidden representations corresponding to those positions.

%% file: sec/4_experiment_setting.tex
\begin{table*}[t]
\centering
 \setlength{\tabcolsep}{0.8mm}{
\begin{tabular}{c|cccclcc}
\toprule
Name & \begin{tabular}[c]{@{}c@{}}Structural\\ type\end{tabular} & \begin{tabular}[c]{@{}c@{}}Semantic\\ Type\end{tabular} & \begin{tabular}[c]{@{}l@{}}Open /\\ Binary\end{tabular}& \begin{tabular}[c]{@{}c@{}}Image\\ Example\end{tabular} & \multicolumn{1}{c}{Question Example}                            & Answer & Num. \\ \hline
ChooseAttr &Choose & Attribute     & Open& \multirow{3}{*}{\raisebox{-0.78\height}{\includegraphics[width=1.8cm,height=1.25cm]{images/dataset/original_image_174.jpg}}}&What was used to make the door, wood or metal?                   & Wood                        & 1000                       \\
ChooseCat &Choose                               & Category& Open    &  & Which piece of furniture is striated, bed or door? & Bed                      & 1000                       \\
ChooseRel &Choose                               & Relation & Open     & &Is the door to the right or to the left of the bed?   & Right                       & 964                       \\
CompareAttr &Compare                              & Attribute    & Open &\multirow{3}{*}{\raisebox{-0.78\height}{\includegraphics[width=1.8cm,height=1.25 cm]{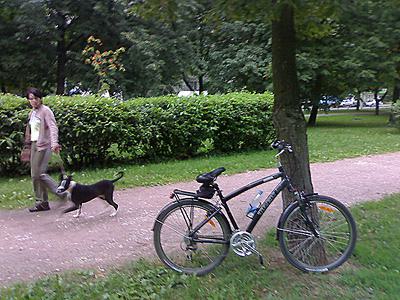}}}&What is common to the bike and the dog?                       & Color                       & 570                        \\
LogicalObj &Logical                              & Object & Binary       & & Are there either women or men that are running?                  & No                          & 991                        \\ 
QueryAttr &Query                                & Attribute   & Open & & In which part of the image is the dog?                      & Left                        & 1000                       \\ \bottomrule
\end{tabular}}
\caption{
Different types of questions in our VQA dataset. 
The questions are categorized based on two dimensions: structure and semantics. The structural types define the question format, including: \textit{Choose} for selecting between alternatives, \textit{Compare} for comparisons between objects, \textit{Logical} for logical inference, and \textit{Query} for open-ended questions. The semantic types focus on the subject matter, covering \textit{Object} existence, and \textit{Attribute}, \textit{Category}, \textit{Relation} of objects. Additionally, questions are labeled as \textit{Open} for open-ended queries or \textit{Binary} for \textit{yes}/\textit{no} answers. The dataset is derived from the GQA dataset \cite{hudson2019gqa}. Due to space limitations, we present two images, noting that 50\% of question samples in our dataset have unique images.
}
\label{tab: question type}
\end{table*}

\section{Experimental setting}
\label{sec: exp}
\paragraph{Setup}
Our paper investigates the inner working mechanism of MLLMs 
, focusing on visual question answering (\textit{VQA}). Typically, the \textit{VQA} setup involves an image and a corresponding question about this image, which the model needs to answer. 
We first investigate where the information from different modalities (image and textual question) is processed in MLLMs, and then how it is integrated within the model. Finally, we explore how the MLLM makes the final decision using this multimodal information.

\vspace{-0.3cm}
\paragraph{Tasks and data}
We collect our data from the validation set of \textit{GQA} dataset \citep{hudson2019gqa}. 
\textit{GQA} is a dataset designed to support visual reasoning and compositional question-answering, offering the semantic and visual richness of real-world images. It is derived from the Visual Genome dataset, which includes detailed scene graph structures \citep{krishna2017visual}.
In \textit{GQA}, the questions are categorized through two dimensions: structure and semantics. The former defines the question format (5 classes) and the latter refers to the semantic information for the main subject of the question (5 classes). The answers to these questions consist of only one word or phrase, which is easy to evaluate.
Based on the two dimensions, the questions in \textit{GQA} are categorized into 15 groups. 
We exclude most groups that consist of simple binary questions (\textit{yes/no}) and demonstrate poor performance on the model investigated in this paper. Finally, we select 6 out of 15 groups (4 structural and 4 semantic classes) in which their performance is higher than 80\% in average performance, as shown in \Cref{tab: question type}. The difficulty of our selected groups ranges from simple multimodal perception tasks to more complex multimodal reasoning. For example, \textit{ChooseAttr} and  \textit{ChooseCat} ask about basic object attributes and categories for one object in the image, \textit{ChooseRel} and \textit{QueryAttr} involve spatial reasoning, and \textit{CompareAttr} and \textit{LogicalObj} require more challenging comparisons and logical reasoning between two objects in the image. For each selected group, we sample an average of 920 image-question pairs that are correctly predicted by most models used in this paper. For each model, we only use correctly predicted samples for analysis (Each model achieves an accuracy greater than 95\% on the dataset we collected). More details about the dataset and the process of collection can be found in \Cref{app: dataset collection}. 
\vspace{-0.3cm}
\paragraph{Format}
Formally, given an \textit{image} $i$ and a \textit{question} $q$ (the \textit{question} may contain \textit{answer options} $os=[o1, o2]$), the model is expected to generate the \textit{answer} $a$ in the \textit{last position} of the input sequence. In addition, the correct one in the options is referred to as the \textit{true option} ($o_t$) while the other ones are denoted as the \textit{false option} ($o_f$). Since the image, question and options might contain multiple input tokens, we use $\sI, \sQ, \sO_t, \sO_f$ to represent the set of input positions corresponding to \textit{image}, \textit{question}, \textit{true option} and \textit{false option}, respectively.
\vspace{-0.3cm}
\paragraph{Evaluation} 
We quantify the information flow between different input parts by evaluating the relative change in the probability of the answer word which is caused by blocking connections between different input parts (\textit{attention knockout}). 
Formally, given an \textit{image}-\textit{question} pair, the MLLM generates the \textit{answer} $a$ with the highest probability $p_1$ from the output distribution $P_N$ defined in \Cref{eq:output_distribution}. After applying \textit{attention knockout} at specific layers, we record the updated probability $p_2$ for the same answer $a$ as in $p_1$. The relative change in probability, $p_c\%$, is calculated as $p_c\%\!=\!((p_2\!-\!p_1)/p_1)\!\times\!100$. In this paper, \textit{attention} \textit{knockout} is applied to each transformer layer (within a defined window) individually and evaluate their respective $p_c$ values.
\paragraph{Models}
We investigate the current state-of-the-art and open-source multimodal large language models from the LLaVA series: \textit{LLaVA-1.5-7b}, \textit{LLaVA-1.5-13b} \citep{Liu2023llava1_5}, \textit{LLaVA-v1.6-Vicuna-7b} \citep{liu2024llavanext} and \textit{Llama3-LLaVA-NEXT-8b} \cite{lmms-lab/llama3-llava-next-8b-2024-11-13}, which achieve state-of-the-art performance across a diverse range of 11 tasks including GQA. These models are trained on similar publicly available data but with different architectures and model sizes, which allows us to explore cross-modal interaction and processing over different architectures and minimize interference of unknown factors from training data. All these models have the same image encoder (\textit{CLIP-ViT-L-336px} \cite{radford2021learning}) but with different LLM: \textit{Vicuna-v1.5-7b} \cite{Zheng2023Vicuna} with 32 layers (transformer blocks) in \textit{LLaVA-1.5-7b} and \textit{LLaVA-v1.6-Vicuna-7b}, \textit{Vicuna-v1.5-13b} \cite{Zheng2023Vicuna} with 40 layers in \textit{LLaVA-1.5-13b}  and \textit{Llama3-8b} \cite{dubey2024llama} with 32 layers in \textit{Llama3-LLaVA-NEXT-8b}, where \textit{Vicuna-v1.5} is the standard and dense transformer architecture \cite{vaswani_attention_2017} and \textit{Llama3} adopts grouped query attention \cite{ainslie2023gqa}. In terms of image processing, \textit{LLaVA-1.5-7b} and \textit{LLaVA-1.5-13b} directly feed the original fixed-length image patch features from the image encoder into the LLM as input tokens. In contrast, \textit{LLaVA-v1.6-Vicuna-7b} and \textit{Llama3-LLaVA-NEXT-8b} employ a dynamic high-resolution technique, which dynamically adjusts image resolution, resulting in variable-length image patch features with higher resolution. 
Due to space limitations, we will primarily present the results for the model \textit{LLaVA-1.5-13b} in the subsequent sections of this paper, while similar findings for other models are presented in \Cref{app: exp on other model}.

%% file: sec/5_where_information_flow.tex
\section{Contribution of different modalities to the final prediction}
\label{sec:where information flow}
\begin{figure}[t]
\centering
\includegraphics[width=1\linewidth]{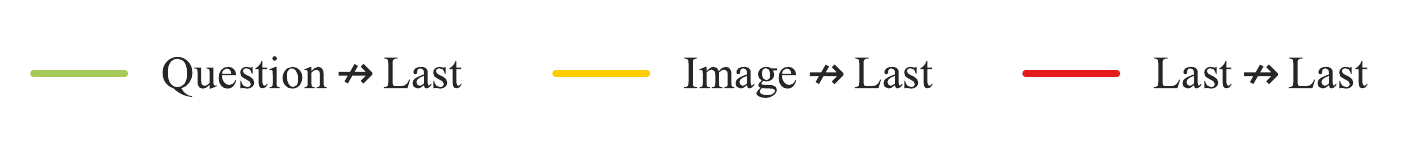}
\centering
\subfloat[ChooseAttr]{
    \includegraphics[width=0.32\linewidth]{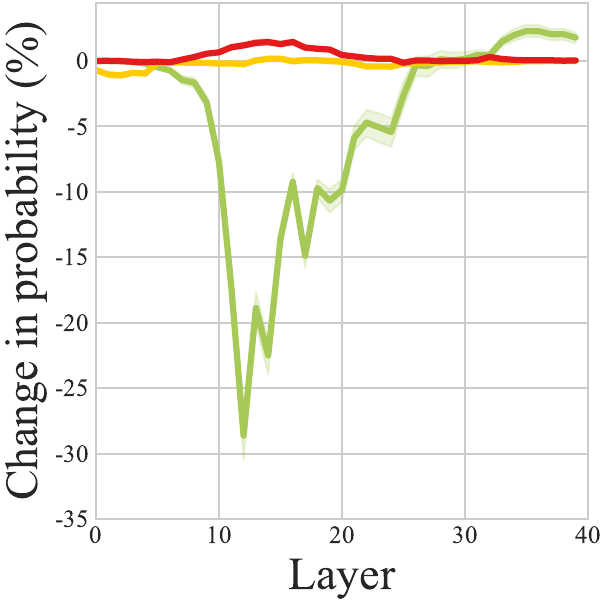}
}
\subfloat[ChooseCat]{
    \includegraphics[width=0.32\linewidth]{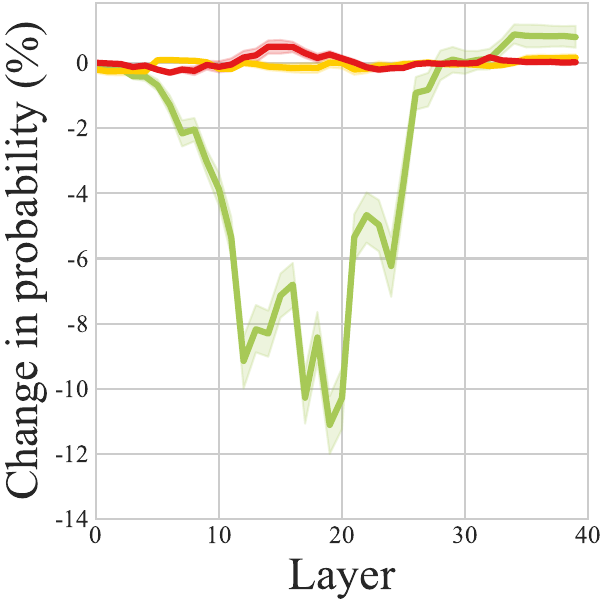}
}
\subfloat[ChooseRel]{
    \includegraphics[width=0.32\linewidth]{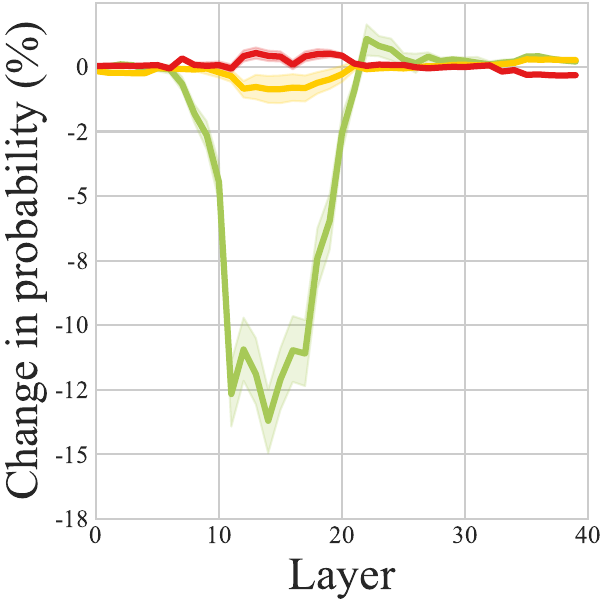}
}\\
\subfloat[CompareAttr]{
    \includegraphics[width=0.32\linewidth]{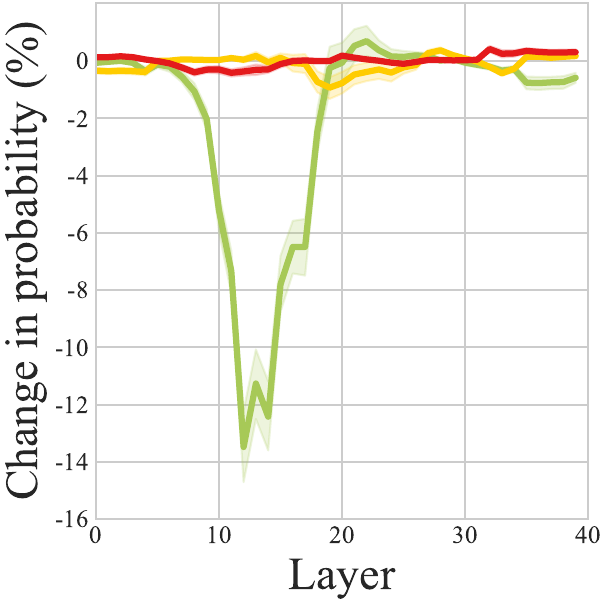}
}
\subfloat[LogicalObj]{
    \includegraphics[width=0.32\linewidth]{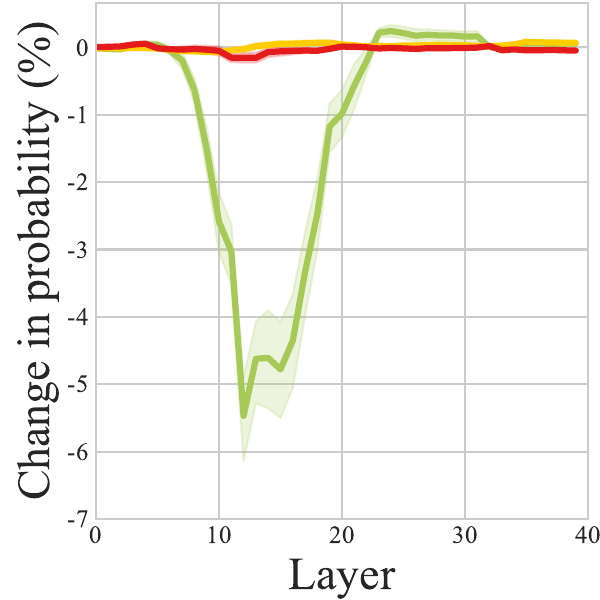}
}
\subfloat[QueryAttr]{
    \includegraphics[width=0.32\linewidth]{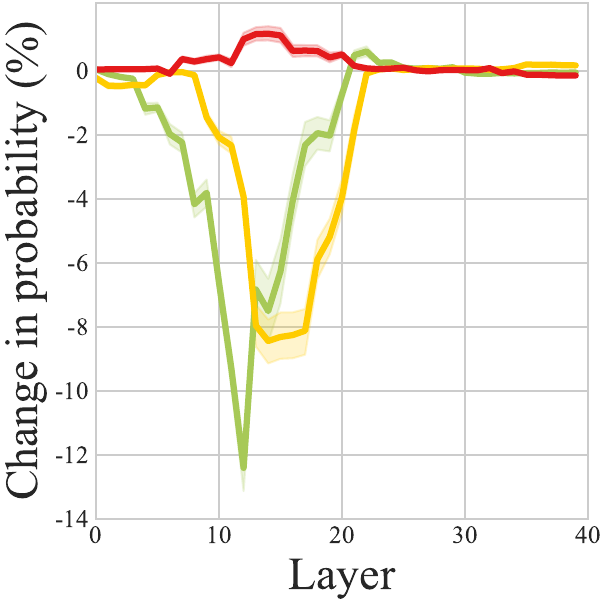}
}
\caption{The relative changes in prediction probability on \textit{LLaVA-1.5-13b} with six \textit{VQA} tasks. The \textit{Question}$\nrightarrow$ \textit{Last}, \textit{Image}$\nrightarrow$ \textit{Last} and  \textit{Last}$\nrightarrow$ \textit{Last} represent preventing \textit{last position} from attending to \textit{Question}, \textit{Image} and itself respectively.}
\vspace{-10pt}
\label{fig:tolast_13b}
\end{figure}

\begin{figure}[t]
\centering
\includegraphics[width=0.33\linewidth]{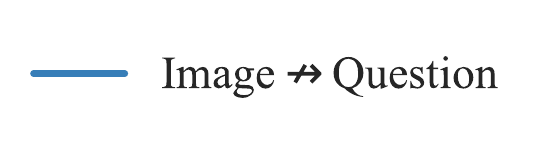}
\centering
\subfloat[ChooseAttr]{
    \includegraphics[width=0.32\linewidth]{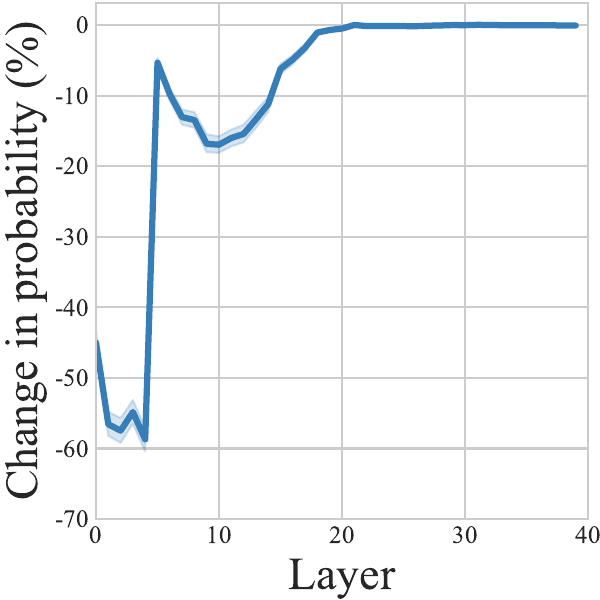}
}
\subfloat[ChooseCat]{
    \includegraphics[width=0.32\linewidth]{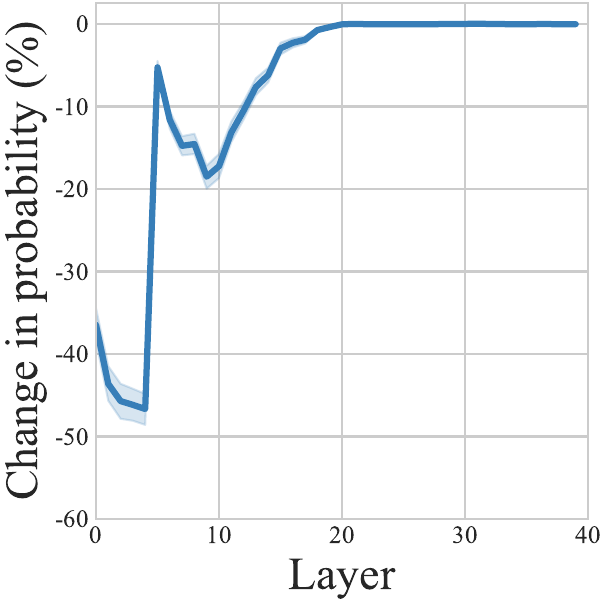}
}
\subfloat[ChooseRel]{
    \includegraphics[width=0.32\linewidth]{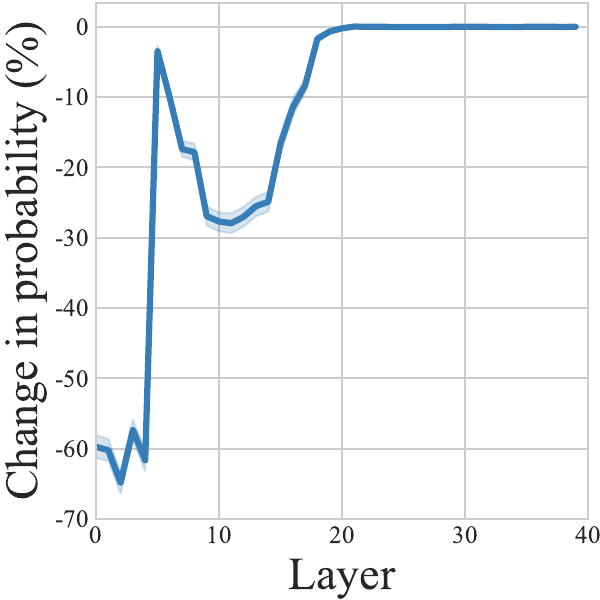}
}\\
\subfloat[CompareAttr]{
    \includegraphics[width=0.32\linewidth]{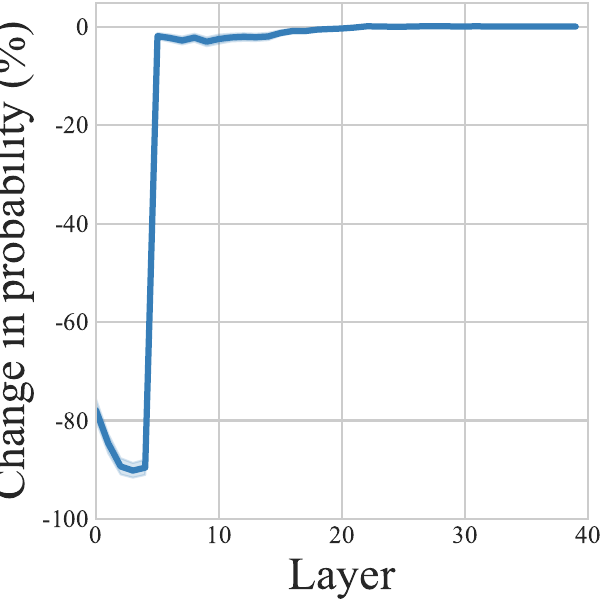}
}
\subfloat[LogicalObj]{
    \includegraphics[width=0.32\linewidth]{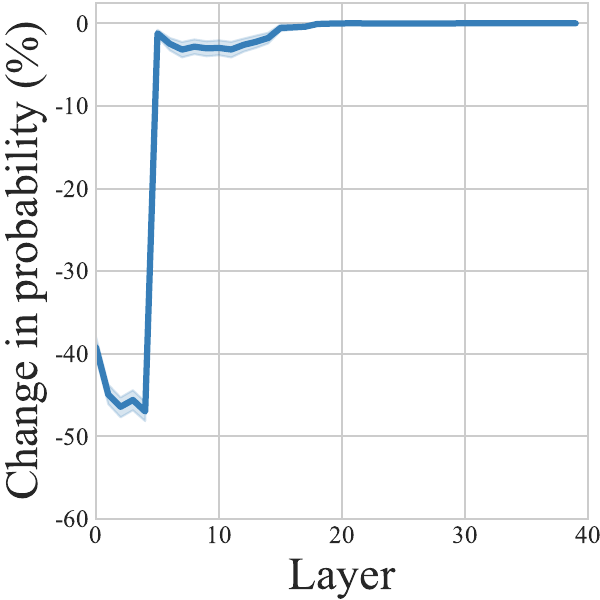}
}
\subfloat[QueryAttr]{
    \includegraphics[width=0.32\linewidth]{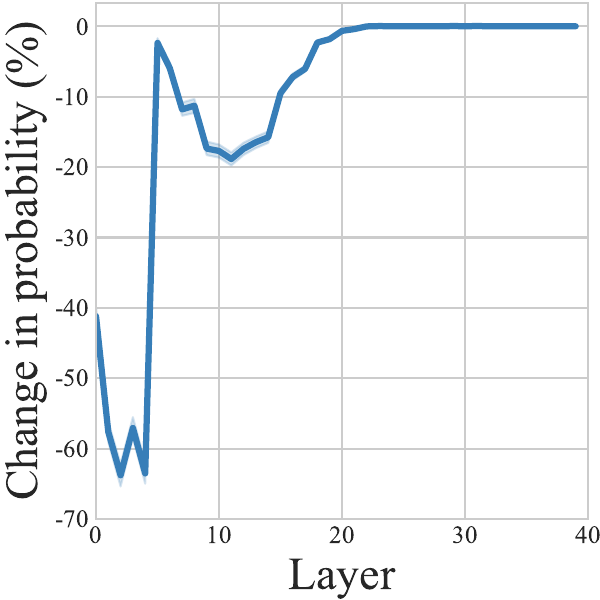}
}
\caption{
The relative changes in prediction probability when blocking attention edges from the \textit{question} positions to the \textit{image} positions on \textit{LLaVA-1.5-13b} with six \textit{VQA} tasks.
}
\vspace{-13pt}
\label{fig:imageToquestion_13b}
\end{figure}

For a successful answer prediction for the task of \textit{VQA}, the MLLM will process the input image-question pair $[i, q]$ and generate the final answer from the output layer of the model corresponding to the \textit{last position}.  
We first investigate whether the different modalities directly contribute to the final prediction. 
\vspace{-0.3cm}
\paragraph{Experiment 1} For each layer $\ell$ in the MLLM, we block 
the target set (the \textit{last position}) from attending to each source set ($\sI$ or $\sQ$) respectively at the layers within a window of $k=9$ layers around the $\ell$-th layer\footnote{We experimented with different values of $k$, as described in \Cref{app: different win}, and observed similar trends as in the analysis we present in this section.}, and measure the change in the probability of the correct answer word. The \textit{last position} means $N$-th position in the input sequence and it is also the first generated sub-word for the predicted answer. Typically, the answers contain a single word or phrase, which might sometimes be tokenized into several sub-word tokens. Therefore, we also conduct the same experiment and observe the probability change at the final generated sub-word of the predicted answer. Both the first and final generated sub-words yield similar results. Thus, we present all the results of the first generated sub-words in the main body of the paper, with details on the final sub-words provided in \Cref{app: last generated sub-word}.
\vspace{-0.3cm}
\paragraph{Observation 1: the contribution to the prediction at the \textit{\textbf{last position}} is derived from other input components, rather than the input itself at this position.} First of all, as an auto-regressive model, it is assumed that the input generated from preceding steps at the final position already encompasses the crucial information required for predicting the correct answer. 
However, as shown in \Cref{fig:tolast_13b}, when we block the attention edge from the \textit{last position} to itself (\textit{Last} $\nrightarrow$ \textit{Last}), there is negligible change observed in the probability of final prediction. 
This implies that the input at the last position does not encompass crucial information for the final prediction of the model. The prediction is, therefore, mainly influenced by other parts of the input sequence. 
\vspace{-0.8cm}
\paragraph{Observation 2: The information from the \textit{question} positions plays a direct and predominant role in influencing the final prediction.} As shown in \Cref{fig:tolast_13b}, blocking attention from the \textit{last position} to the hidden representations in $\sQ$ (\textit{Question} $\nrightarrow$ \textit{Last}) results in significant reduction in prediction probabilities across all six tasks. 
For example, in the \textit{ChooseAttr} task, this decreases the prediction probability by up to $\sim30\%$. This highlights the critical flow of information from $\sQ$ to the \textit{last position}, directly affecting the final prediction. It is worth noting that this information flow pattern is observed primarily in the middle layers, where performance reductions consistently occur across all six tasks.
In contrast, information from the \textit{image} positions ($\sI$) does not directly and significantly impact the final prediction in most tasks, except for \textit{QueryAttr}, where a slight information flow from $\sI$ to the \textit{last position} is observed. However, this direct influence is negligible compared to its indirect effects, discussed below. 
The additional experiment about the information flow between different parts of \textit{question}, such as \textit{options} $os$ and object words, and \textit{last position} can be found in \Cref{app: fine-grain analysis}. 
\vspace{-0.4cm}
\paragraph{Experiment 2}  As the MLLM is auto-regressive and the input format is \textit{image} followed by the \textit{question} in our setting, the information from the \textit{image} ($\sI$) can propagate to the positions of the \textit{question} ($\sQ$), but not the other way around.
To establish whether this indeed occurs, for each layer $\ell$, we block $\sQ$ from attending to $\sI$ with the same window size ($k=9$) around the $\ell$-th layer and observe the change in the probability of the answer word at the \textit{last position} as above. 
\vspace{-0.8cm}
\paragraph{Observation: Information flow from the \textit{image} positions to \textit{question} positions occurs twice} As shown in \Cref{fig:imageToquestion_13b}, blocking the \textit{question} positions from attending to the \textit{image} positions leads to a reduction in prediction probability. This is visible in lower layers, in two different parts of the model. We first observe a sharp drop in layers $\sim0-4$ and then a second smaller drop around $10$th layer. This indicates a two-stage integration process of visual information into the representations of the \textit{question}. %that critical information flows from the position of \textit{image} $\sI$ to that of \textit{question} $\sQ$ in these layers, affecting the final prediction.
In the first drop, attention knockout reduces the prediction probability by an average of $\sim60\%$ across all six tasks. In the second drop, tasks such as \textit{ChooseAttr}, \textit{ChooseCat}, \textit{ChooseRel}, and \textit{QueryAttr} show another average reduction of $\sim21\%$ while \textit{CompareAttr} and \textit{LogicalObj} exhibit smaller decreases. Despite the variability in the magnitude of the reduction, the layers responsible for information flow remain consistent across all tasks, which is also observed during the first drop. 
The additional experiment about the information flow between \textit{image} and different parts of \textit{question}, such as \textit{option} $os$ and object words, can be found in \Cref{app: fine-grain analysis}.

\vspace{-0.5cm}
\paragraph{Overall information flow}
Given the input sequence: \textit{image} and \textit{question} with the corresponding sets of positions $\sI$ and $\sQ$ respectively, the MLLM first propagates information twice from the \textit{image} positions to the \textit{question} positions in the lower-to-middle layers of the MLLM. Subsequently, in the middle layers, the information flows from the \textit{question} positions to the \textit{last position} for the final prediction. 
Overall, this reveals the existence of distinct and disjoint stages in the computation process of different layers in MLLM, where critical information transfer points from different positions corresponding to different modalities are observed to influence the final predictions of the model. These findings are also observed in the other three MLLMs (\cref{app: exp on other model}).

%% file: sec/6_how_information_integration.tex
\section{How is the linguistic and visual information integrated?}
\label{sec:how integrate}
\begin{figure}[t]
\centering
\includegraphics[width=1.0\linewidth]{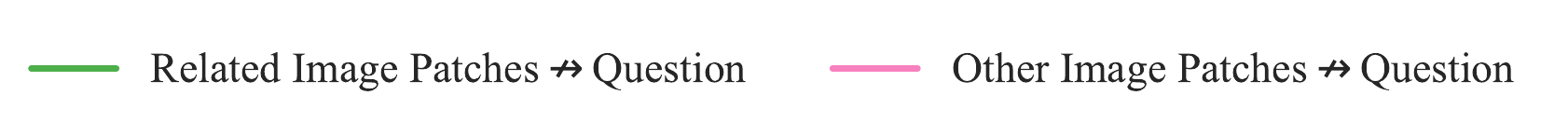}
\\
\centering
\subfloat[ChooseAttr]{
    \includegraphics[width=0.32\linewidth]{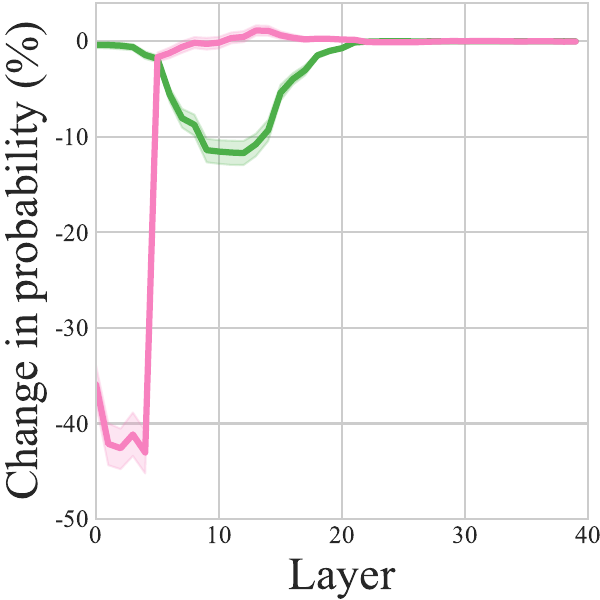}
}
\subfloat[ChooseCat]{
    \includegraphics[width=0.32\linewidth]{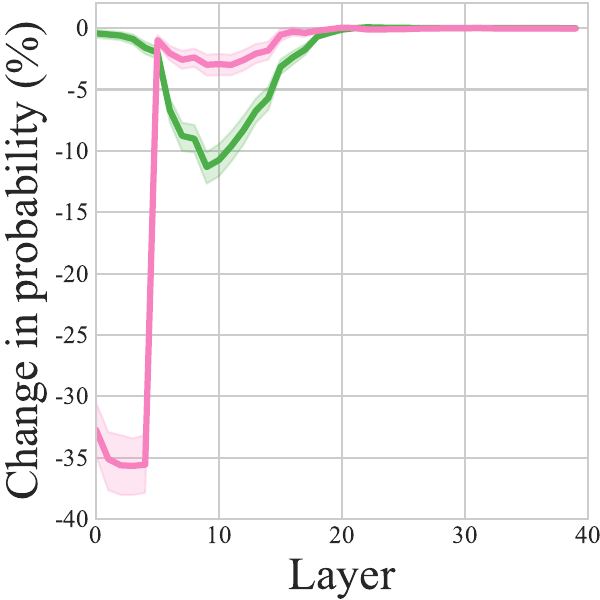}
}
\subfloat[ChooseRel]{
    \includegraphics[width=0.32\linewidth]{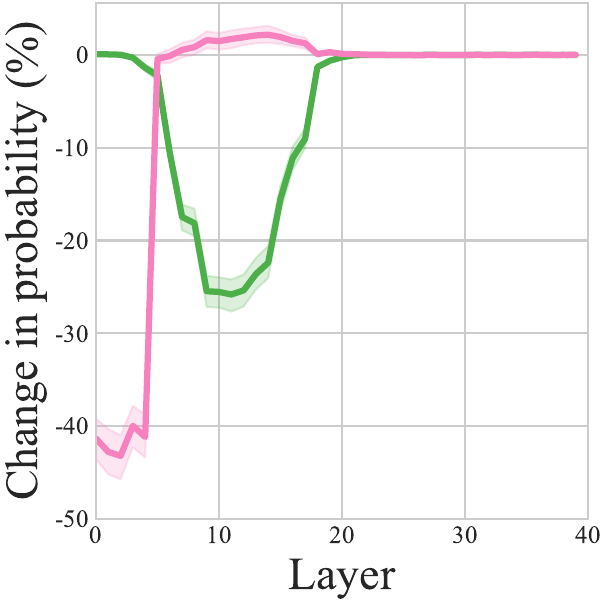}
}\\
\subfloat[CompareAttr]{
    \includegraphics[width=0.32\linewidth]{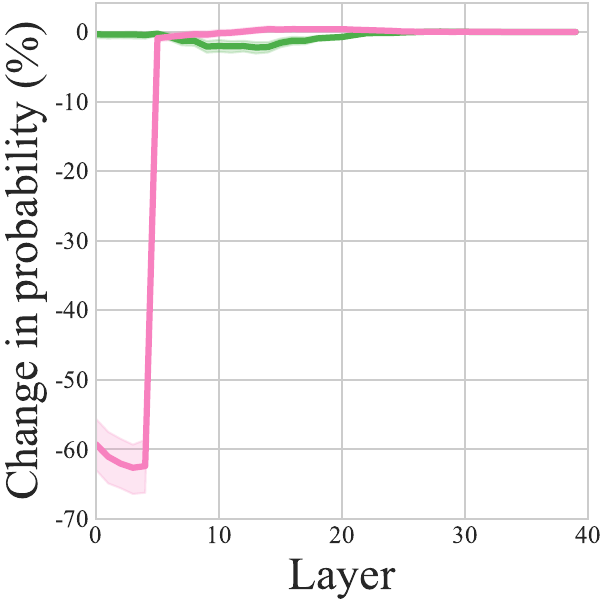}
}
\subfloat[LogicalObj]{
    \includegraphics[width=0.32\linewidth]{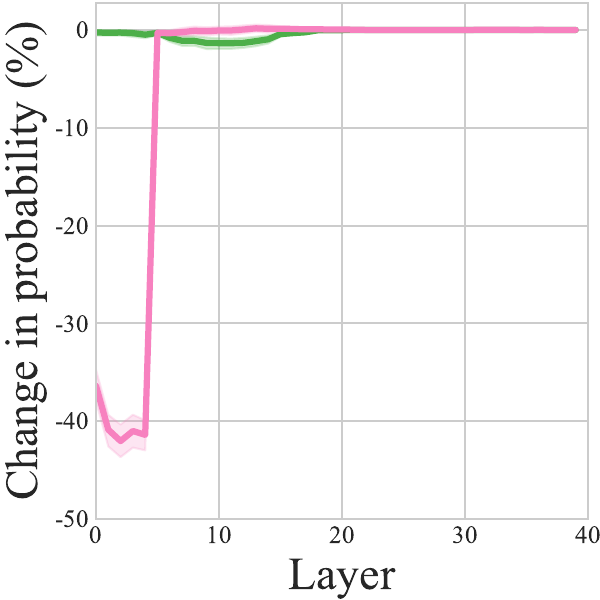}
}
\subfloat[QueryAttr]{
    \includegraphics[width=0.32\linewidth]{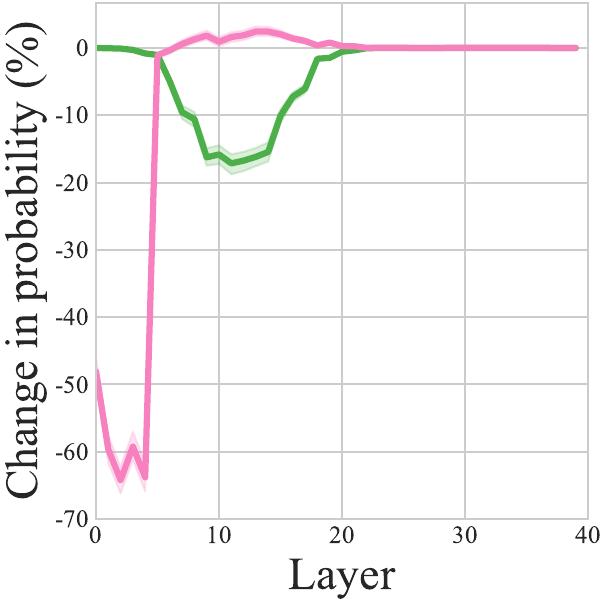}
}
\caption{The relative changes in prediction probability on \textit{LLaVA-1.5-13b} with six \textit{VQA} tasks.
\textit{Related Image Patches}$\nrightarrow$\textit{question} and \textit{Other Image Patches}$\nrightarrow$\textit{question} represent blocking the position of \textit{question} from attending to that of different image patches, region of interest and remainder, respectively.}
\vspace{-12pt}
\label{fig:imagepatchesToquestion_13b}
\end{figure}
The results of the above analysis suggest a two-stage integration process of the two modalities within an MLLM. In this section, we further investigate how the information about specific visual and linguistic concepts is integrated across these two stages.% and how the multimodal semantic representation is constructed.
%\subsection{Multimodal information integration}
\vspace{-0.45cm}
\paragraph{Experiment} 
To investigate how the model uses the image to answer the question, we conducted attention knockout experiments at the level of individual objects and individual words. 
The dataset used in the paper consists of questions targeting specific objects and each object is annotated with the bounding box for a certain image region. 
Based on whether an image patch includes the corresponding bounding boxes (objects), we divide the input image patch features $\mV$ into two groups: $\mV_{\text{obj}}$ corresponding to the patches containing the objects mentioned in the question, and $\mV_{\text{oth}}$ containing the remaining patches. Then, for each layer $\ell$, we use the same \textit{attention knockout} method to block the target set $\sQ$ from attending each source set, $\sI_{\text{obj}}$ and $\sI_{\text{oth}}$, corresponding to the position of $\mV_{\text{obj}}$ and $\mV_{\text{oth}}$ in the input sequence respectively, at the layers with a window of $k=9$ layers around the $\ell$-th layer, and observe the change in the probability of the correct answer word.
\vspace{-0.5cm}
\paragraph{Observation: Shifting focus from comprehensive representation to specific regions of interest}
As illustrated in \Cref{fig:imagepatchesToquestion_13b}, blocking the attention edges between the position of $\mV_{\text{obj}}$ and the \textit{question} (\textit{related image patches}$\nrightarrow$\textit{question}) and between the position of $\mV_{\text{oth}}$ and the \textit{question} (\textit{other image patches}$\nrightarrow$\textit{question}) appear to account for the two performance drops observed in \Cref{fig:imageToquestion_13b}, individually. Specifically, \textit{other image patches}$\nrightarrow$\textit{question} clearly results in a significant and predominant reduction in prediction probability during the first stage of cross-modal integration, while \textit{related image patches}$\nrightarrow$\textit{question} plays a dominant role at the second stage. It is noteworthy that both types of cross-modal information transfer occur in similar layers within the MLLM across all six tasks. Even for the \textit{CompareAttr} and \textit{LogicalObj} tasks, although slight changes in probability are observed during the second stage, the layers in which this happens remain consistent with those of the other tasks.
This suggests in the lower layers, the model integrates the information from the whole image into the \textit{question} positions building a more generic representation. And it is only in the later layers, that the model starts to pay attention to the specific regions in the image relevant to the \textit{question},
fusing the more fine-grained linguistic and visual representations. 
The other MLLMs also present similar results as shown in \Cref{app: exp on other model}. The additional more fine-grained analysis on intervention of the attention edge between object words in \textit{question} and image region can be found in \Cref{app: fine-grain analysis}. Moreover, we find compared with \textit{LLaVA-1.5-13b}, the model \textit{LLaVA-1.5-7b} with smaller size has less information flow from the position of $\mV_{\text{oth}}$ to that of \textit{question} in the first stage, as shown in \Cref{app: exp on other model}.

%% file: sec/7_how_final_decision.tex
\section{How is the final answer generated?}
\label{sec: How final decision}

%After understanding the inner integration mechanism between visual and linguistic information in MLLMs, we further explore the process by which answers are generated within these models.

\begin{figure}[t]
\centering
\includegraphics[width=0.8\linewidth]{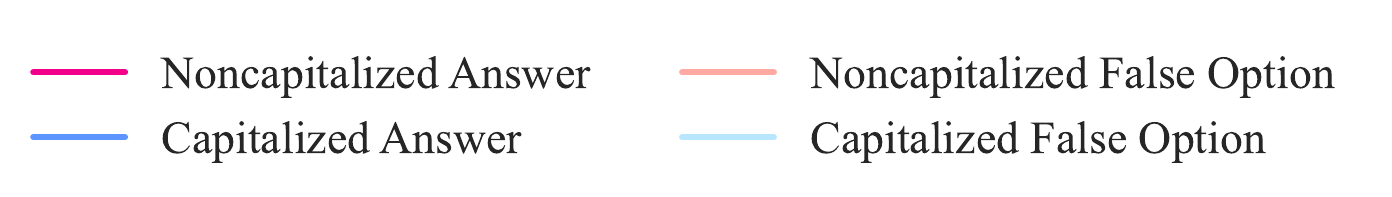}

\centering
\subfloat[ChooseAttr]{
    \includegraphics[width=0.32\linewidth]{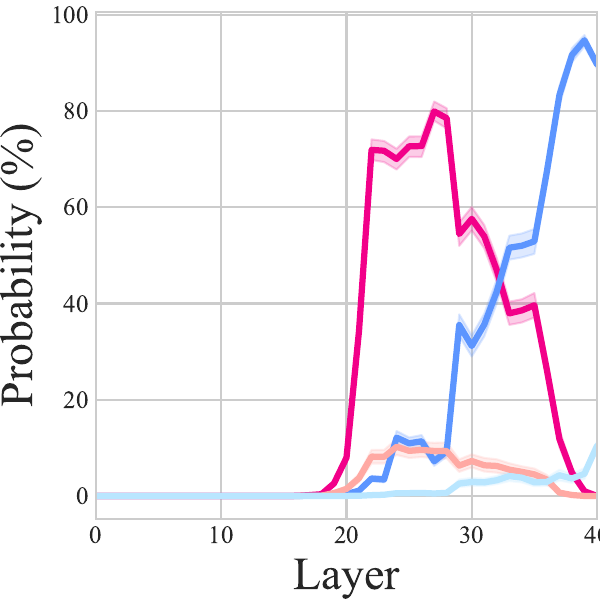}
}
\subfloat[ChooseCat]{
    \includegraphics[width=0.32\linewidth]{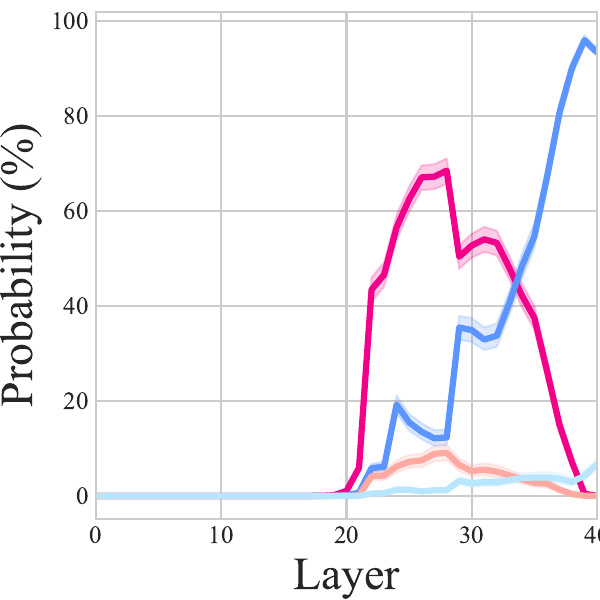}
}
\subfloat[ChooseRel]{
    \includegraphics[width=0.32\linewidth]{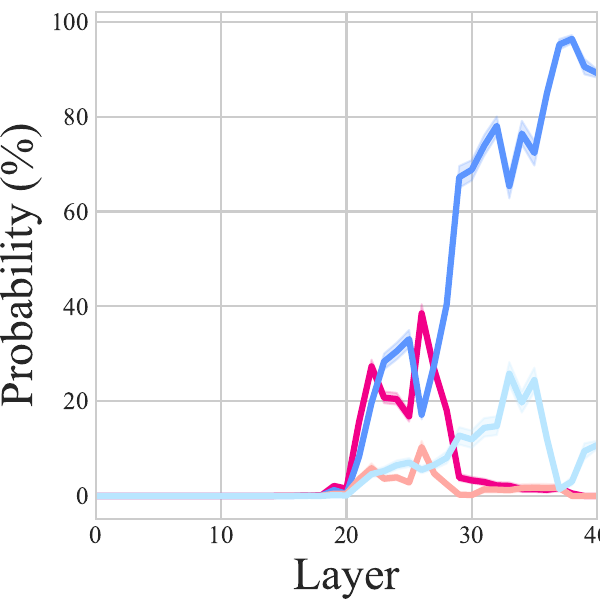}
}\\
\subfloat[CompareAttr]{
    \includegraphics[width=0.32\linewidth]{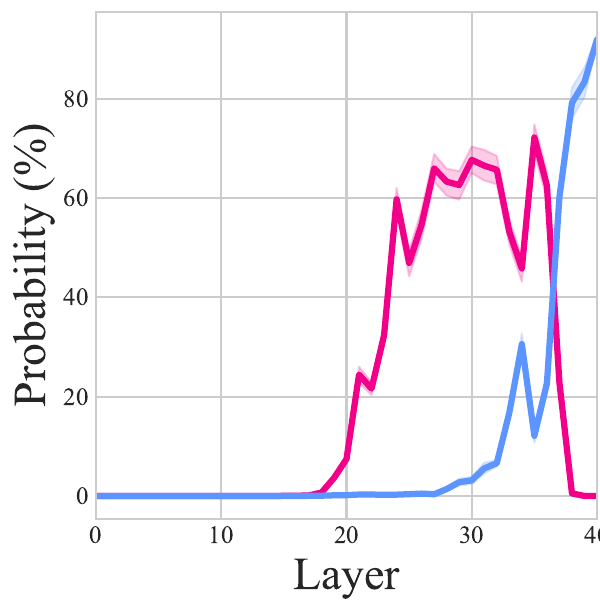}
}
\subfloat[LogicalObj]{
    \includegraphics[width=0.32\linewidth]{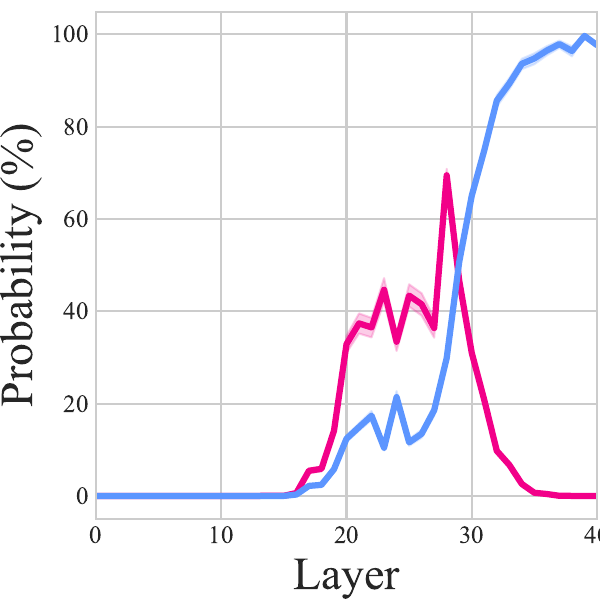}
}
\subfloat[QueryAttr]{
    \includegraphics[width=0.32\linewidth]{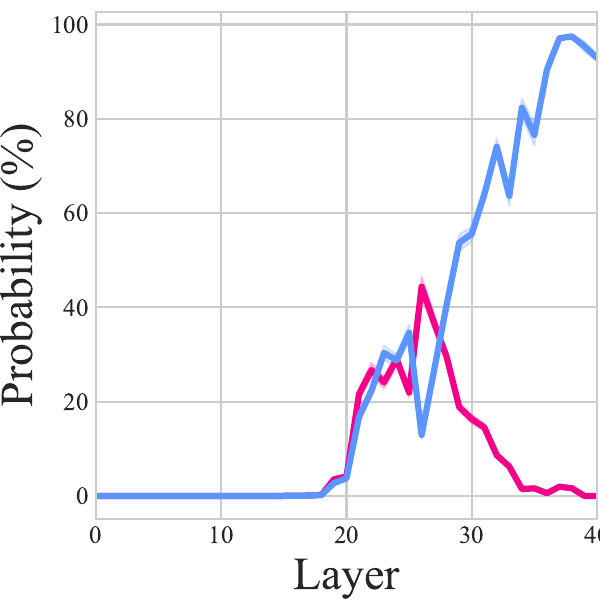}
}
\caption{The probability of the answer word at the \textit{last position} across all layers in \textit{LLaVA-1.5-13b} with six VQA tasks. \textit{Capitalized Answer} and \textit{Noncapitalized Answer} represent the answer word with or without the uppercase of the initial letter, respectively. As the tasks of \textit{ChooseAttr}, \textit{ChooseCat} and \textit{ChooseRel} contain \textit{false option}, we also provide the probability of it.}
\vspace{-12pt}
\label{fig:last_position_prob_13b}
\end{figure}

\paragraph{Experiment}
\vspace{-5pt}
To track the process of answer generation in the MLLM, motivated by the approach of logit lens \cite{lesswrongInterpretingGPT},  we monitor the probability of the correct answer from the hidden representations at the \textit{last position} of the input sequence across all layers. Formally, for each layer $\ell$ at the last position $N$, we use the unembedding matrix $\mE$ (as defined in \Cref{eq:output_distribution}) to compute the probability distribution over the entire vocabulary $\mathcal{V}$: 
\begin{equation} 
P_N^\ell = \text{softmax}\big(\mE\vh_N^\ell\big), 
\end{equation} 
where the probability of the target answer word $w_a$ is given by the corresponding entry in $P_N^\ell$, denoted as $P_N^\ell(w_a)$. As the tokenizer in most MLLMs distinguishes the case of the word, especially the initial letter of the word, we monitor the probability of the answer word with both those starting with uppercase (\textit{Capitalized Answer}) and lowercase letters (\textit{Noncapitalized Answer}).
\vspace{-10pt}
\paragraph{Observation 1: The model is able to predict the correct answer starting at the layer immediately following multimodal integration} 
As illustrated in \Cref{fig:last_position_prob_13b}, the probability of the answer word with a lowercase initial letter (\textit{Noncapitalized Answer}) rises sharply from near $0$ to a range of $\sim$20\% to $\sim$70\% across the six \textit{VQA} tasks, around the model's middle layers. This implies that the model rapidly acquires the capability to predict correct answers in these middle layers, where the phase of multimodal information integration has just fully completed (see \Cref{fig:imagepatchesToquestion_13b}) and the multimodal information is still transforming from \textit{question} position to \textit{last position} (see \Cref{fig:tolast_13b}).
\vspace{-10pt}
\paragraph{Observation 2: Semantic generation is followed by syntactic refinement} 
As shown in \Cref{fig:last_position_prob_13b}, across all \textit{VQA} tasks, the probability of \textit{Noncapitalized Answer} starts to gradually decrease to nearly zero after an increase in middle layers. In contrast, the probability of \textit{Capitalized Answer} remains low in the initial layers following $20$th but starts to increase in subsequent layers. This indicates the model has already semantically inferred the answer by about halfway through layers and in the higher layers, the model starts to refine the syntactic correctness of the answer. Similar findings on other models are shown in \Cref{app: exp on other model}.

% \begin{figure*}[t]
%     \centering
% \subfloat[Image to patches]{
%     \includegraphics[width=0.3\linewidth]{images/GQA_val_correct_question_with_choose_NoPositionChoose_Only1Step_window9_Image___Question_first.pdf}
% \label{fig:indect_image_with_object_patches_to_ques_firstgen}
% }
% \subfloat[Other patches to question]{
%     \includegraphics[width=0.3\linewidth]{images/GQA_val_correct_question_with_choose_NoPositionChoose_Only1Step_window9_Image_Without_Central_Object___Question_first.pdf}
% \label{fig:indect_image_without_object_patches_to_ques_firstgen}
% }
% \subfloat[Object patches to question]{
%     \includegraphics[width=0.3\linewidth]{images/GQA_val_correct_question_with_choose_NoPositionChoose_Only1Step_window9_Image_Central_Object___Question_first.pdf}
% \label{fig:indect_image_central_object_patches_to_ques_firstgen}
% }
% \caption{fine-grained image information transfer}
% \end{figure*}

%% file: sec/8_conclusion.tex
\section{Conclusion}
In this paper, we unveil the inner working mechanisms of auto-regressive multimodal large language models in handling multimodal tasks. Our experiments reveal that different multimodal tasks exhibit similar processing patterns within the model. Specifically, when provided with an input consisting of an image and a question, within the lower-to-middle layers, the model initially propagates the overall image information into the hidden representations of the \textit{question} in the lower layers and then the model selectively transfers only the question-relevant image information into the hidden representations of the \textit{question}, facilitating multimodal information integration. In the middle layers, this integrated multimodal information is propagated to the hidden representation of the \textit{last position} for the final prediction. In addition, we find that the answers are initially generated in lowercase form in middle layers and then converted to uppercase for the first letter in higher layers.
These findings enhance the transparency of such models, offering new research directions for better understanding the interaction of the two modalities in MLLMs and ultimately leading to improved model designs.

%% file: sec/X_suppl.tex
\clearpage
\setcounter{page}{1}
\maketitlesupplementary

\begin{table}[t]
\setlength{\tabcolsep}{2pt}
\begin{tabular}{l|ccccc}
& {\color[HTML]{000000} \textbf{Object}} & {\color[HTML]{000000} \textbf{Attribute}} & {\color[HTML]{000000} \textbf{Category}} & {\color[HTML]{000000} \textbf{Relation}} & {\color[HTML]{000000} \textbf{Global}} \\ \midrule
{\color[HTML]{000000} \textbf{Verify}}  & \cellcolor[HTML]{F9CBAA}86.21\%        & \cellcolor[HTML]{F9CBAA}83.00\%           & --                                        & \cellcolor[HTML]{F9CBAA}87.82\%          & \cellcolor[HTML]{F9CBAA}95.56\%        \\
{\color[HTML]{000000} \textbf{Query}}   & --                                      & \cellcolor[HTML]{C9E4B4}71.20\%           & \cellcolor[HTML]{C9E4B4}62.88\%          & \cellcolor[HTML]{C9E4B4}52.84\%          & \cellcolor[HTML]{C9E4B4}55.74\%        \\
{\color[HTML]{000000} \textbf{Choose}}  & --                                      & \cellcolor[HTML]{C9E4B4}90.17\%           & \cellcolor[HTML]{C9E4B4}92.03\%          & \cellcolor[HTML]{C9E4B4}87.19\%          & \cellcolor[HTML]{C9E4B4}96.76\%        \\
{\color[HTML]{000000} \textbf{Logical}} & \cellcolor[HTML]{F9CBAA}88.92\%        & \cellcolor[HTML]{F9CBAA}76.17\%           & --                                        & --                                        & --                                      \\
{\color[HTML]{000000} \textbf{Compare}} &--                                      & \cellcolor[HTML]{CBCEFB}71.23\%           & --                                        & --                                        & --  
\end{tabular}
\caption{The accuracy of the validation set of  GQA dataset\cite{hudson2019gqa} on \textit{LLaVA-1.5-13b} \cite{Liu2023llava1_5}. \colorbox[HTML]{F9CBAA}{\makebox[0.5cm][c]{\rule{0pt}{0.1cm}}} and \colorbox[HTML]{C9E4B4}{\makebox[0.5cm][c]{\rule{0pt}{0.1cm}}} represent binary (\textit{yes}/\textit{no}) and open question respectively. \colorbox[HTML]{CBCEFB}{\makebox[0.5cm][c]{\rule{0pt}{0.1cm}}} represents that this category contains both binary and open questions.}
\label{tab:gqa acc}
\end{table}

\section{Dataset collection}
\label{app: dataset collection}

We collect our data from the validation set of the \textit{GQA} dataset \citep{hudson2019gqa}, which is designed for visual reasoning and compositional question-answering. Derived from the Visual Genome dataset \citep{krishna2017visual}, \textit{GQA} provides real-world images enriched with detailed scene graphs. Questions in \textit{GQA} dataset are categorized along two dimensions: structure (5 classes, defining question formats) and semantics (5 classes, specifying the main subject’s semantic focus). Structural classes include:
(1) \textit{verify} (\textit{yes}/\textit{no} questions),
(2) \textit{query} (open questions),
(3) \textit{choose} (questions with two alternatives),
(4) \textit{logical} (logical inference), and
(5) \textit{compare} (object comparisons).
Semantic classes are:
(1) \textit{object} (existence questions),
(2) \textit{attribute} (object properties or positions),
(3) \textit{category} (object identification within a class),
(4) \textit{relation} (questions about relational subjects/objects), and
(5) \textit{global} (overall scene properties like weather or location).
Based on the combination of these two dimensions, the questions in \textit{GQA} are categorized into 15 groups, as shown in \Cref{tab:gqa acc}. 

We select 6 out of 15 groups according to the following steps. First, we exclude the \textit{verify} type, as it is quite simple and involves only straightforward binary questions (\textit{e.g.,} "Is the apple red?"). Then we focus on types with an average accuracy above 80\% on \textit{LLaVA-1.5-13b} model \cite{Liu2023llava1_5}, retaining \textit{ChooseAttr}, \textit{ChooseCat}, \textit{ChooseRel}, and \textit{LogicalObj}. \textit{ChooseGlo} is excluded due to its limited sample size in the validation set of \textit{GQA} (only 556 instances). 
After that, to enhance question-type diversity, we select high-performing subtypes (accuracy$>$80\%) in \textit{CompareAttr} and \textit{QueryAttr} from the GQA dataset. Specifically, we use the \textit{positionQuery} subtype for spatial-relation questions in \textit{QueryAttr} and the \textit{twoCommon} subtype for comparing common attributes between two objects in \textit{CompareAttr}. Finally, for each type of the six selected types, we sample at most 1000 data that are predicted correctly on model \textit{LLaVA-1.5-13b} from the validation set of \textit{GQA} resulting in our final data in this paper, as shown in \Cref{tab: question type}. 

%In addition, as \textit{ChooseAttr} contains much more data, which allows us to explore the influence of the option order \citep{liu2023mmbench}, we ensure that 50\%  of sampled questions present the \textit{true option} before the \textit{false option}. 

\begin{figure}[t]
\centering
\includegraphics[width=0.8\linewidth]{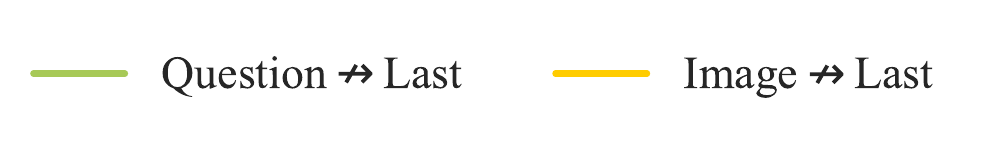}
\subfloat[Window 1]{
    \includegraphics[width=0.32\linewidth]{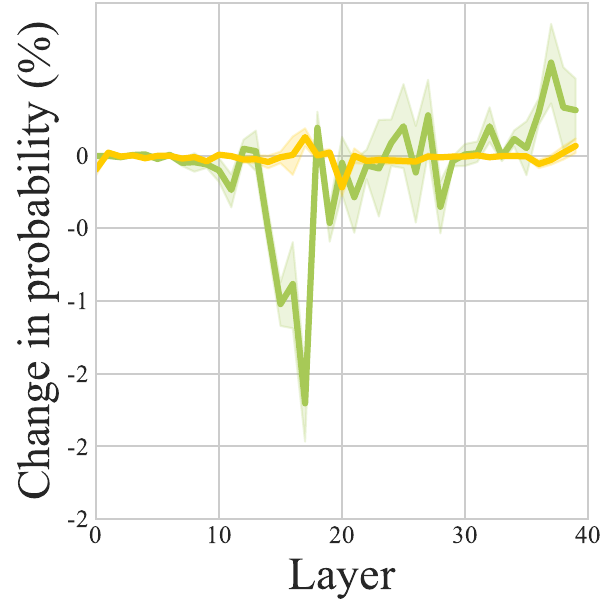}
}
\subfloat[Window 3]{
    \includegraphics[width=0.32\linewidth]{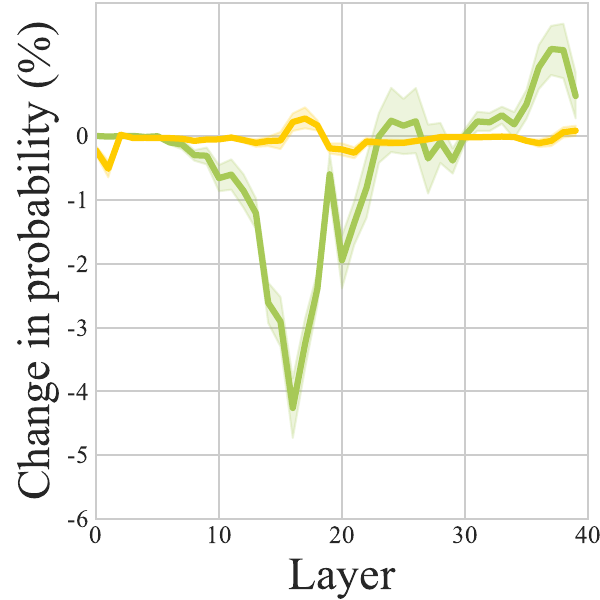}
}
\subfloat[Window 5]{
    \includegraphics[width=0.32\linewidth]{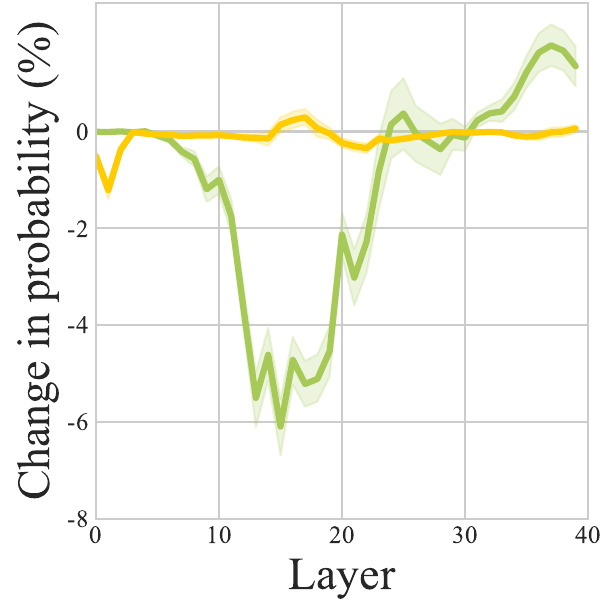}
}\\
\subfloat[Window 7]{
    \includegraphics[width=0.32\linewidth]{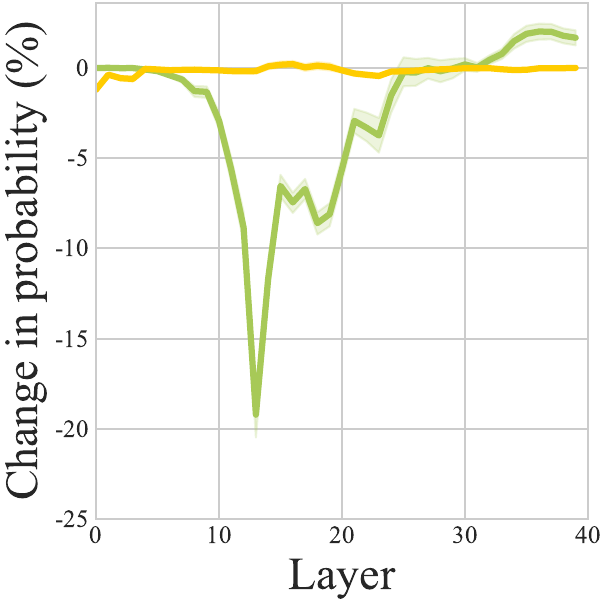}
}
\subfloat[Window 11]{
    \includegraphics[width=0.32\linewidth]{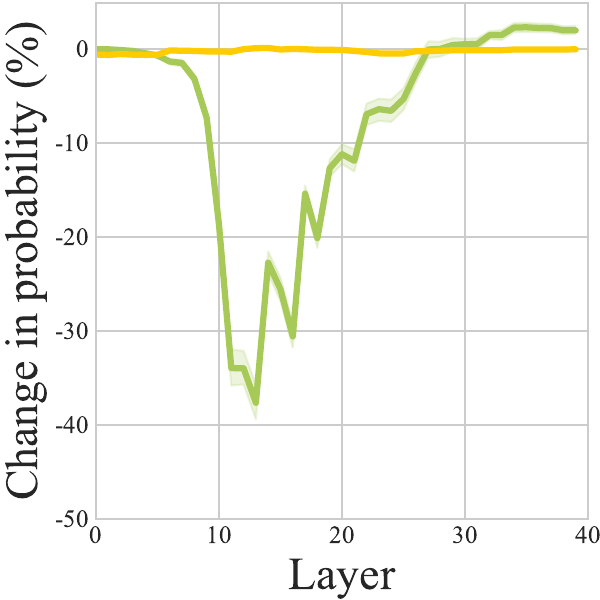}
}
\subfloat[Window 15]{
    \includegraphics[width=0.32\linewidth]{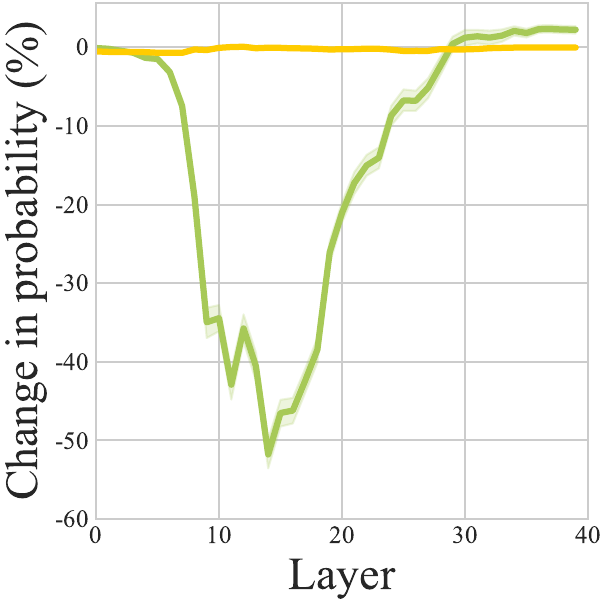}
}
\caption{The relative changes in prediction probability on \textit{LLaVA-1.5-13b} with the tasks of \textit{ChooseAttr} for different window size. The \textit{Question}$\nrightarrow$ \textit{Last} and \textit{Image}$\nrightarrow$ \textit{Last} represent preventing \textit{last position} from attending to \textit{Question} and \textit{Image} respectively.}
\label{fig:tolast_LLaVA-1.5-13b_diff_wins}
\end{figure}

\section{Informaion flow for different window size k}
\label{app: different win}
\begin{figure}[t]
\centering
\includegraphics[width=0.4\linewidth]{images/llava_v1_5_13b/image2question/legend.pdf}
\subfloat[Window 1]{
    \includegraphics[width=0.32\linewidth]{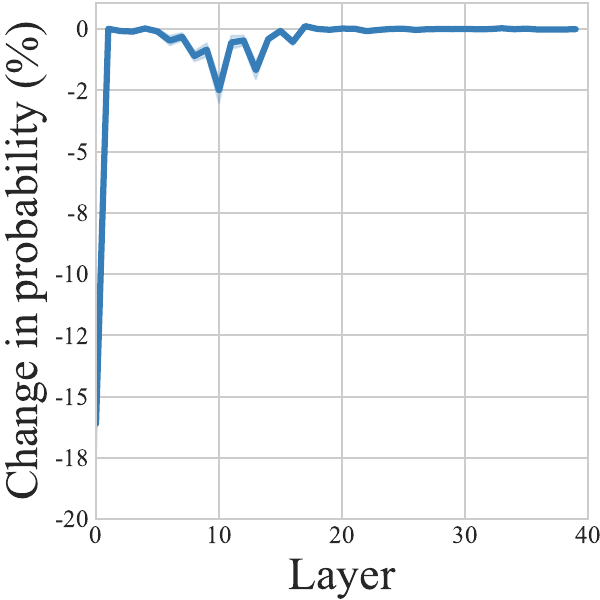}
}
\subfloat[Window 3]{
    \includegraphics[width=0.32\linewidth]{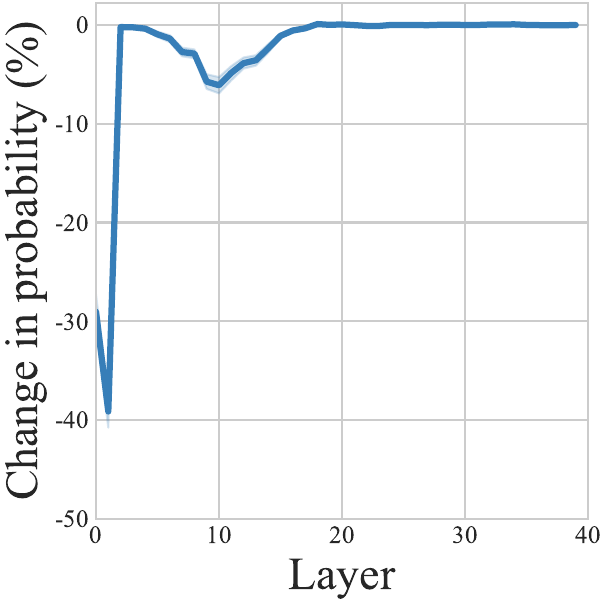}
}
\subfloat[Window 5]{
    \includegraphics[width=0.32\linewidth]{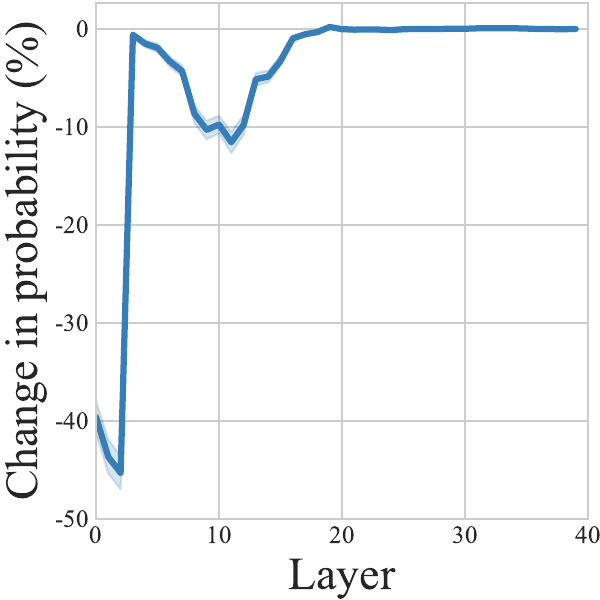}
}\\
\subfloat[Window 7]{
    \includegraphics[width=0.32\linewidth]{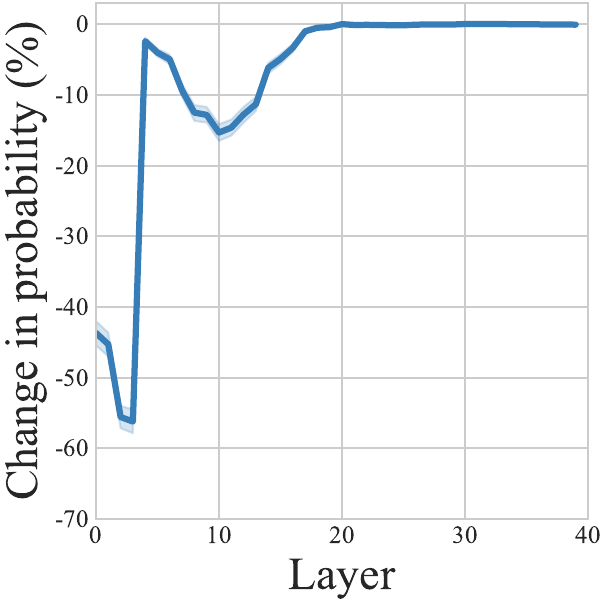}
}
\subfloat[Window 11]{
    \includegraphics[width=0.32\linewidth]{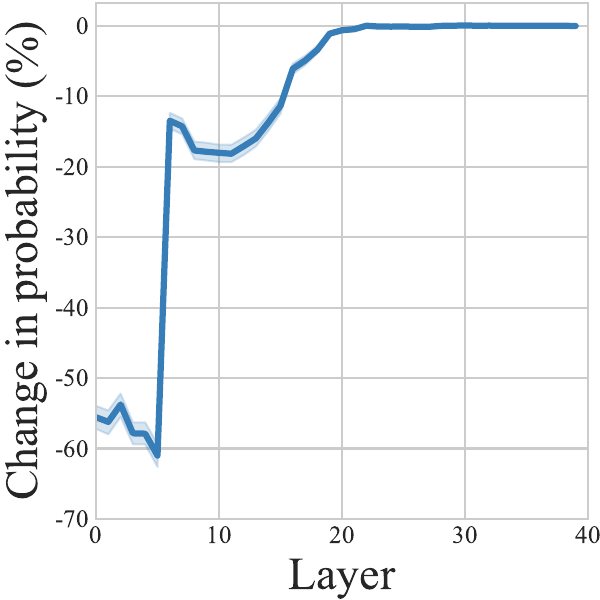}
}
\subfloat[Window 15]{
    \includegraphics[width=0.32\linewidth]{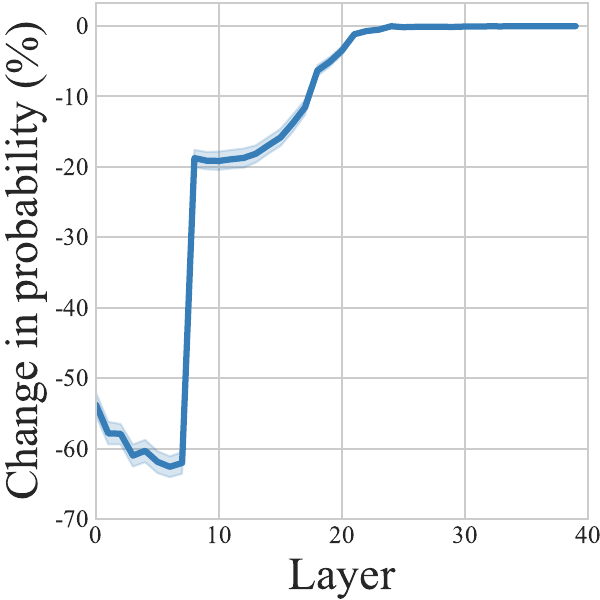}
}
\caption{
The relative changes in prediction probability when blocking attention edges from the \textit{question} positions to the \textit{image} positions on \textit{LLaVA-1.5-13b} with the tasks of \textit{ChooseAttr} for different window sizes.
}
\label{fig:imageToquestion_LLaVA-1.5-13b_diff_wins}
\end{figure}

In the main body of the paper, we use a window size $k=9$ for an easier analysis of the internal working mechanism of the multimodal large language models when performing multimodal tasks. We present the relative change in probability on \textit{ LLaVA-1.5-13b} and the task of \textit{ChooseAttr} with different window sizes of $k=1,3,5,7,9,11,15$. The resulting information flow between different parts of the input sequence (\textit{image} and \textit{question}) and \textit{last position}, and between \textit{image} and \textit{question} are shown in \Cref{fig:tolast_LLaVA-1.5-13b_diff_wins} and \Cref{fig:imageToquestion_LLaVA-1.5-13b_diff_wins}, respectively. Overall, the observations on the information flow are consistent across different window sizes $k$. Specifically, the critical information flow from \textit{question} to \textit{last position} occurs in the middle layers while the critical information flow from \textit{image} to \textit{last position} is not observed across different window sizes, as shown in \Cref{fig:tolast_LLaVA-1.5-13b_diff_wins}. For critical information from \textit{image} to \textit{question},  the two different critical information flows are observed across different window sizes where both occur in lower layers and sequentially follow each other, as illustrated in \Cref{fig:imageToquestion_LLaVA-1.5-13b_diff_wins}. In addition, we observe that as the window size increases, the two information flows gradually merge into one, which is because the larger window encompasses layers involved in both information flows. 
Moreover, the decrease in the prediction probability becomes more pronounced with the increase of the value $k$. This is expected, as blocking more attention edges in the computation hinders the model's ability to properly contextualize the input.

\section{Changes in probability of the last sub-word generation}
\label{app: last generated sub-word}
\begin{figure}[t]
\centering
\includegraphics[width=1\linewidth]{images/llava_v1_5_13b/toLast/legend.pdf}

\centering
\subfloat[ChooseAttr]{
    \includegraphics[width=0.32\linewidth]{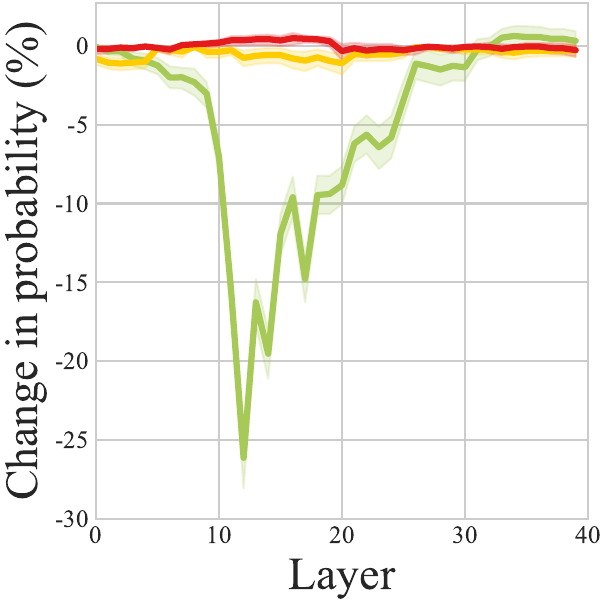}
}
\subfloat[ChooseCat]{
    \includegraphics[width=0.32\linewidth]{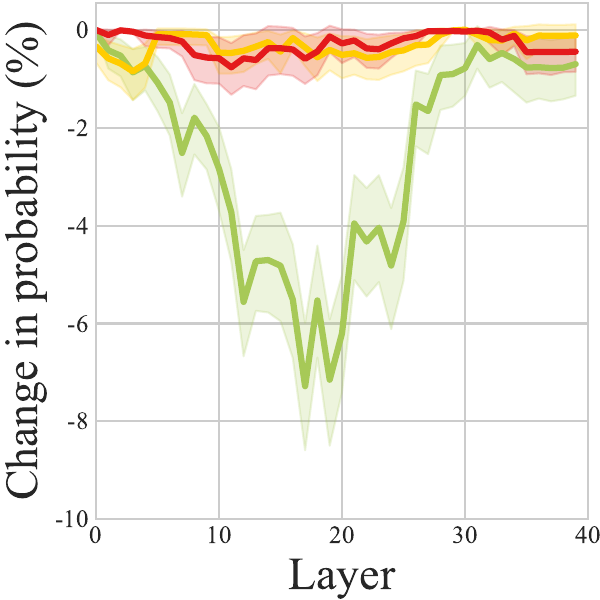}
}
\subfloat[ChooseRel]{
    \includegraphics[width=0.32\linewidth]{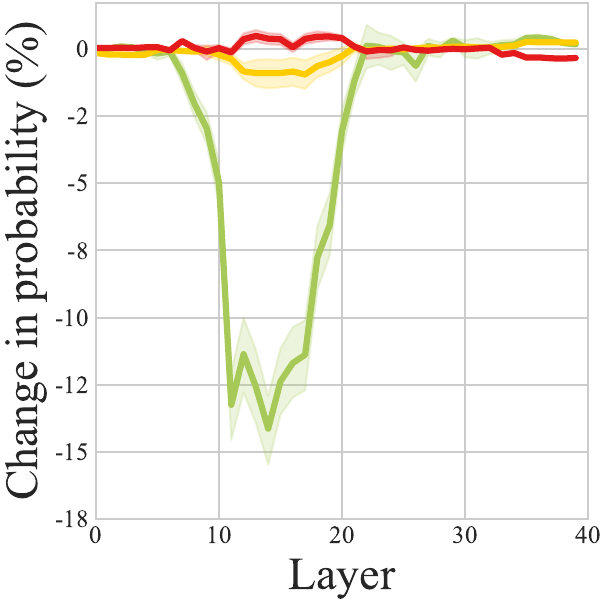}
}\\
\subfloat[CompareAttr]{
    \includegraphics[width=0.32\linewidth]{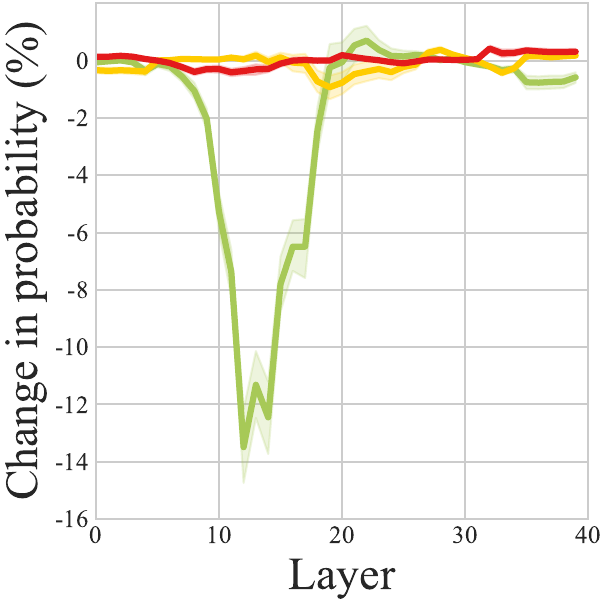}
}
\subfloat[LogicalObj]{
    \includegraphics[width=0.32\linewidth]{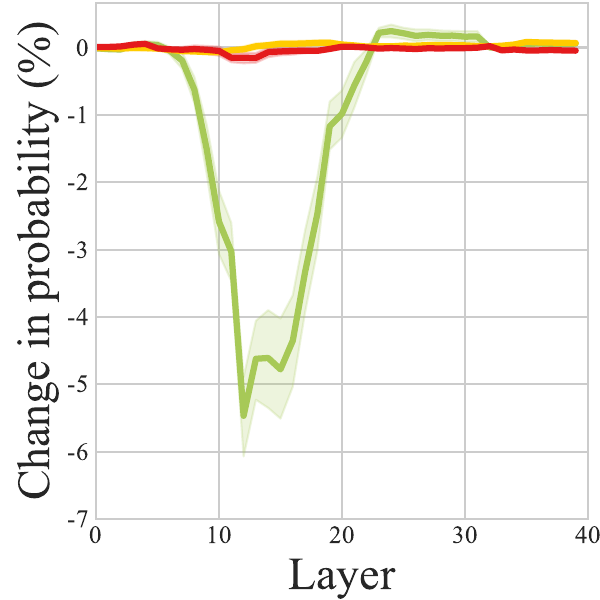}
}
\subfloat[QueryAttr]{
    \includegraphics[width=0.32\linewidth]{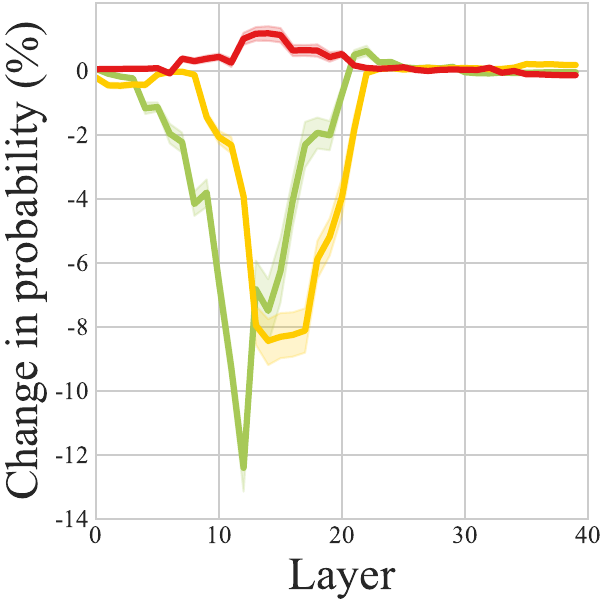}
}
\caption{The relative changes in prediction probability for the final generated sub-word of the \textit{answer} on \textit{LLaVA-1.5-13b} with six \textit{VQA} tasks. The \textit{Question}$\nrightarrow$ \textit{Last}, \textit{Image}$\nrightarrow$ \textit{Last} and  \textit{Last}$\nrightarrow$ \textit{Last} represent preventing \textit{last position} from attending to \textit{Question}, \textit{Image} and itself respectively.}
\label{fig:tolast_LLaVA-1.5-13b_last_generated_sub-word}
\end{figure}

\begin{figure}[t]
\centering
\includegraphics[width=0.4\linewidth]{images/llava_v1_5_13b/image2question/legend.pdf}
\centering
\subfloat[ChooseAttr]{
    \includegraphics[width=0.32\linewidth]{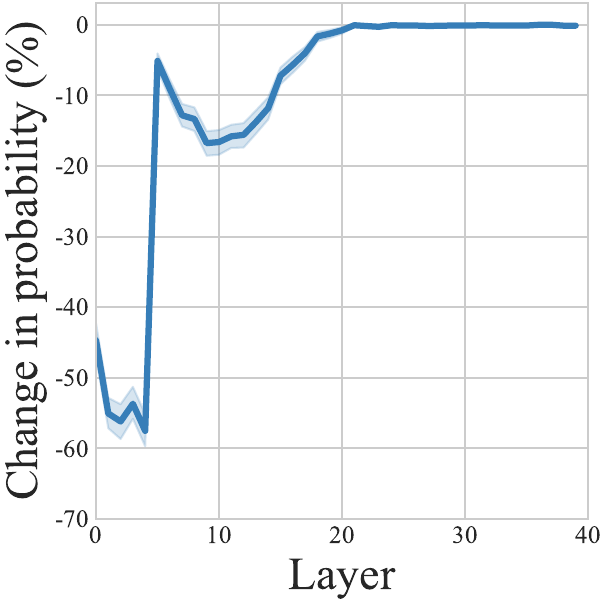}
}
\subfloat[ChooseCat]{
    \includegraphics[width=0.32\linewidth]{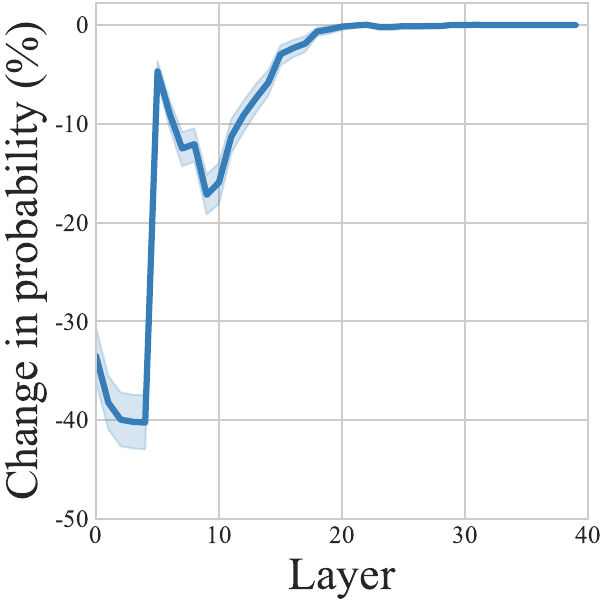}
}
\subfloat[ChooseRel]{
    \includegraphics[width=0.32\linewidth]{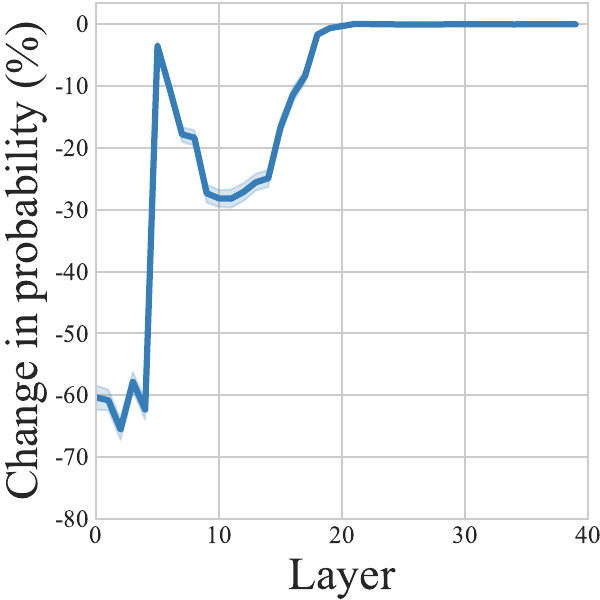}
}\\
\subfloat[CompareAttr]{
    \includegraphics[width=0.32\linewidth]{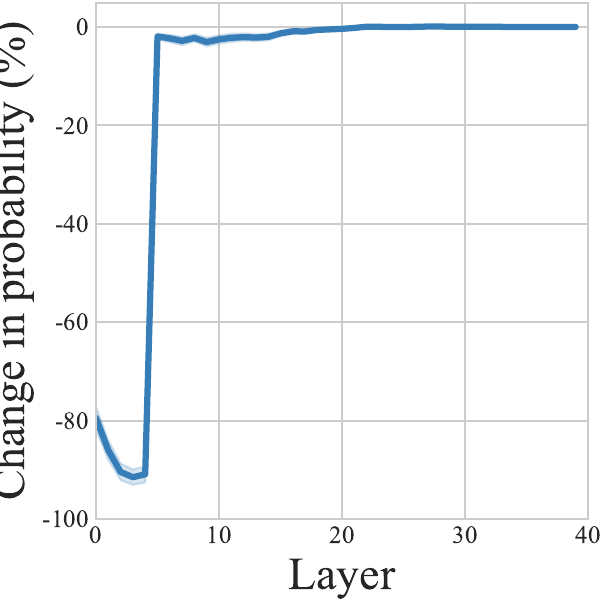}
}
\subfloat[LogicalObj]{
    \includegraphics[width=0.32\linewidth]{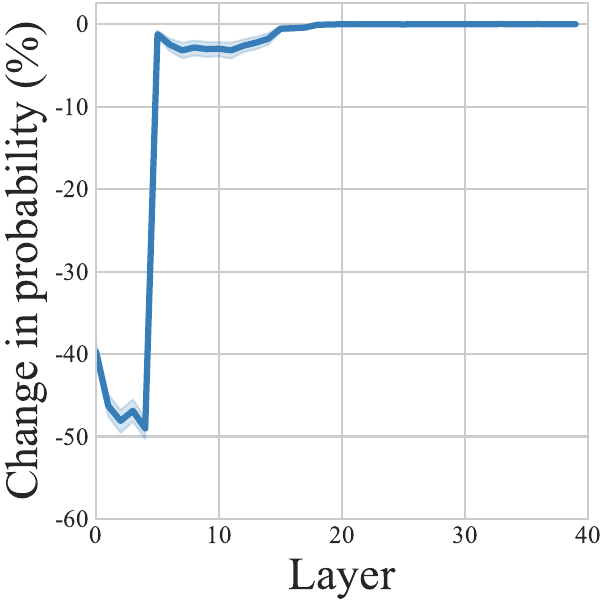}
}
\subfloat[QueryAttr]{
    \includegraphics[width=0.32\linewidth]{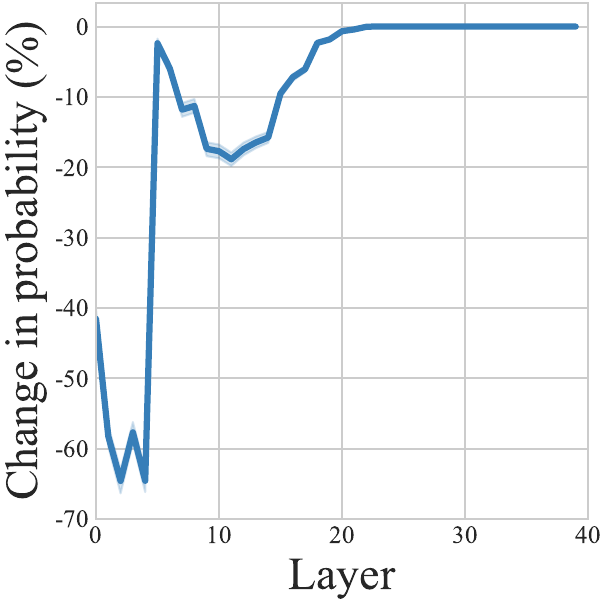}
}
\caption{
The relative changes in prediction probability for the final generated sub-word of the \textit{answer} when blocking attention edges from the \textit{question} positions to the \textit{image} positions on \textit{LLaVA-1.5-13b} with six \textit{VQA} tasks.
}
\label{fig:imageToquestion_LLaVA-1.5-13b_sub-word}
\end{figure}

\begin{figure}[t]
\centering
\includegraphics[width=1.0\linewidth]{images/llava_v1_5_13b/imagePatches2question/legend.pdf}
\\
\centering
\subfloat[ChooseAttr]{
    \includegraphics[width=0.32\linewidth]{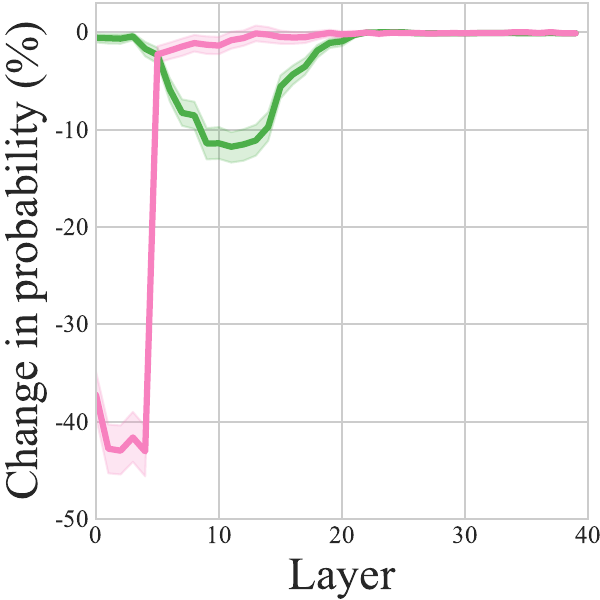}
}
\subfloat[ChooseCat]{
    \includegraphics[width=0.32\linewidth]{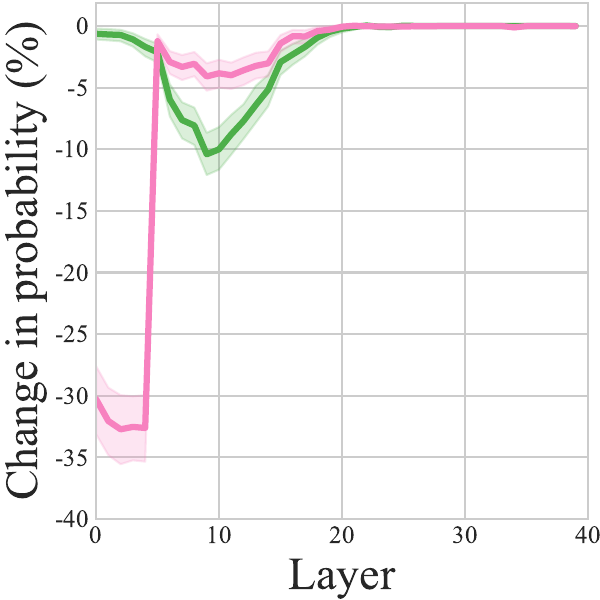}
}
\subfloat[ChooseRel]{
    \includegraphics[width=0.32\linewidth]{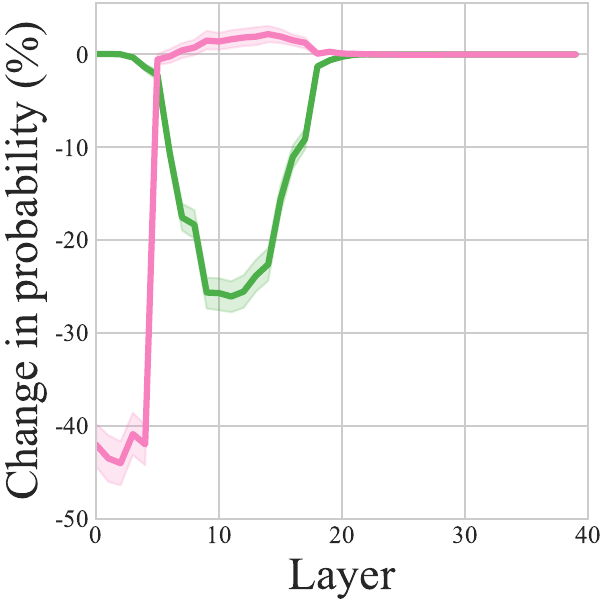}
}\\
\subfloat[CompareAttr]{
    \includegraphics[width=0.32\linewidth]{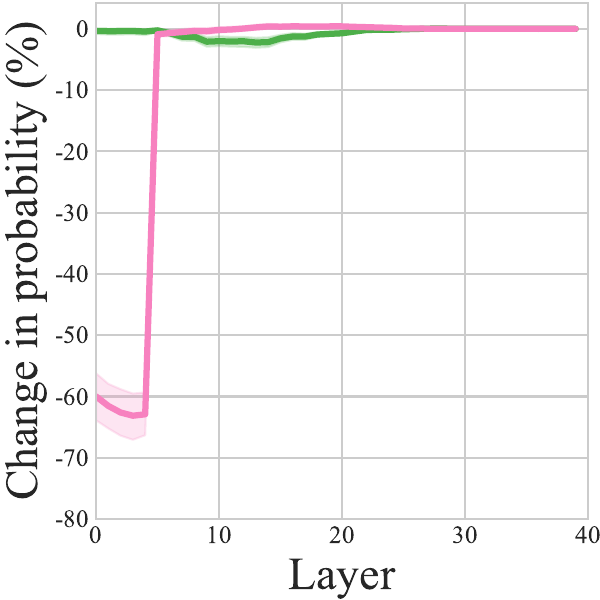}
}
\subfloat[LogicalObj]{
    \includegraphics[width=0.32\linewidth]{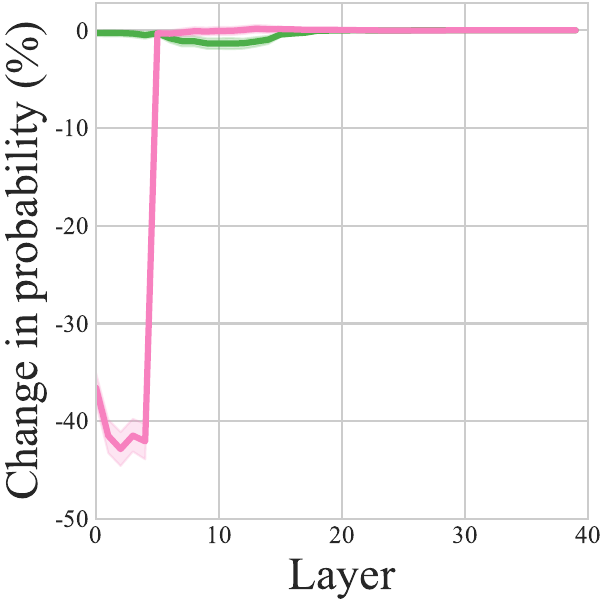}
}
\subfloat[QueryAttr]{
    \includegraphics[width=0.32\linewidth]{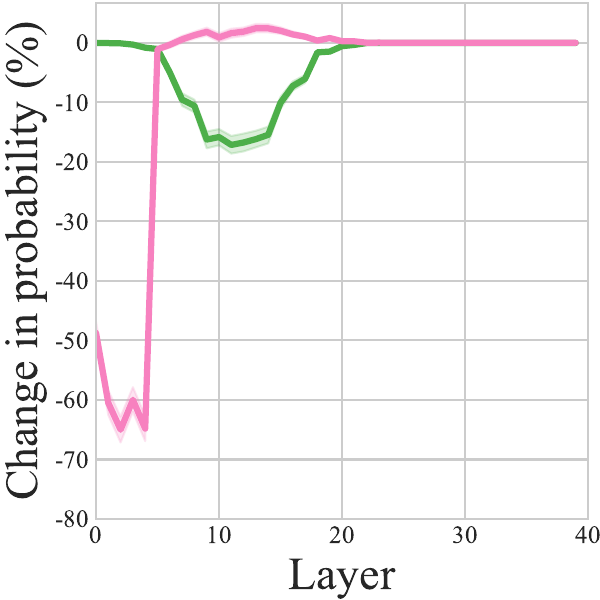}
}
\caption{The relative changes in prediction probability for the final generated sub-word of the \textit{answer} on \textit{LLaVA-1.5-13b} with six \textit{VQA} tasks.
\textit{Related Image Patches}$\nrightarrow$\textit{question} and \textit{Other Image Patches}$\nrightarrow$\textit{question} represent blocking the position of \textit{question} from attending to that of different image patches, region of interest and remainder, respectively.}
\label{fig:imagepatchesToquestion_LLaVA-1.5-13b_sub-word}
\end{figure}

In this paper, the \textit{answer} in our used dataset normally contains one word or one phrase, which might result in several sub-word tokens. 
In the main body of the paper, we present the relative change in probability of the first generated sub-word while the final generated sub-words also yield similar results. Specifically. we conduct the same experiments as in the main body of the paper: six tasks (\textit{ChooseAttr}, \textit{ChooseCat}, \textit{ChooseRel}, \textit{LogicalObj}, \textit{QueryAttr} and \textit{CompareAttr}) on \textit{LLaVA-1.5-13b} model. Instead of calculating the relative change in probability for the first generated sub-word token, we calculate that for the final generated sub-word token of the correct answer word.  As shown in \Cref{fig:tolast_LLaVA-1.5-13b_last_generated_sub-word}, \Cref{fig:imageToquestion_LLaVA-1.5-13b_sub-word} and \Cref{fig:imagepatchesToquestion_LLaVA-1.5-13b_sub-word}, the information flow from different parts of the input sequence (\textit{image} and \textit{question}) to \textit{last position}, from \textit{image} to \textit{question} and from different image patches (\textit{related image patches} and \textit{other image patches}) to \textit{question} are consistent with the observations in \Cref{fig:tolast_13b}, \Cref{fig:imageToquestion_13b} and \Cref{fig:imagepatchesToquestion_13b} in the main body of the paper.

\section{Constructing multimodal semantic representations}
\label{app: constructing multimodal semantic representations}
We have investigated how multimodal information is integrated through the MHAT module in \Cref{sec:how integrate}. We now take a closer look at how the multimodal semantic representation is constructed. 

\paragraph{Experiment} 
To identify which module in the transformer contributes to the formulation of multimodal semantic information within hidden representations, we employ a \textit{module knockout} approach to evaluate the significance of individual transformer modules. As shown in \Cref{eq:hidden}, the hidden representation at layer $\ell$ is computed by adding $\va_i^{\ell}$ and $\vf_i^{\ell}$ to $\vh_i^{\ell-1}$, where $\va_i^{\ell}$ and $\vf_i^{\ell}$ are derived from the MHAT (\Cref{eq:attention}) and MLP (\Cref{eq:ffn}) modules, respectively. This allows us to determine which module contributes to constructing semantic information by selectively zeroing out the outputs of MHAT or MLP---two additive modules in the transformer layer.
Specifically, for each layer $\ell$, we intervene by setting $\va_i^{\ell^\prime}$ or $\vf_i^{\ell^\prime}$ ($i \in \sQ$) to zero across 9 consecutive layers {\small \( \{\ell^\prime\}_{\ell^\prime=\ell}^{\min\{\ell+8, L\}} \)}. We then measure the importance of constructing multimodal semantic information by observing the semantic change of the hidden representation corresponding to \textit{question} position $\sQ$ at the final layer $L$.
Our focus on layer $L$ is inspired by \citet{geva_dissecting_2023}, who highlight that semantic information peaks in the final layer. %Normally, automatically measuring semantic information is challenging. 
We follow \citet{wang_label_2023}, who evaluate the semantic content of a hidden representation using top-k words from this representation. We estimate semantic content using the top-10 words predicted from each hidden representation in $\sQ$, derived from \Cref{eq:output_distribution}, where $\vh_N^L$ is replaced by $\vh_i^L$ ($i \in \sQ$). We then quantify the change in semantic content of hidden representation resulting from our interventions using Jaccard Similarity:
\begin{equation}
    J(\sW_o, \sW_i) = \frac{|\sW_o \cap \sW_i|}{|\sW_o \cup \sW_i|}
\end{equation}
where \( \sW_o \) and \( \sW_i \) denote the sets of $10 \cdot |\sQ|$ predicted words from the original and intervened models, respectively.

\begin{figure}[tb]
\centering
\includegraphics[width=0.4\linewidth]{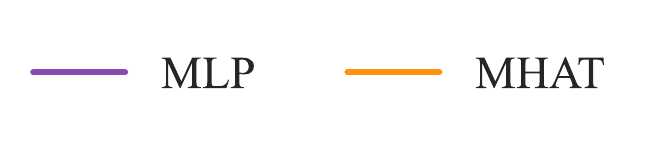}
\centering
\subfloat[ChooseAttr]{
\includegraphics[width=0.32\linewidth]{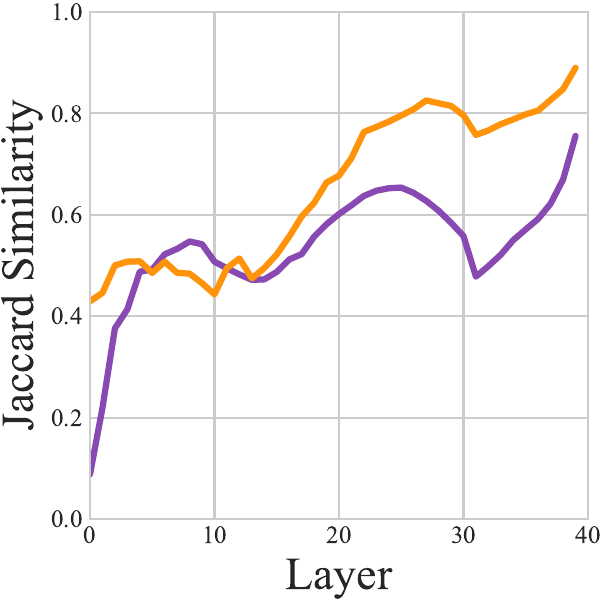}
}
\subfloat[ChooseCat]{
\includegraphics[width=0.32\linewidth]{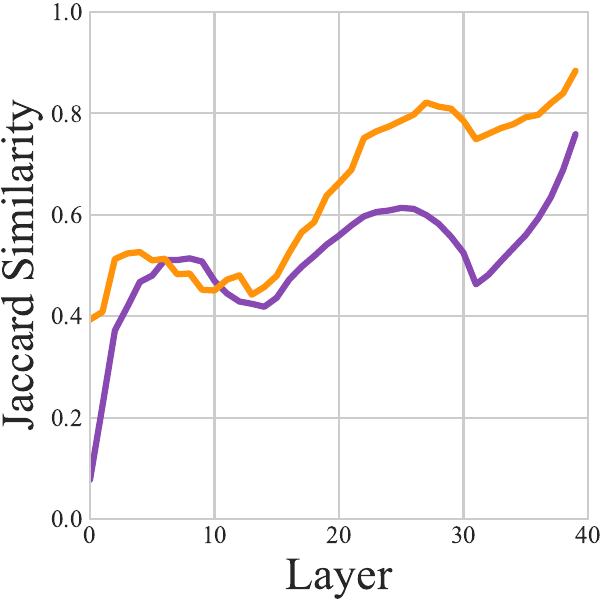}
}
\subfloat[ChooseRel]{
\includegraphics[width=0.32\linewidth]{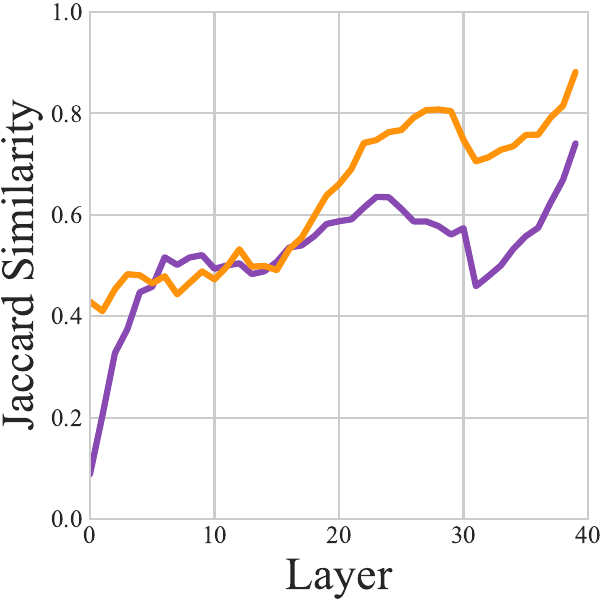}
}\\
\subfloat[CompareAttr]{
\includegraphics[width=0.32\linewidth]{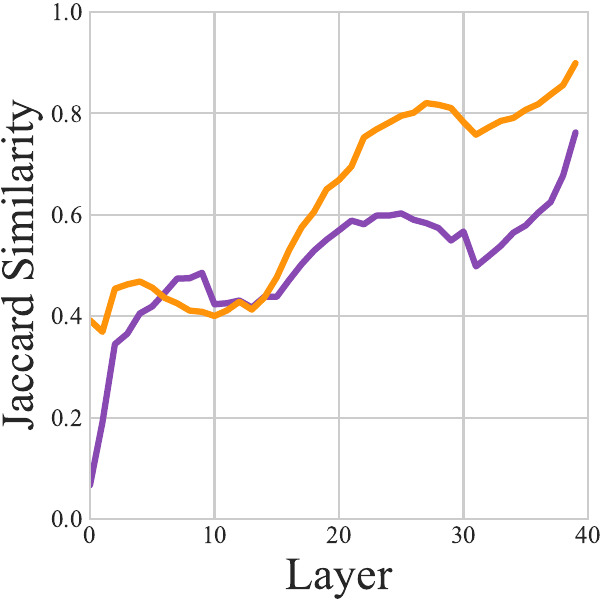}
}
\subfloat[LogicalObj]{
\includegraphics[width=0.32\linewidth]{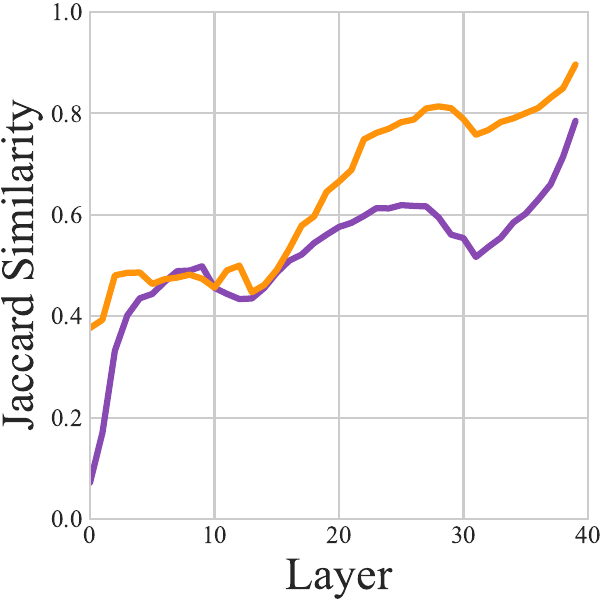}
}
\subfloat[QueryAttr]{
\includegraphics[width=0.32\linewidth]{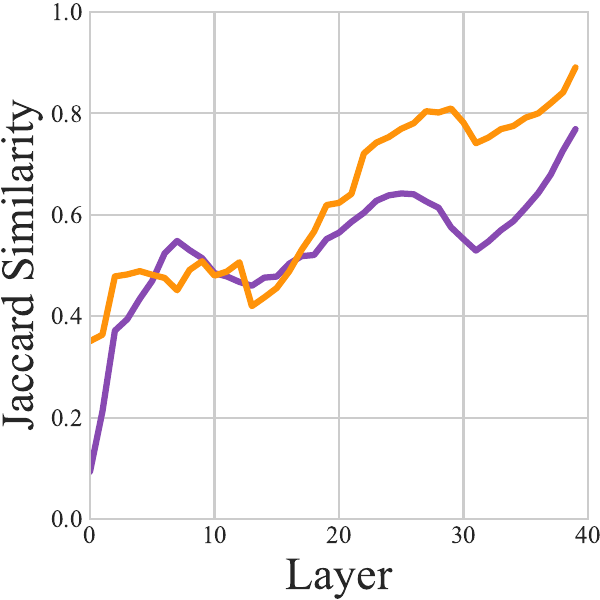}
}
\caption{The Jaccard similarity between the predicted words of the original model and those of the intervened model, with the MLP and MHAT modules removed individually (\textit{LLaVA-1.5-13b}).}
\label{fig:semantic_13b}
\end{figure}

\begin{figure}[tb]
\centering
\includegraphics[width=0.4\linewidth]{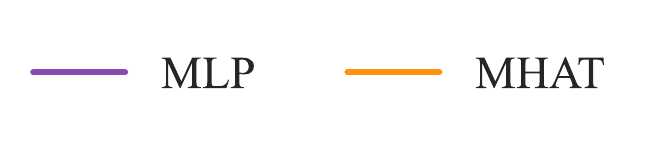}
\centering
\subfloat[ChooseAttr]{
\includegraphics[width=0.32\linewidth]{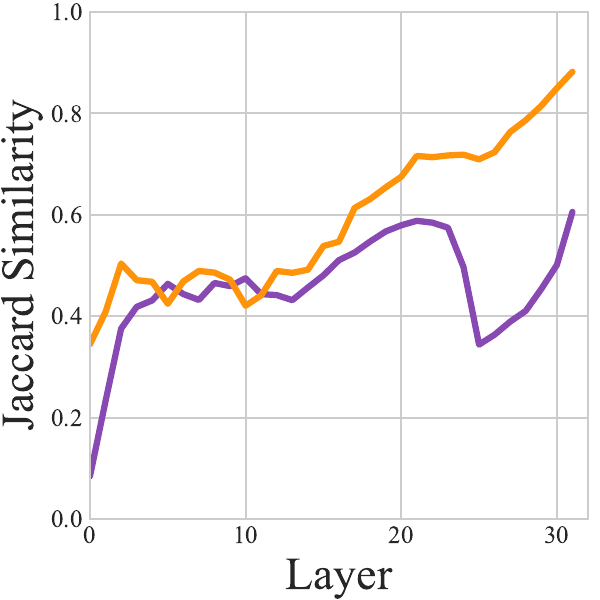}
}
\subfloat[ChooseCat]{
\includegraphics[width=0.32\linewidth]{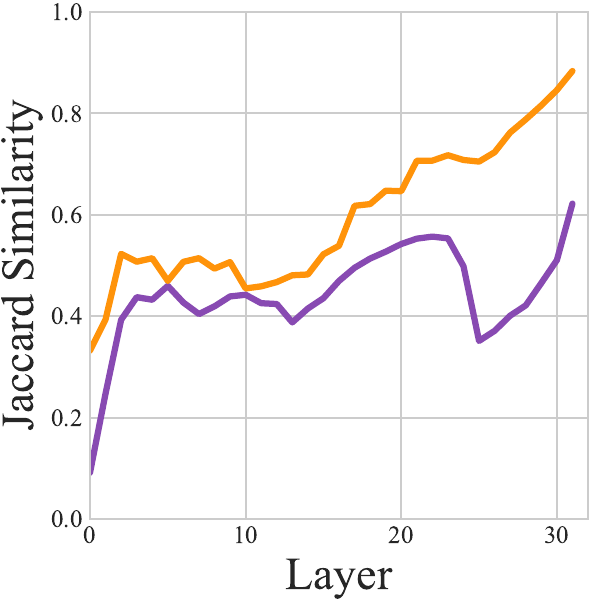}
}
\subfloat[ChooseRel]{
\includegraphics[width=0.32\linewidth]{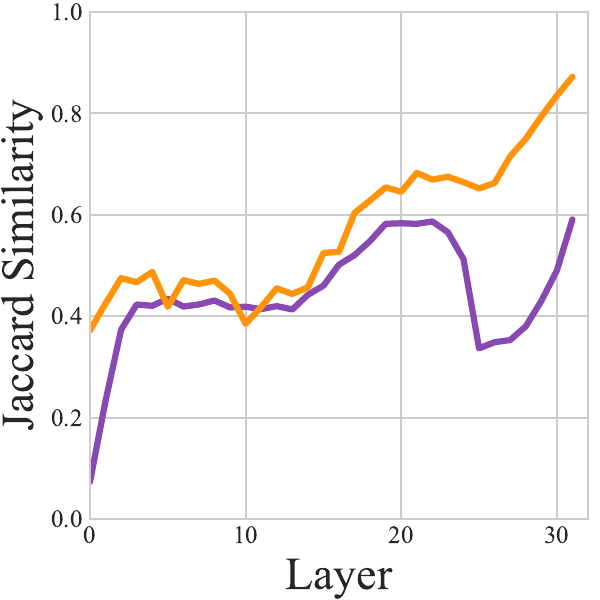}
}\\
\subfloat[CompareAttr]{
\includegraphics[width=0.32\linewidth]{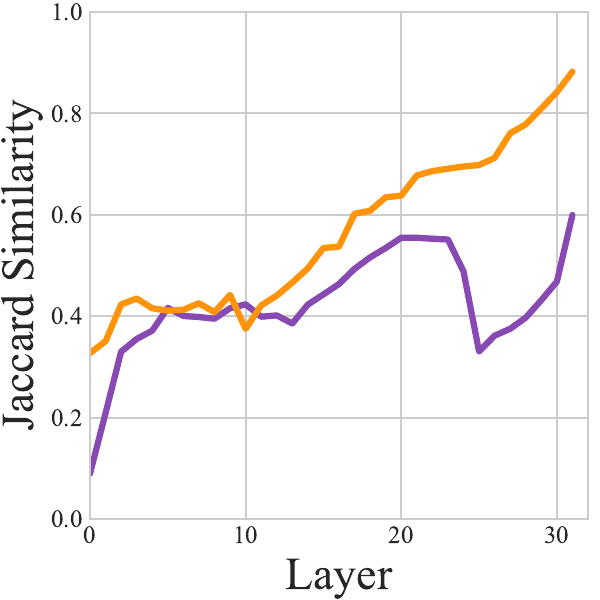}
}
\subfloat[LogicalObj]{
\includegraphics[width=0.32\linewidth]{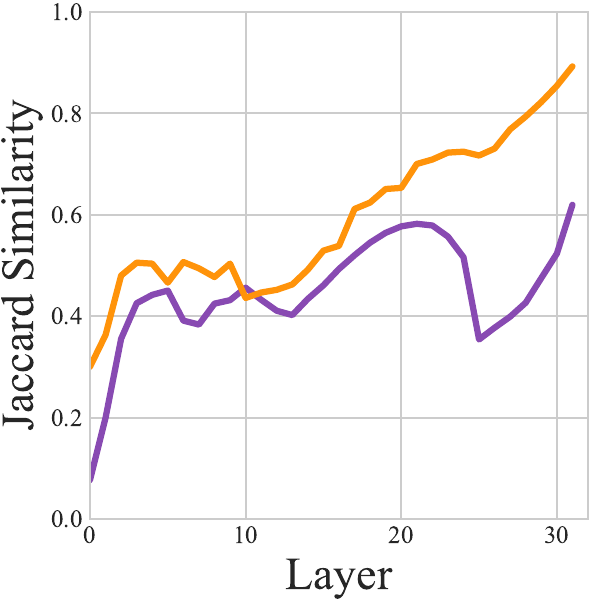}
}
\subfloat[QueryAttr]{
\includegraphics[width=0.32\linewidth]{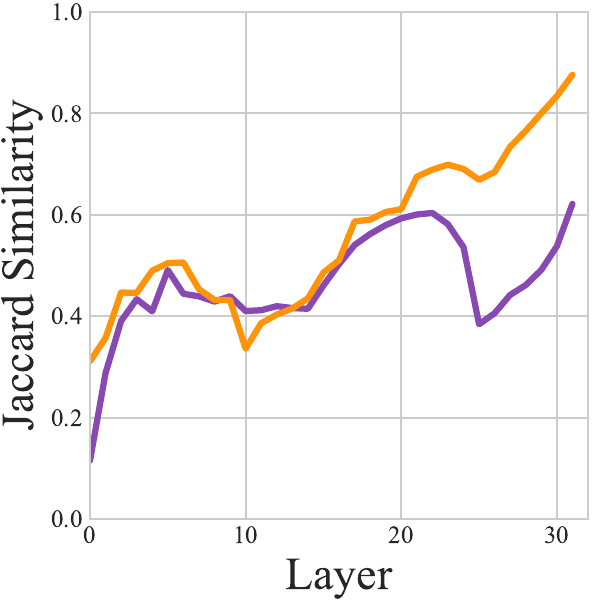}
}
\caption{The Jaccard similarity between the predicted words of the original model and those of the intervened model, with the MLP and MHAT modules removed individually (\textit{LLaVA-1.5-7b}).}
\label{fig:semantic_7b}
\end{figure}
\begin{figure*}[tbh]
\centering
\includegraphics[width=0.30\linewidth]{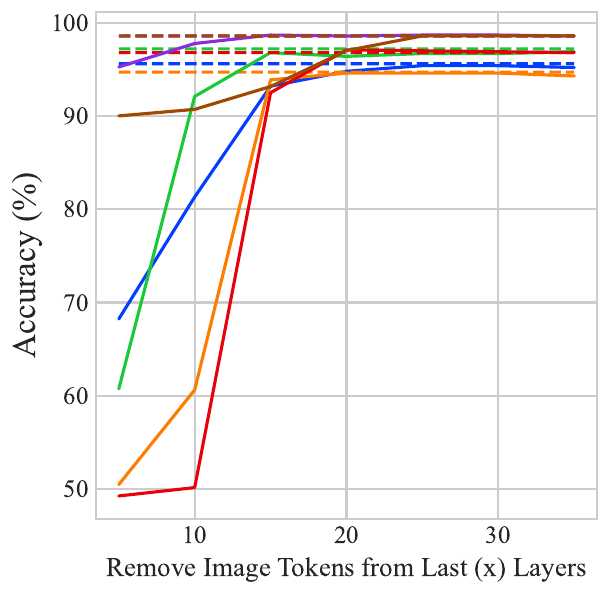}
\includegraphics[width=0.30\linewidth]{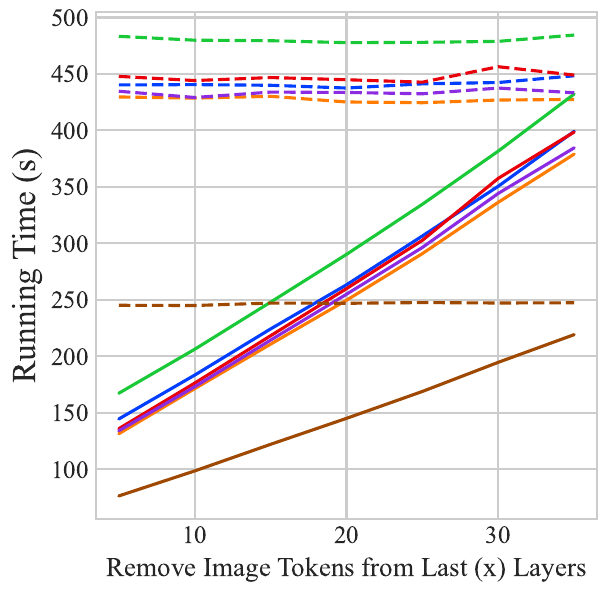}
\includegraphics[trim=60 0 60 0, clip, width=0.32\linewidth]{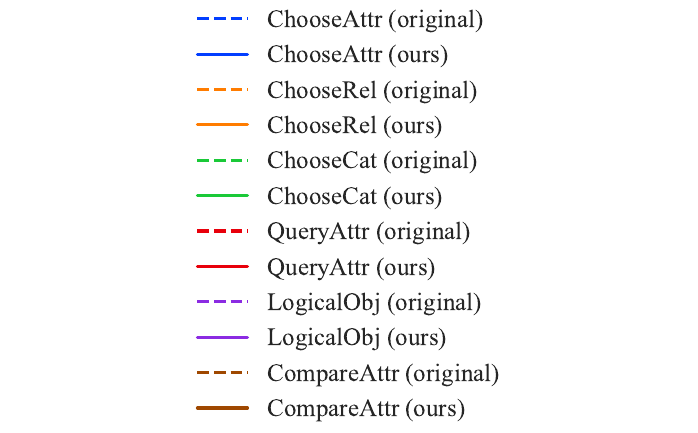}
\caption{Accuracy and inference time of the original \textit{LLaVA-v1.6-13B} and a variant removing image tokens in last certain layers ($X$) across six VQA datasets.
}
\label{fig:application}
\end{figure*}

\paragraph{Observation: The MLP module plays a greater role in constructing semantic representations compared to the MHAT Module}
As shown in \Cref{fig:semantic_13b} for model  \textit{LLaVA-1.5-13b}, removing the MLP module severely impacts semantic representation, reducing average Jaccard Similarity across six tasks by $\sim$90\% when MLP is removed in the first layer and $\sim$25\% in the last layer. In contrast, removing the MHAT module has a smaller effect, with reductions of $\sim$60\% and $\sim$10\% at the first and last layers, respectively. This highlights the MLP module’s important role in generating multimodal semantic representations. These results align with the findings from \cite{Geva_2021, dai_knowledge_2022, meng2022locating}, who demonstrate that factual information is primarily stored in the MLP module, emphasizing its contribution to enriching semantic information. This is also observed in the model \textit{LLaVA-1.5-7b}, as shown in \Cref{fig:semantic_7b}.

\section{Attention sink in image encoder}

\begin{figure}[t]
\centering
\includegraphics[width=1\linewidth]{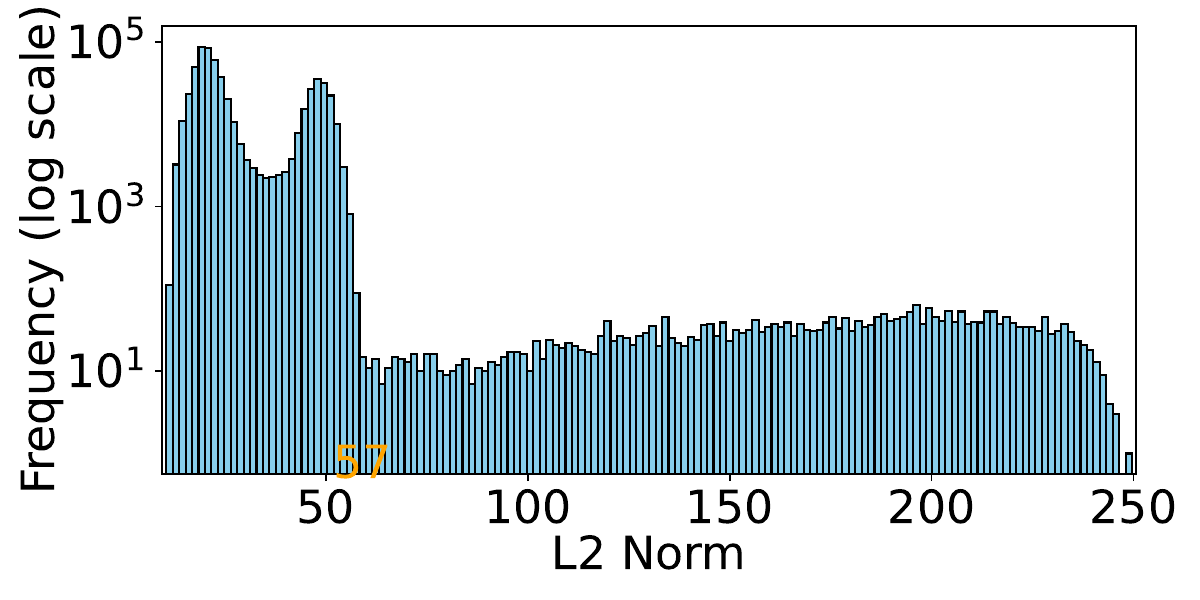}
\caption{Distribution of image patch norms}
\label{fig:attention_sink}
\end{figure}

\begin{figure}[t]
\centering
\includegraphics[width=1.0\linewidth]{images/llava_v1_5_13b/imagePatches2question/legend.pdf}
\\
\includegraphics[width=0.32\linewidth]{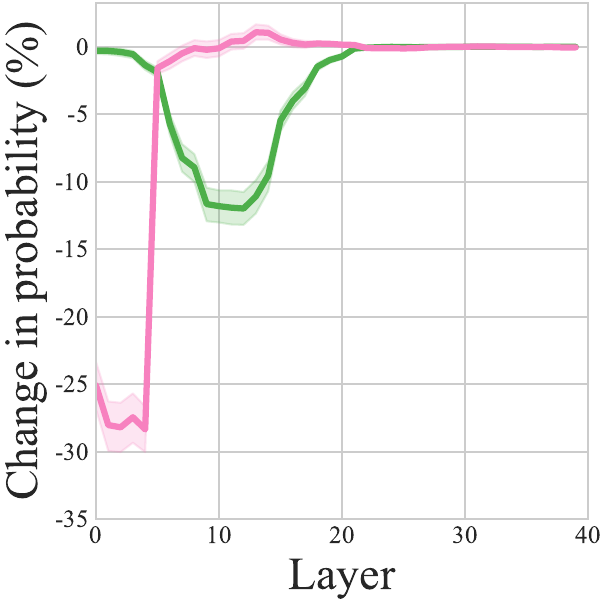}
\includegraphics[width=0.32\linewidth]{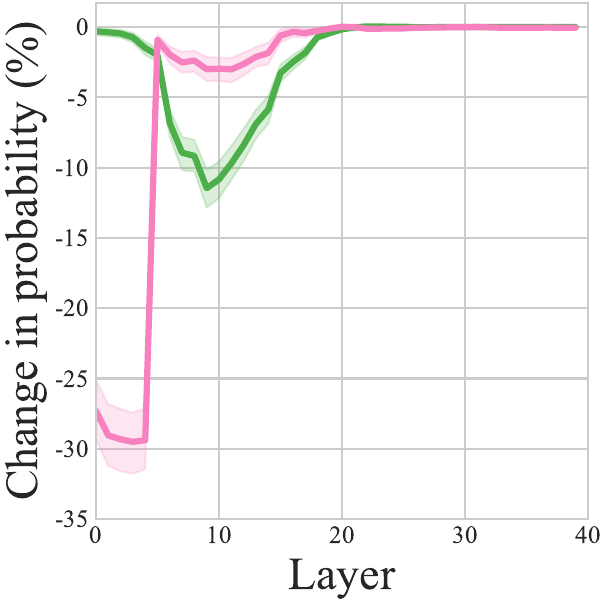}
\includegraphics[width=0.32\linewidth]{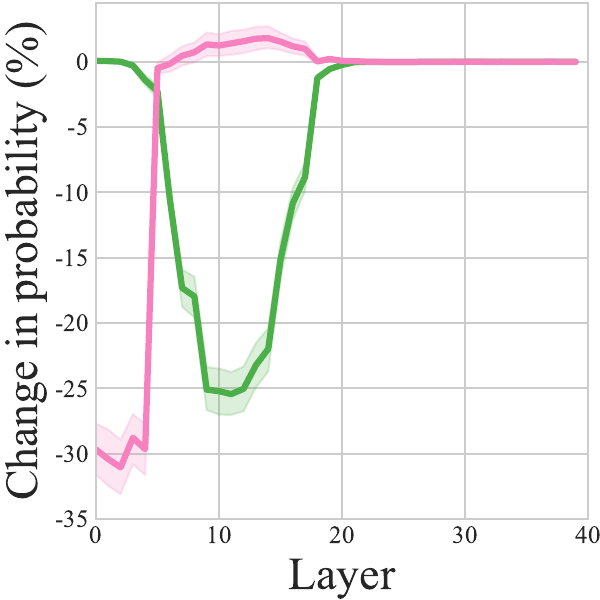}
\includegraphics[width=0.32\linewidth]{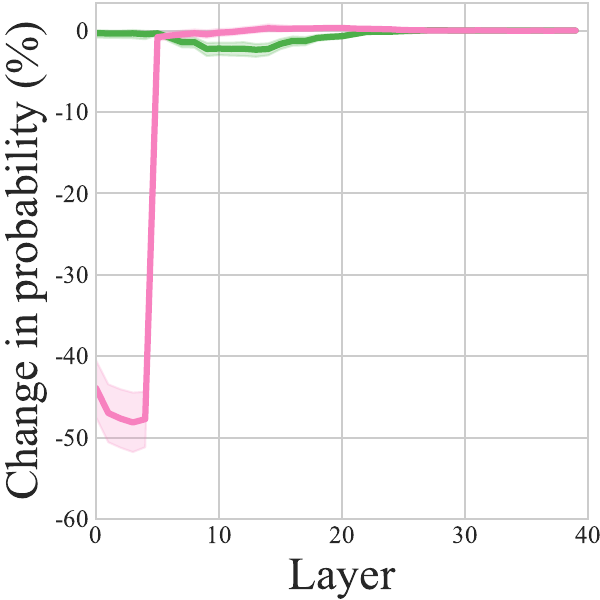}
\includegraphics[width=0.32\linewidth]{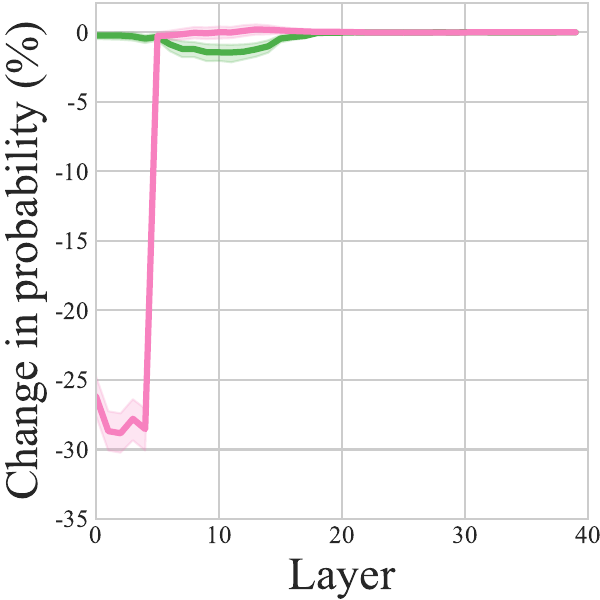}
\includegraphics[width=0.32\linewidth]{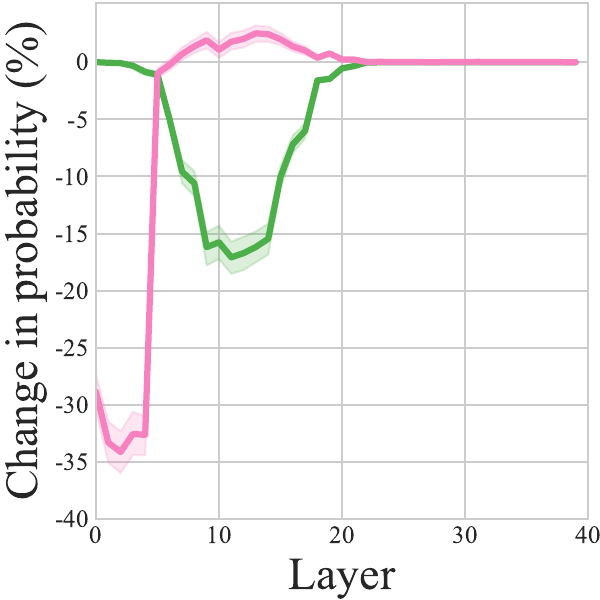}
\caption{The relative changes in prediction probability for the first generated sub-word of the \textit{answer} on \textit{LLaVA-1.5-13b} with six \textit{VQA} tasks.
\textit{Related Image Patches}$\nrightarrow$\textit{question} and \textit{Other Image Patches}$\nrightarrow$\textit{question} represent blocking the position of \textit{question} from attending to that of different image patches (excluding those with high norms), region of interest and remainder, respectively.}
\label{fig:attention_sink}
\end{figure}

The work in \cite{darcet_vision_2024} shows high-norm image patch tokens hold global rather than local information, we analyze the distribution of token norms over our six datasets and identify a threshold of 57, as shown in \cref{fig:attention_sink}. Excluding patches exceeding 57 (3.3 out of 576 patches per image on average), we split the remaining patches into two groups: \textit{Related Image Patched} and \textit{Other Image Patches} and replicate the experiments from our main body of paper (\Cref{sec:how integrate}). As shown in \cref{fig:attention_sink}, the results across six tasks remain consistent with our original findings in the main body of the paper.

\section{Potential application --- model efficiency}

Our findings in the main body of the paper indicate image information primarily propagates to other tokens in the lower layers, suggesting a potential strategy for improving the efficiency of MLLMs. Specifically, during inference, we can remove image tokens in the higher layers to reduce computational costs without significantly compromising performance (average \#image tokens 2,021.6 v.s \#question tokens 11.3 in \textit{LLaVA-v1.6-13B} over our dataset).
To quantitatively analyze this, we remove image tokens in the last 5, 10, 15, 20, 25, 30, 35 layers respectively and test the performance and inference time. As shown in \cref{fig:application}, removing image tokens from final 20 layers keeps accuracy intact while reducing inference time by $\sim$40\%, demonstrating a promising approach for optimizing MLLMs efficiency.

\section{Difference with unimodal LLM}
\begin{figure}[t]
\centering
\includegraphics[width=1.0\linewidth]{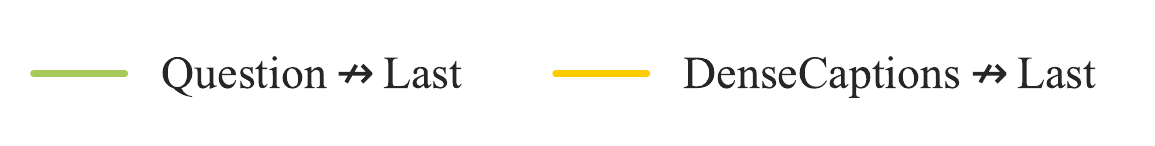}
\\
\includegraphics[width=0.45\linewidth]{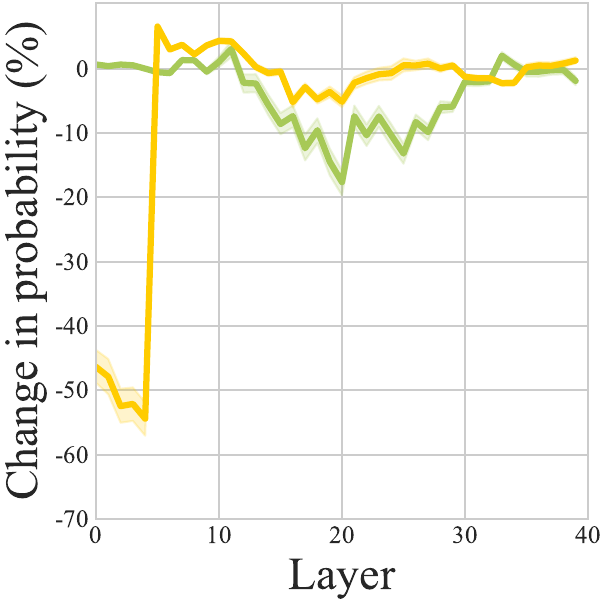}
\includegraphics[width=0.45\linewidth]{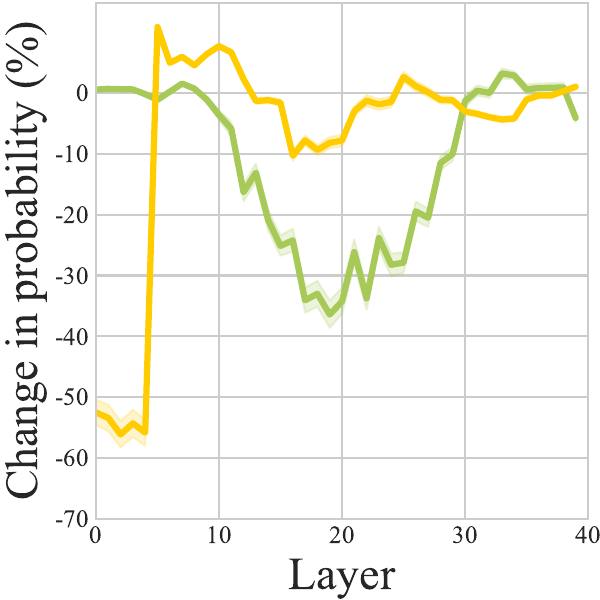}
\caption{Information flow from dense captions or questions to last position in LLM (\textit{vicuna1.5-13b}), the initial LLM in \textit{LLaVA1.5-13b} (MLLM) on two tasks.}
\label{fig:llm}
\end{figure}
To compare the information flow in LLMs and MLLMs, we replace image with dense captions (obtained from VG dataset \cite{krishna2017visual}) and analyze the resulting information flow. As shown in \cref{fig:llm}, we observe that information flow from the question to last position primarily occurs in middle layers which is consistent with MLLMs. However, in LLMs, information flow from dense captions to the last position in lower layers differs from MLLMs where almost no information flow is observed from the image. This suggests that captions and questions follow distinct processing stages, aligning with findings from works in \cite{geva_dissecting_2023}, which identify separate stages of information flow in attribute extraction tasks within LLMs.

\section{Experiment on other factual tasks}
\begin{figure}[t]
\centering
\includegraphics[width=0.45\linewidth]{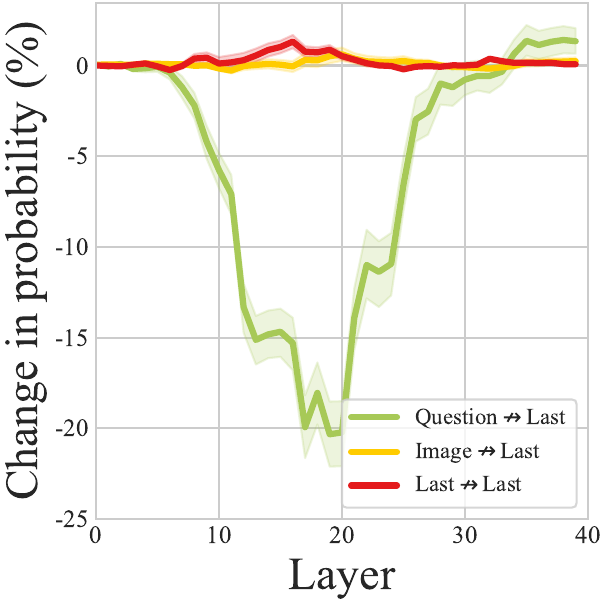}
\includegraphics[width=0.45\linewidth]{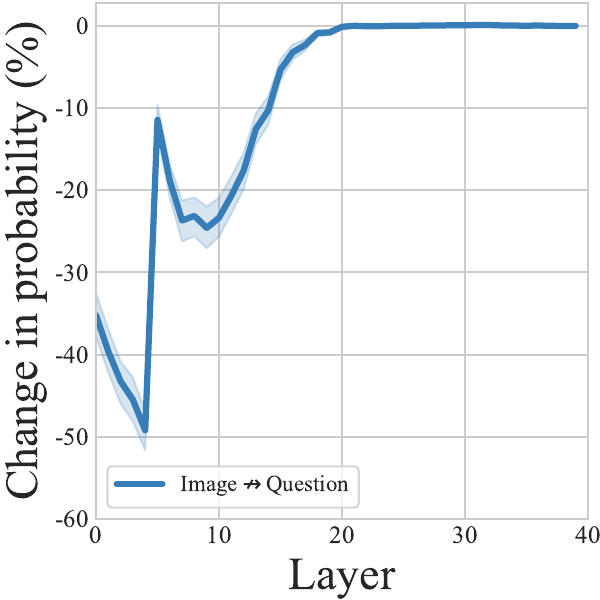}
\includegraphics[width=0.45\linewidth]{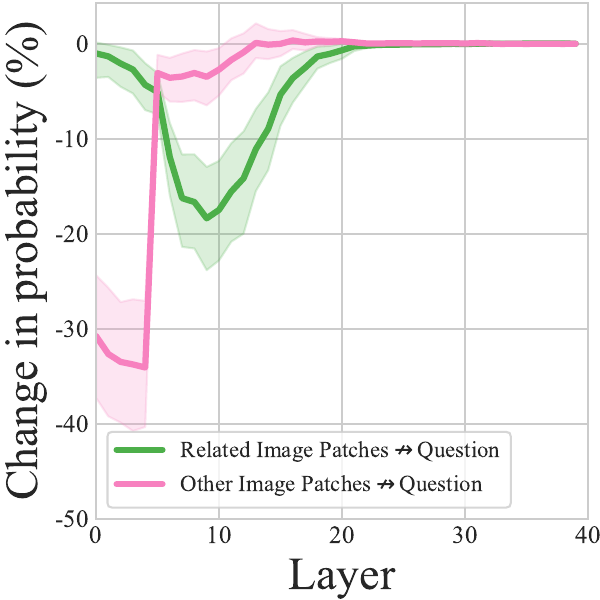}
\includegraphics[width=0.45\linewidth]{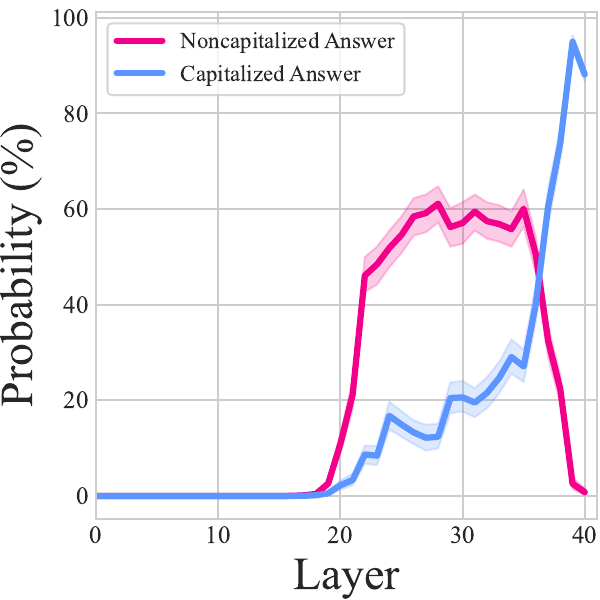}
\caption{The information flow and probability of answer word tracking on AOKVQA dataset.}
\label{fig:AOKVQA task}
\end{figure}

\begin{figure}[t]
\centering
\includegraphics[width=0.45\linewidth]{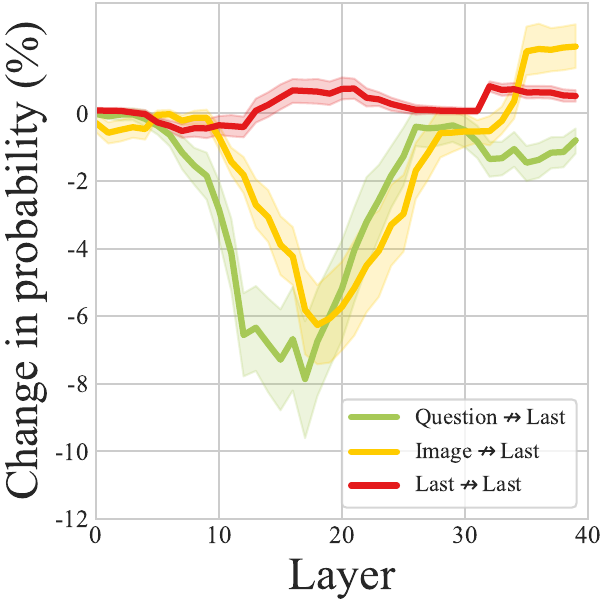}
\includegraphics[width=0.45\linewidth]{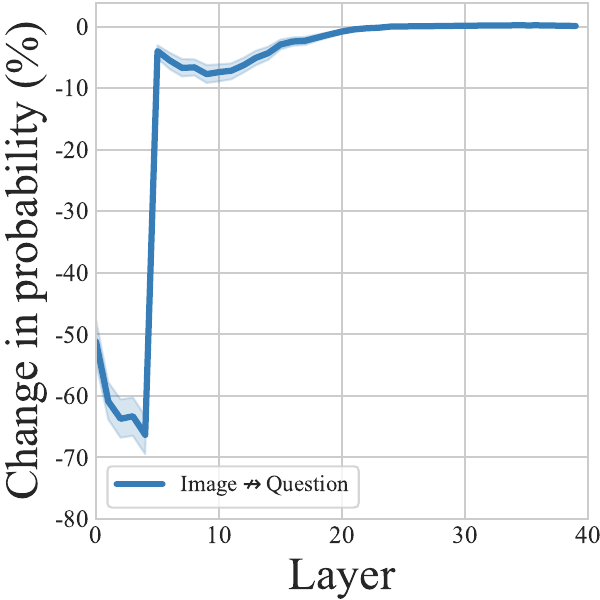}
\includegraphics[width=0.45\linewidth]{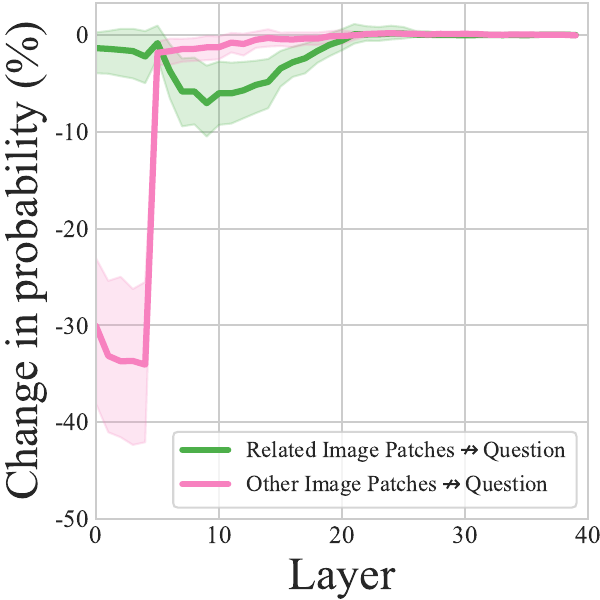}
\includegraphics[width=0.45\linewidth]{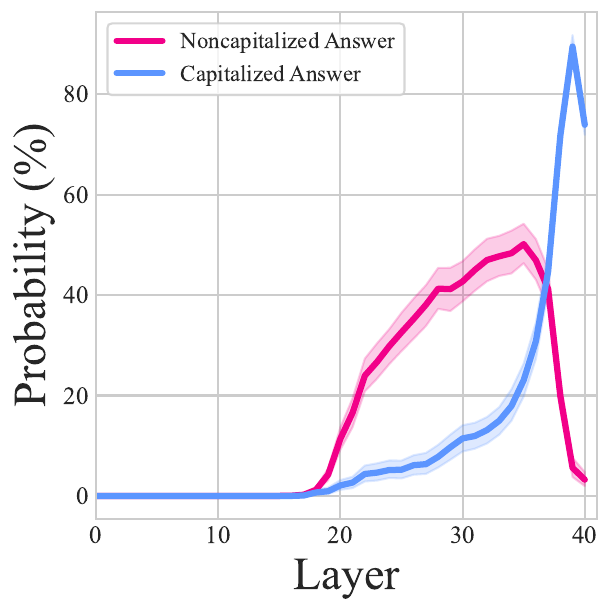}
\caption{The information flow and probability of answer word tracking on OKVQA dataset.}
\label{fig:OKVQA task}
\end{figure}

Since the main focus of our paper is the interaction between modalities, we evaluate tasks where cross-modal alignment plays a crucial role, while excluding those requiring external knowledge for reasoning to minimize confounding factors and ensure a more precise analysis. Nevertheless, to further validate our findings, we extend our experiments to two additional factual multimodal tasks, OKVQA\cite{marino_ok-vqa_2019} and AOKVQA\cite{schwenk_-okvqa_2022} involving more complex, fact-based reasoning requiring external commonsense and world knowledge. \cref{fig:OKVQA task,fig:AOKVQA task} shows the results align with our findings, i.e. from lower to higher layers: a two-stage multimodal integration, information flow from question to last positions and semantic generation and syntactic refinement.

\section{Experiments on other models}
\label{app: exp on other model}
We conduct the same experiments (six \textit{VQA} task types) as in the main body of the paper with the other three models. Six \textit{VQA} task types include (\textit{ChooseAttr}, \textit{ChooseCat}, \textit{ChooseRel}, \textit{LogicalObj}, \textit{QueryAttr} and \textit{CompareAttr}). The other three models include \textit{LLaVA-1.5-7b}, \textit{LLaVA-v1.6-Vicuna-7b} and \textit{Llama3-LLaVA-NEXT-8b}.

\begin{figure}[H]
\centering
\includegraphics[width=1\linewidth]{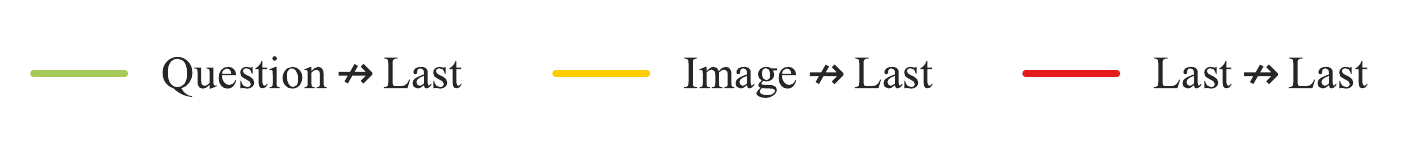}

\centering
\subfloat[ChooseAttr]{
    \includegraphics[width=0.32\linewidth]{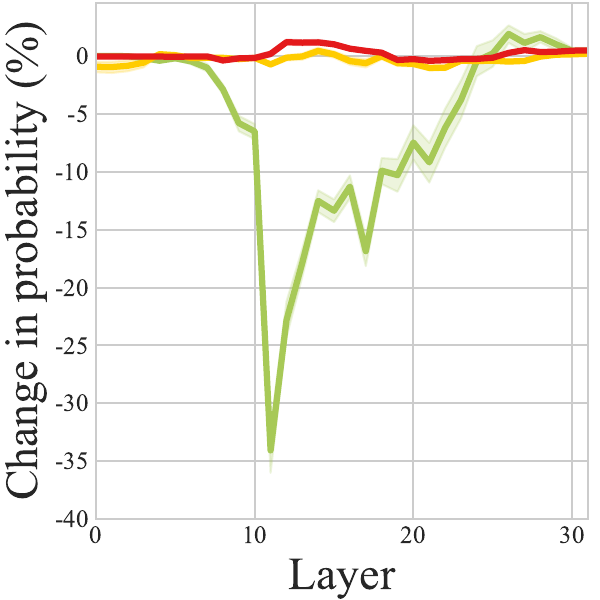}
}
\subfloat[ChooseCat]{
    \includegraphics[width=0.32\linewidth]{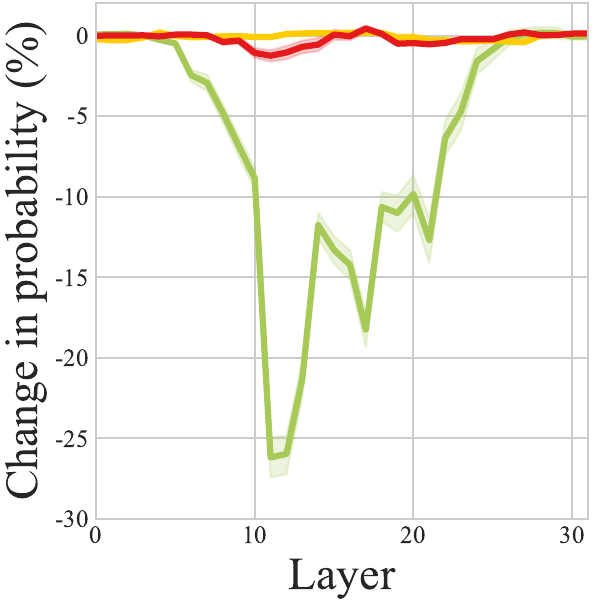}
}
\subfloat[ChooseRel]{
    \includegraphics[width=0.32\linewidth]{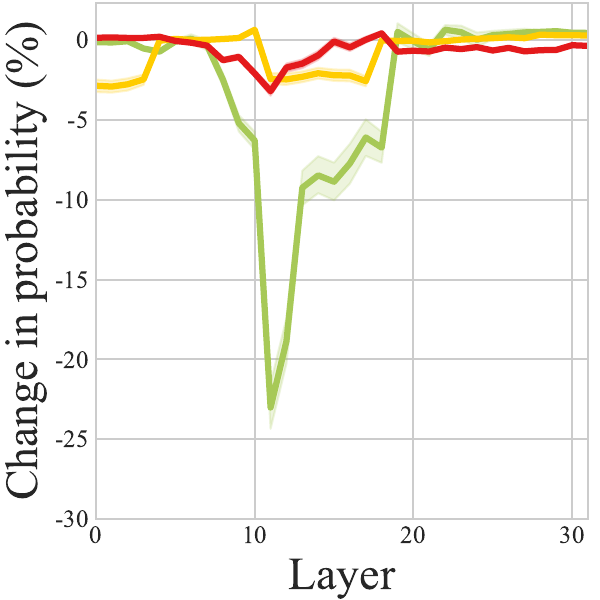}
}\\
\subfloat[CompareAttr]{
    \includegraphics[width=0.32\linewidth]{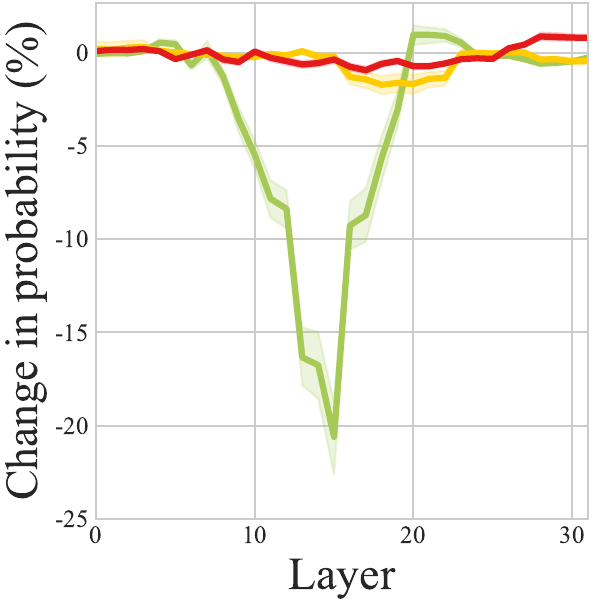}
}
\subfloat[LogicalObj]{
    \includegraphics[width=0.32\linewidth]{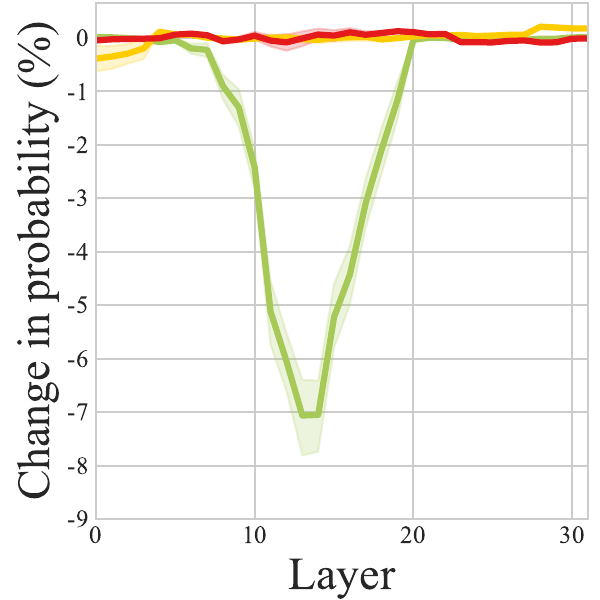}
}
\subfloat[QueryAttr]{
    \includegraphics[width=0.32\linewidth]{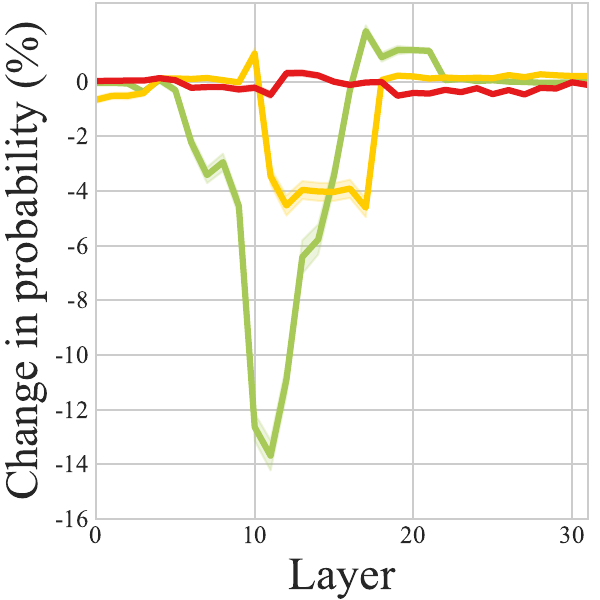}
}
\caption{The relative changes in prediction probability on \textit{LLaVA-1.5-7b} with six \textit{VQA} tasks. The \textit{Question}$\nrightarrow$ \textit{Last}, \textit{Image}$\nrightarrow$ \textit{Last} and  \textit{Last}$\nrightarrow$ \textit{Last} represent preventing \textit{last position} from attending to \textit{Question}, \textit{Image} and itself respectively.}
\label{fig:tolast_LLaVA-1.5-7b}
\end{figure}

\begin{figure}[t]
\centering
\includegraphics[width=0.33\linewidth]{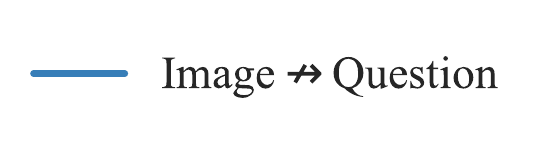}
\centering
\subfloat[ChooseAttr]{
    \includegraphics[width=0.32\linewidth]{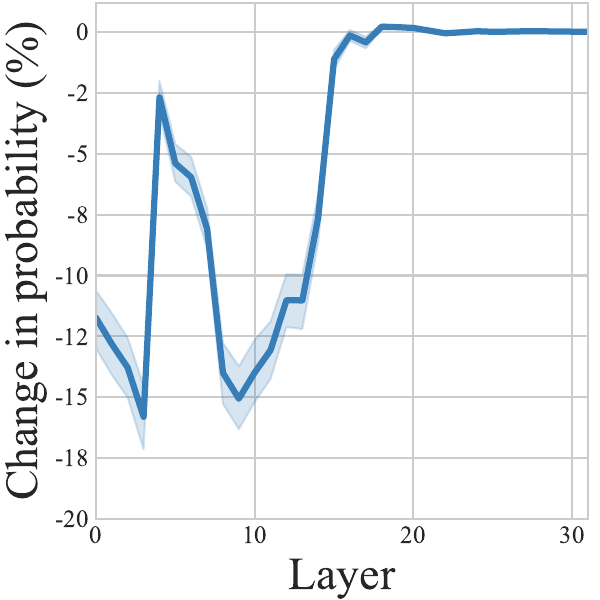}
}
\subfloat[ChooseCat]{
    \includegraphics[width=0.32\linewidth]{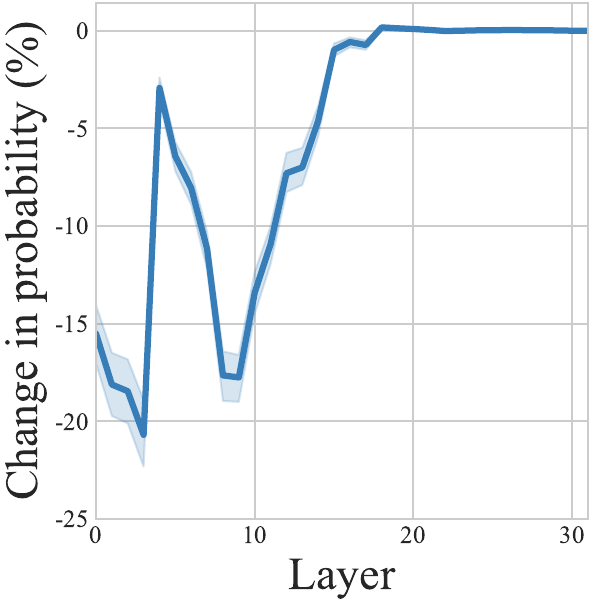}
}
\subfloat[ChooseRel]{
    \includegraphics[width=0.32\linewidth]{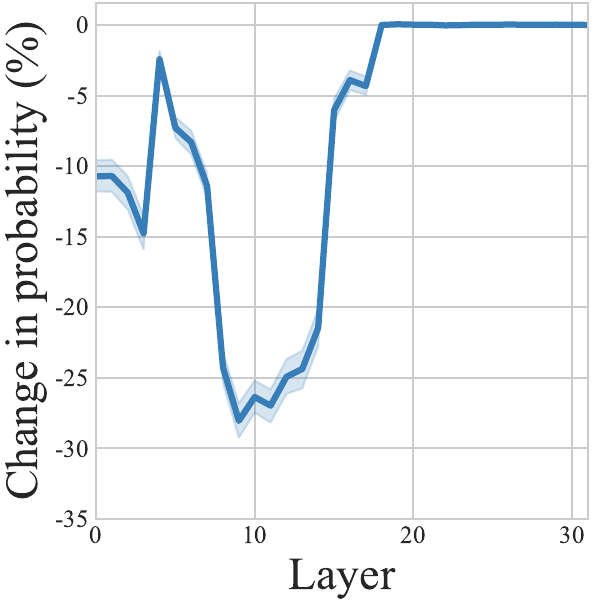}
}\\
\subfloat[CompareAttr]{
    \includegraphics[width=0.32\linewidth]{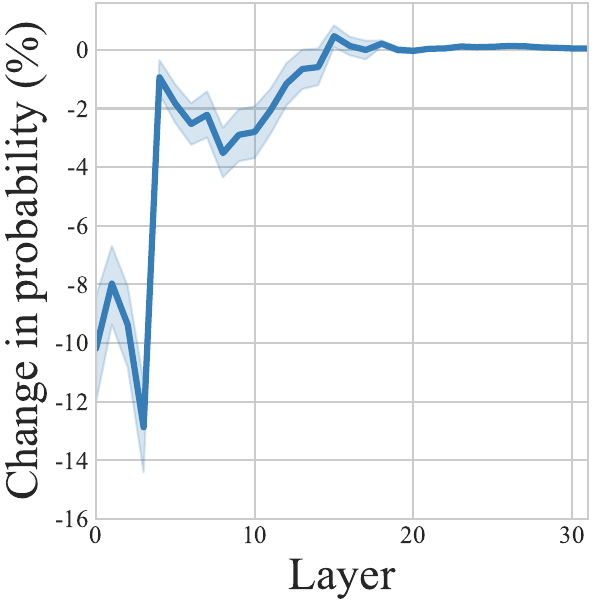}
}
\subfloat[LogicalObj]{
    \includegraphics[width=0.32\linewidth]{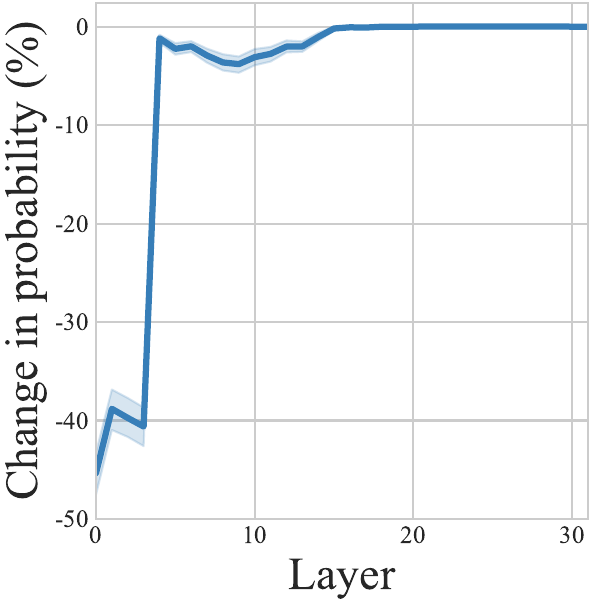}
}
\subfloat[QueryAttr]{
    \includegraphics[width=0.32\linewidth]{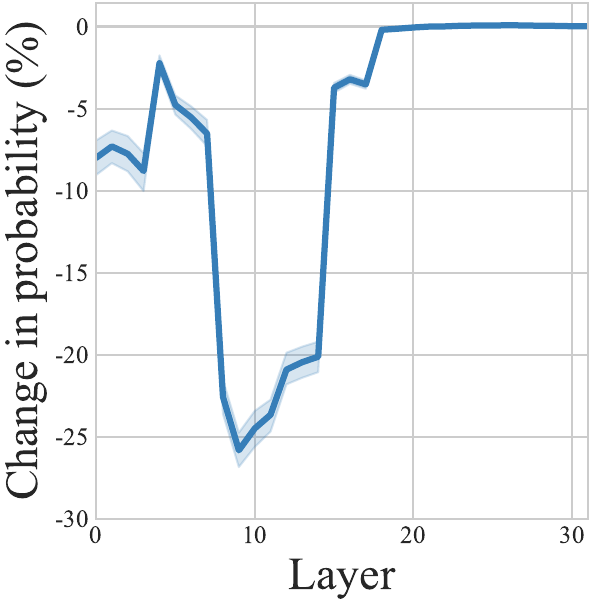}
}
\caption{
The relative changes in prediction probability when blocking attention edges from the \textit{question} positions to the \textit{image} positions on \textit{LLaVA-1.5-7b} with six \textit{VQA} tasks.
}
\label{fig:imageToquestion_LLaVA-1.5-7b}
\end{figure}

\begin{figure}[t]
\centering
\includegraphics[width=1.0\linewidth]{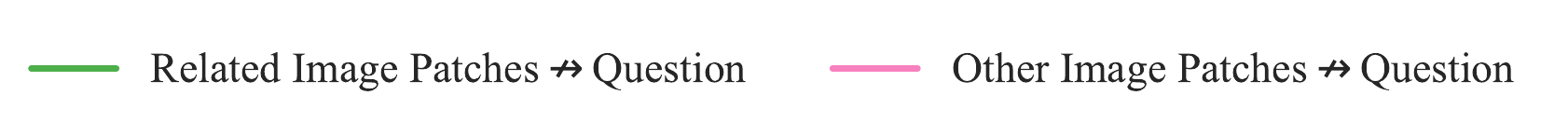}
\\
\centering
\subfloat[ChooseAttr]{
    \includegraphics[width=0.32\linewidth]{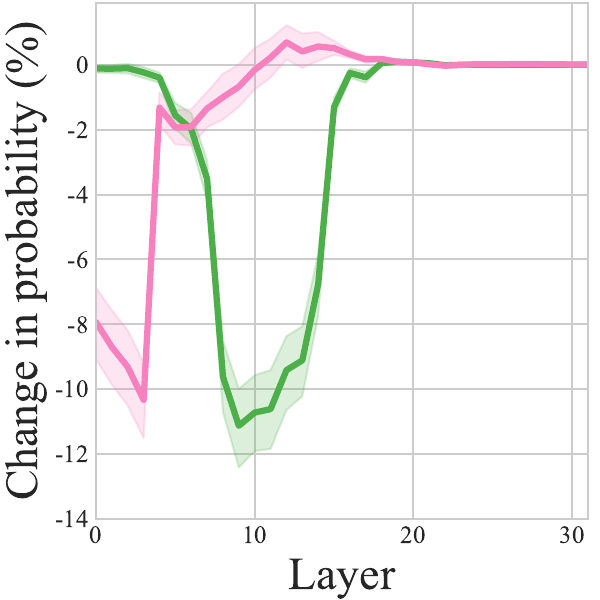}
}
\subfloat[ChooseCat]{
    \includegraphics[width=0.32\linewidth]{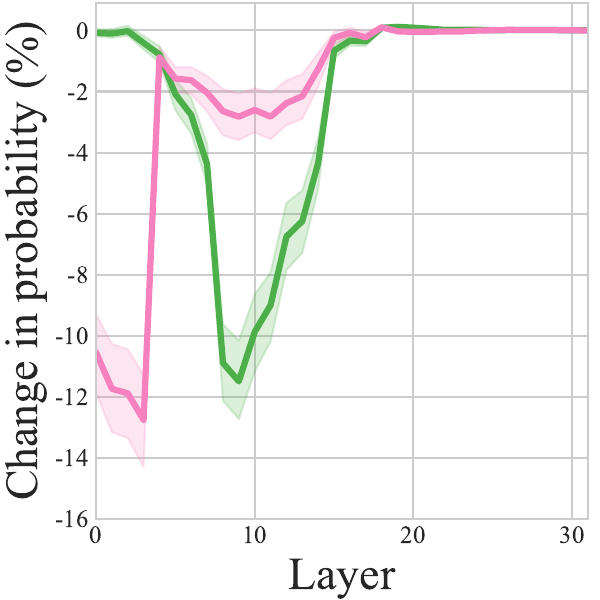}
}
\subfloat[ChooseRel]{
    \includegraphics[width=0.32\linewidth]{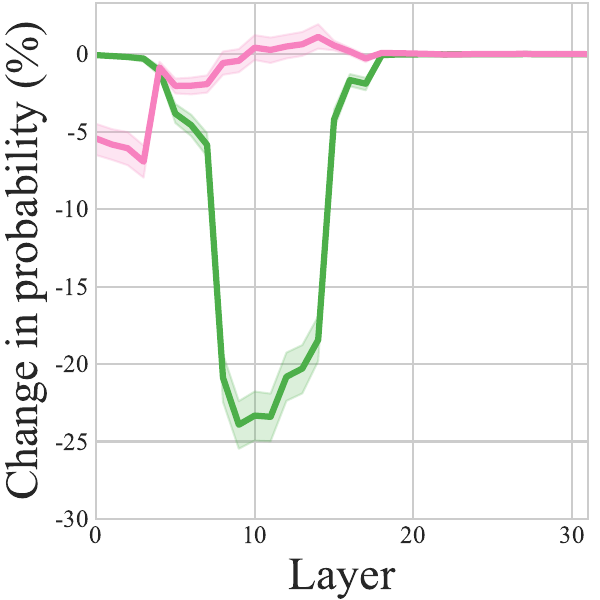}
}\\
\subfloat[CompareAttr]{
    \includegraphics[width=0.32\linewidth]{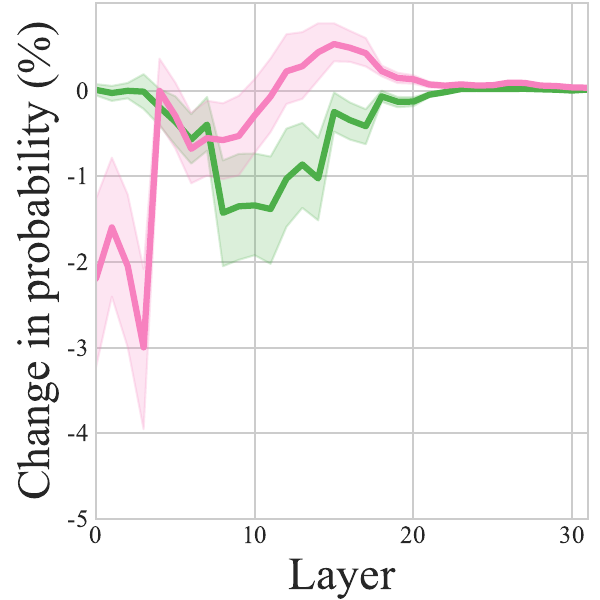}
}
\subfloat[LogicalObj]{
    \includegraphics[width=0.32\linewidth]{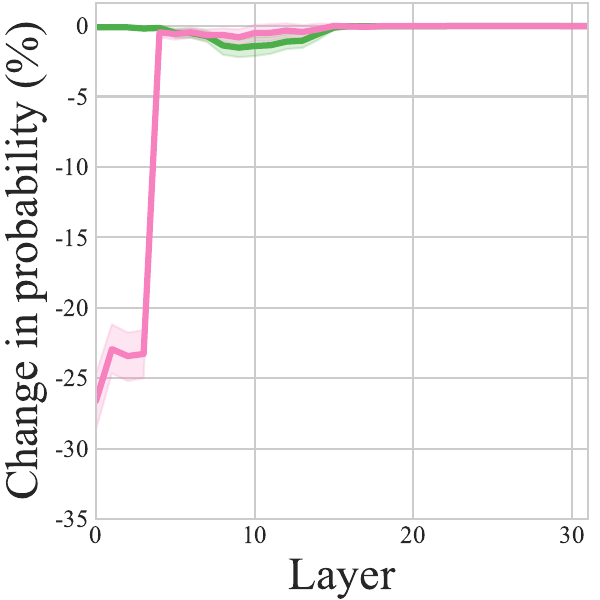}
}
\subfloat[QueryAttr]{
    \includegraphics[width=0.32\linewidth]{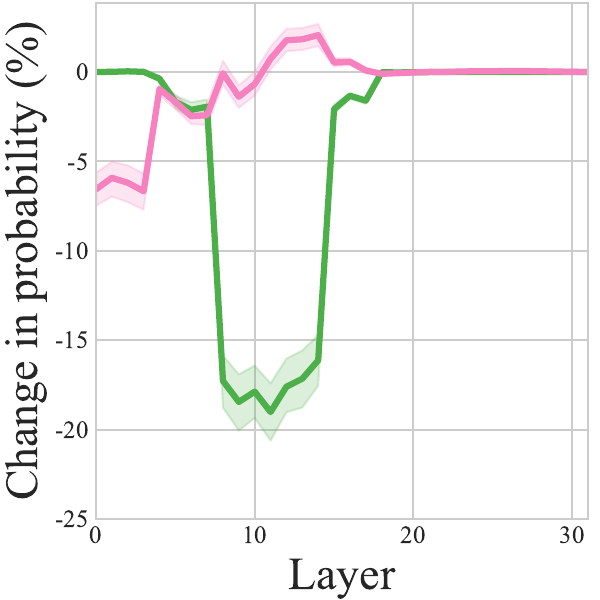}
}
\caption{The relative changes in prediction probability on \textit{LLaVA-1.5-7b} with six \textit{VQA} tasks.
\textit{Related Image Patches}$\nrightarrow$\textit{question} and \textit{Other Image Patches}$\nrightarrow$\textit{question} represent blocking the position of \textit{question} from attending to that of different image patches, region of interest and remainder, respectively.}
\label{fig:imagepatchesToquestion_LLaVA-1.5-7b}
\end{figure}

\subsection{\textit{LLaVA-1.5-7b}}
\begin{figure}[t]
\centering
\includegraphics[width=0.8\linewidth]{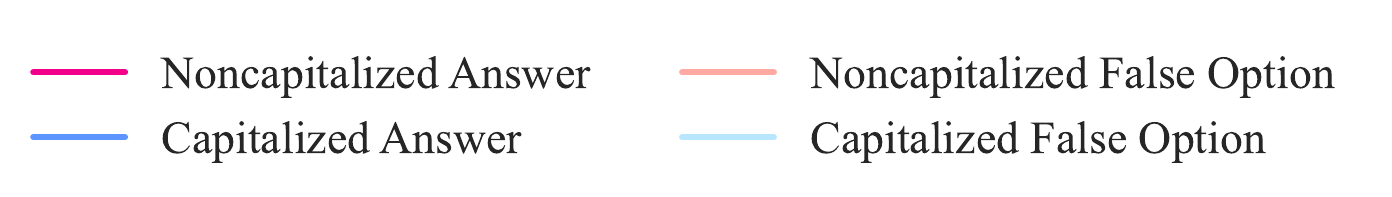}

\centering
\subfloat[ChooseAttr]{
    \includegraphics[width=0.32\linewidth]{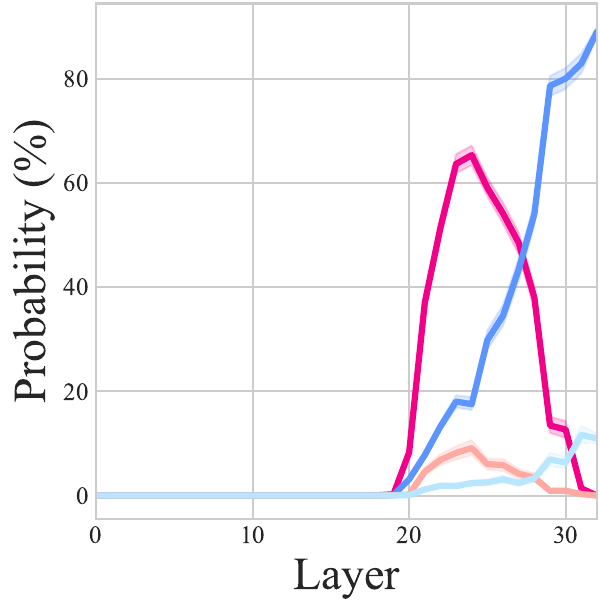}
}
\subfloat[ChooseCat]{
    \includegraphics[width=0.32\linewidth]{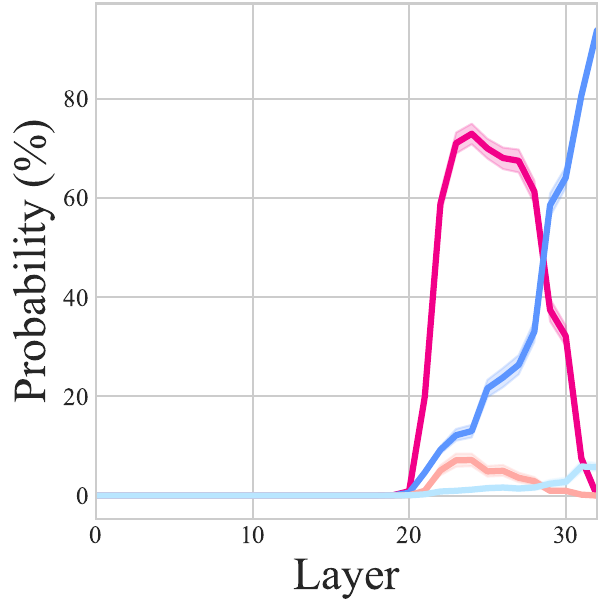}
}
\subfloat[ChooseRel]{
    \includegraphics[width=0.32\linewidth]{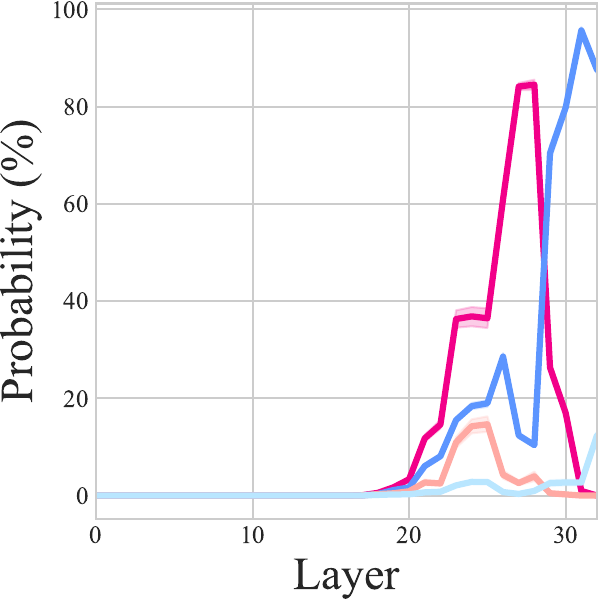}
}\\
\subfloat[CompareAttr]{
    \includegraphics[width=0.32\linewidth]{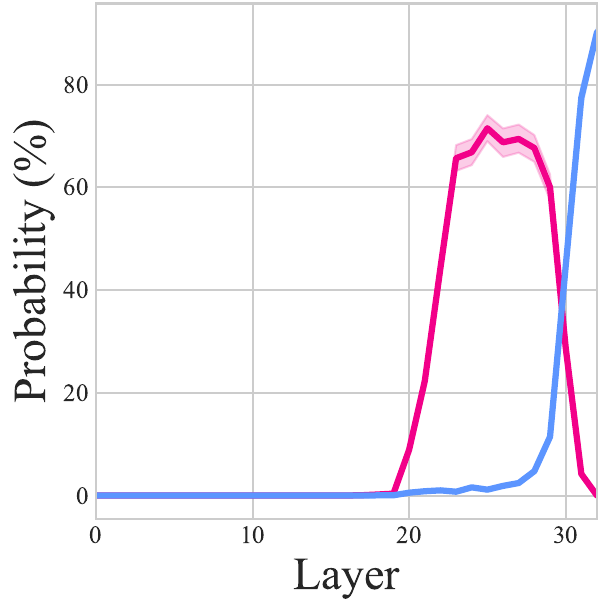}
}
\subfloat[LogicalObj]{
    \includegraphics[width=0.32\linewidth]{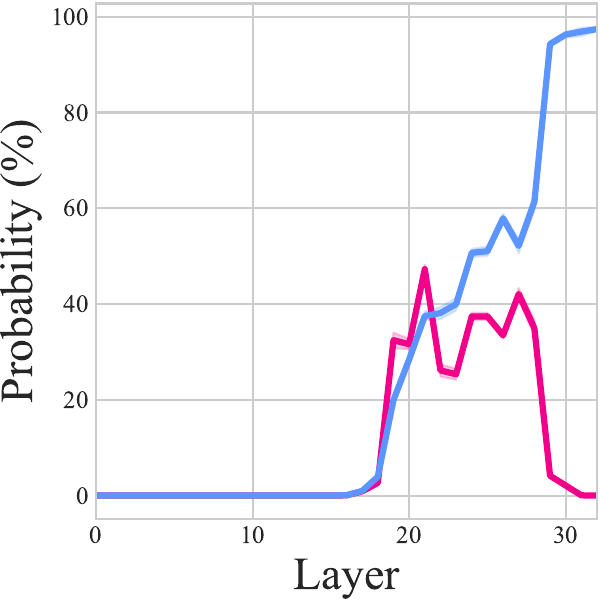}
}
\subfloat[QueryAttr]{
    \includegraphics[width=0.32\linewidth]{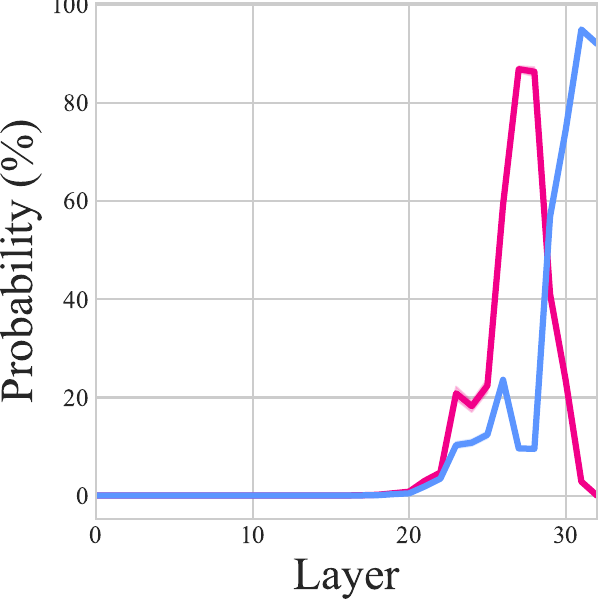}
}
\caption{The probability of the answer word at the \textit{last position} across all layers in LLaVA-1.5-7b with six VQA tasks. \textit{Capitalized Answer} and \textit{Noncapitalized Answer} represent the answer word with or without the uppercase of the initial letter, respectively. As the tasks of \textit{ChooseAttr}, \textit{ChooseCat} and \textit{ChooseRel} contain \textit{false option}, we also provide the probability of it.}
\label{fig:last_position_prob_LLaVA-1.5-7b}
\end{figure}

\textit{LLaVA-1.5-7b} is a small version of \textit{LLaVA-1.5-13b} presented in the main body of the paper. It contains 32 transformer blocks (layers) instead of 40 layers in \textit{LLaVA-1.5-13b}. The information flow from different parts of the input sequence (\textit{image} and \textit{question}) to \textit{last position}, from \textit{image} to \textit{question} and from different image patches (\textit{related image patches} and \textit{other image patches}) to \textit{question}, as shown in \Cref{fig:tolast_LLaVA-1.5-7b}, \Cref{fig:imageToquestion_LLaVA-1.5-7b} and \Cref{fig:imagepatchesToquestion_LLaVA-1.5-7b} respectively, are consistent with the observations for the \textit{LLaVA-1.5-13b} model, as shown in \Cref{fig:tolast_13b}, \Cref{fig:imageToquestion_13b} and \Cref{fig:imagepatchesToquestion_13b} respectively, in the main body of the paper. Specifically, the model first propagates critical information twice from the \textit{image} positions to the \textit{question} positions in the lower-to-middle layers of the MLLM. For the twice multimodal information integration, the first one focuses on producing the generative representations over the whole image while the second one tends to construct question-related representation. Subsequently, in the middle layers, the critical multimodal information flows from the \textit{question} positions to the \textit{last position} for the final prediction. The difference between the two models is the magnitude of reduction in the probability when blocking the attention edge between \textit{image} and \textit{question}. In model \textit{LLaVA-1.5-7b}, the first drop is rather smaller than that in model \textit{LLaVA-1.5-13b}. However, this does not conflict with our conclusion that the information flows from \textit{image} to \textit{question} twice and one after the other in the main body of the paper. Moreover, the probability change of the answer word across all layers as shown in \Cref{fig:last_position_prob_LLaVA-1.5-7b} is also consistent with the result in \Cref{fig:last_position_prob_13b} in the main body of the paper. Specifically, the model first generates the answer semantically in the middle layers and then refines the syntactic correctness of the answer in the higher layers.

\subsection{\textit{LLaVA-v1.6-Vicuna-7b}}

\begin{figure}[t]
\centering
\includegraphics[width=1\linewidth]{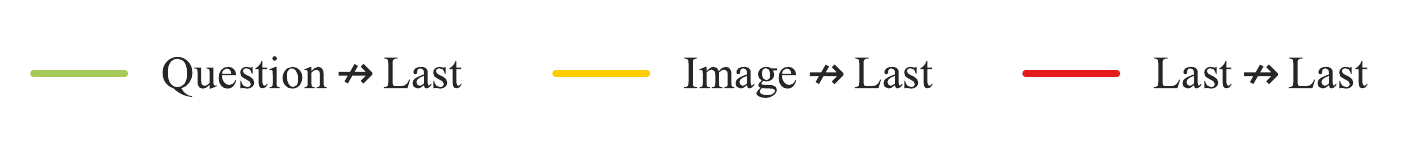}

\centering
\subfloat[ChooseAttr]{
    \includegraphics[width=0.32\linewidth]{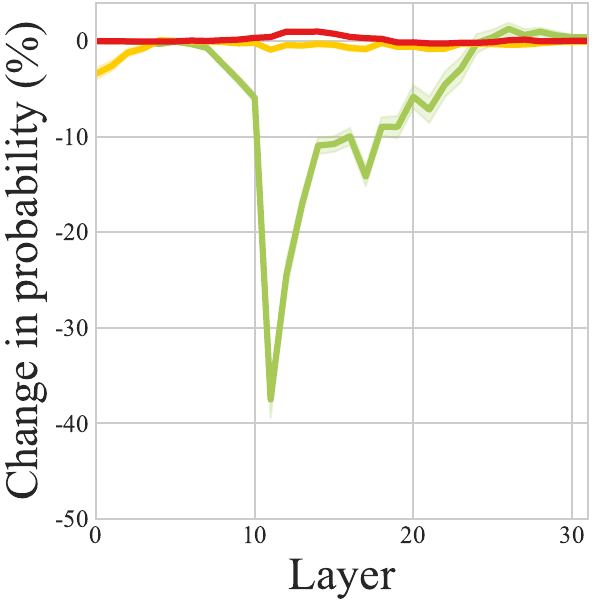}
}
\subfloat[ChooseCat]{
    \includegraphics[width=0.32\linewidth]{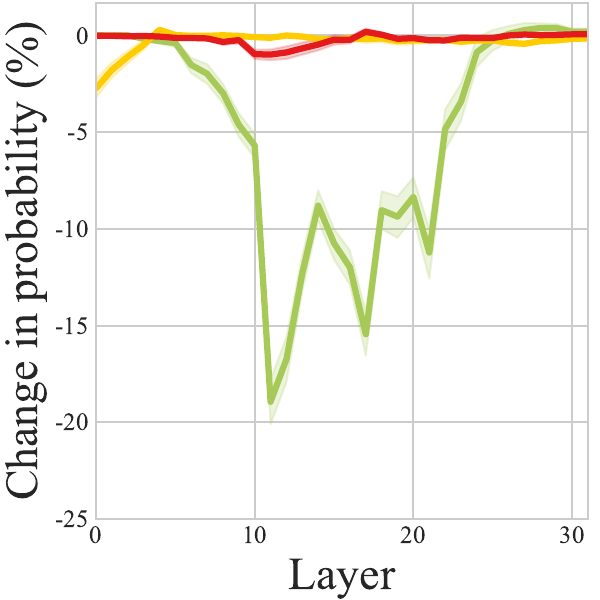}
}
\subfloat[ChooseRel]{
    \includegraphics[width=0.32\linewidth]{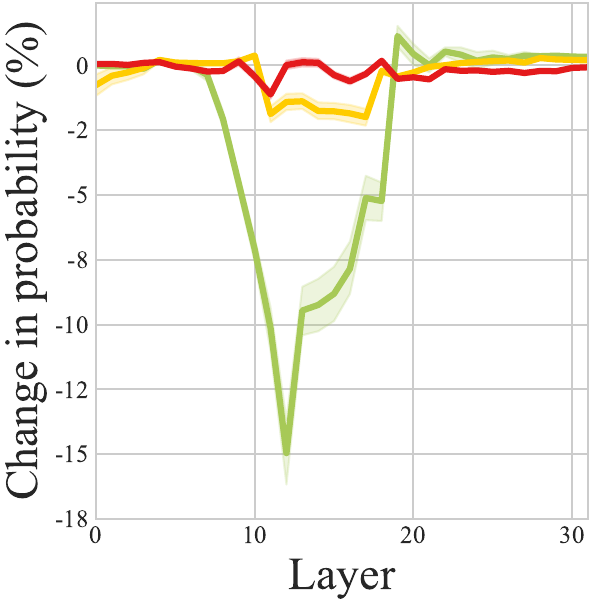}
}\\
\subfloat[CompareAttr]{
    \includegraphics[width=0.32\linewidth]{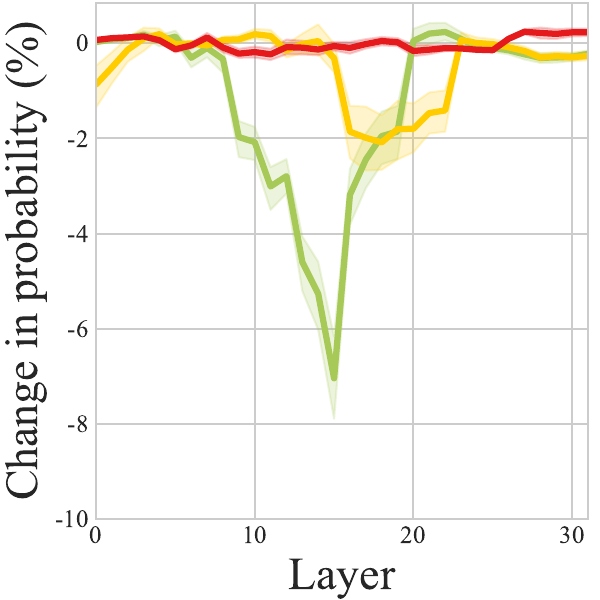}
}
\subfloat[LogicalObj]{
    \includegraphics[width=0.32\linewidth]{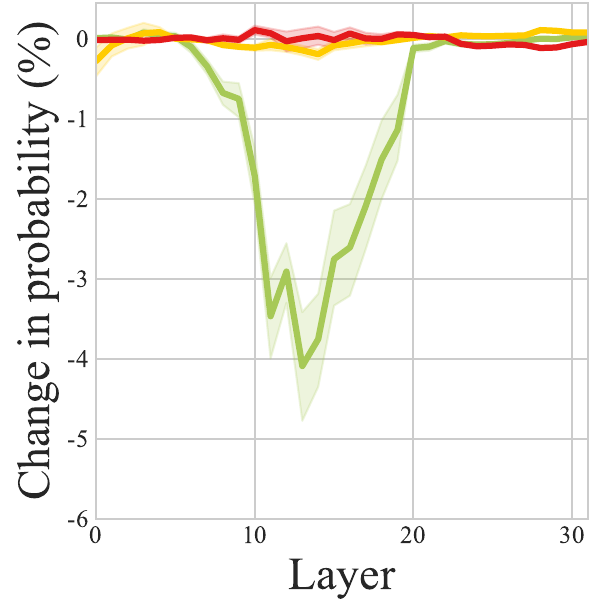}
}
\subfloat[QueryAttr]{
    \includegraphics[width=0.32\linewidth]{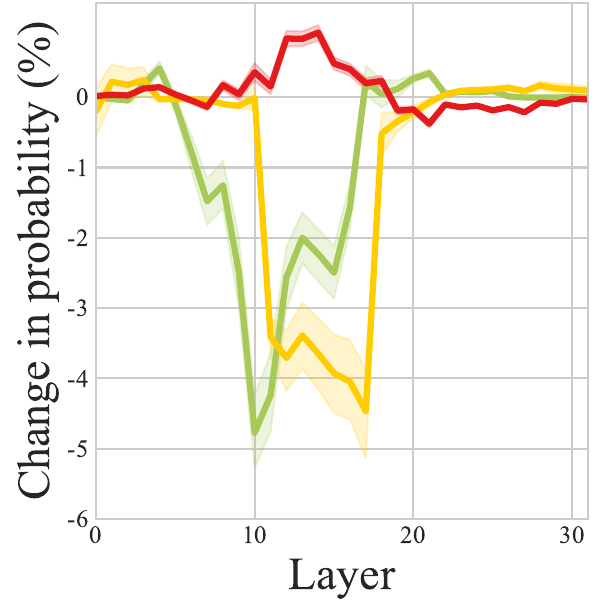}
}
\caption{The relative changes in prediction probability on \textit{LLaVA-v1.6-Vicuna-7b} with six \textit{VQA} tasks. The \textit{Question}$\nrightarrow$ \textit{Last}, \textit{Image}$\nrightarrow$ \textit{Last} and  \textit{Last}$\nrightarrow$ \textit{Last} represent preventing \textit{last position} from attending to \textit{Question}, \textit{Image} and itself respectively.}
\label{fig:tolast_LLaVA-v1.6-Vicuna-7b}
\end{figure}

\begin{figure}[t]
\centering
\includegraphics[width=0.33\linewidth]{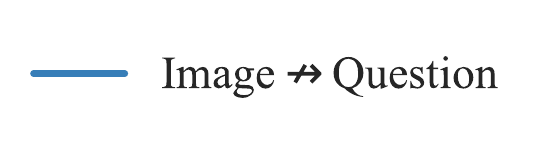}
\centering
\subfloat[ChooseAttr]{
    \includegraphics[width=0.32\linewidth]{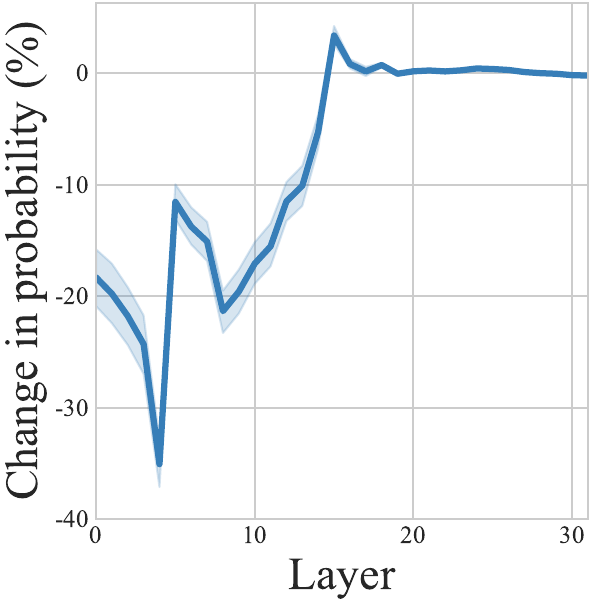}
}
\subfloat[ChooseCat]{
    \includegraphics[width=0.32\linewidth]{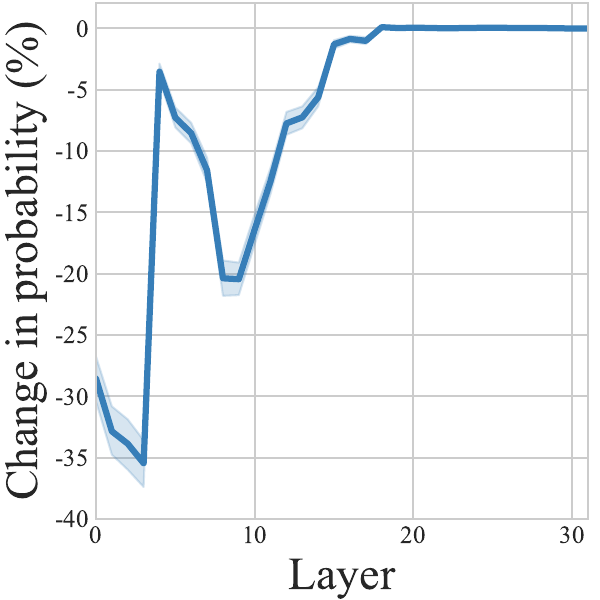}
}
\subfloat[ChooseRel]{
    \includegraphics[width=0.32\linewidth]{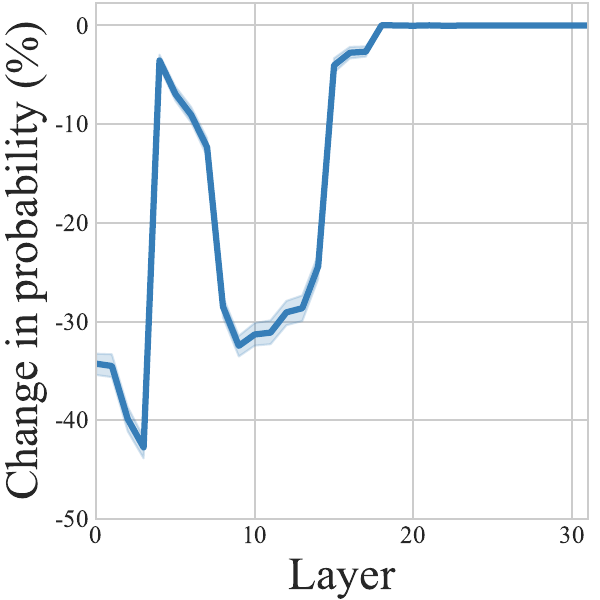}
}\\
\subfloat[CompareAttr]{
    \includegraphics[width=0.32\linewidth]{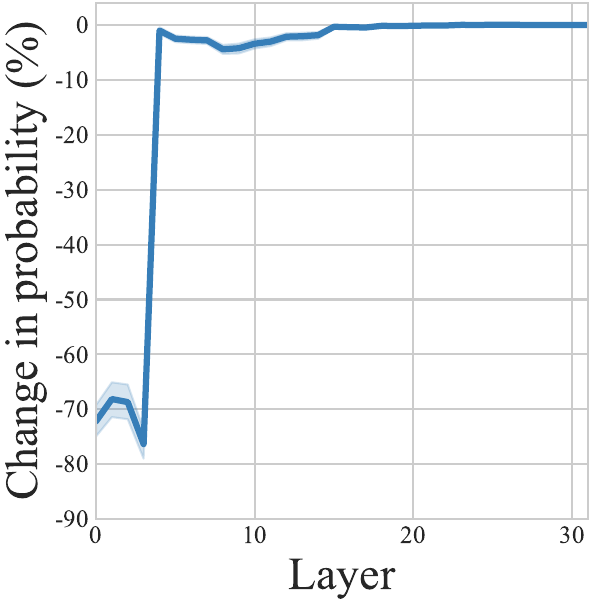}
}
\subfloat[LogicalObj]{
    \includegraphics[width=0.32\linewidth]{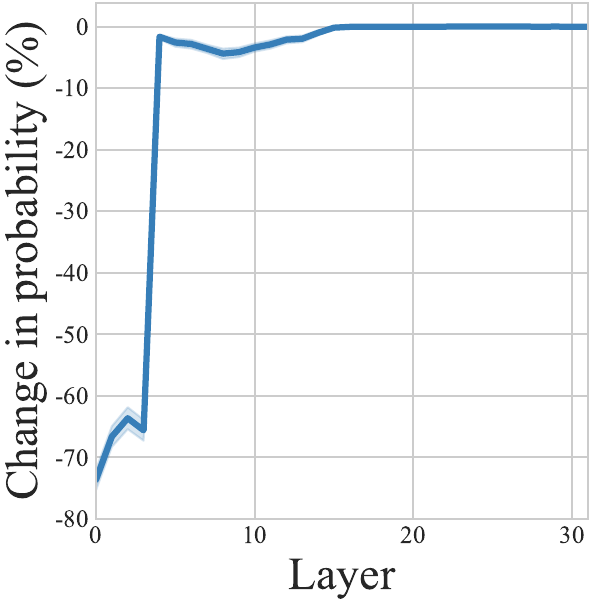}
}
\subfloat[QueryAttr]{
    \includegraphics[width=0.32\linewidth]{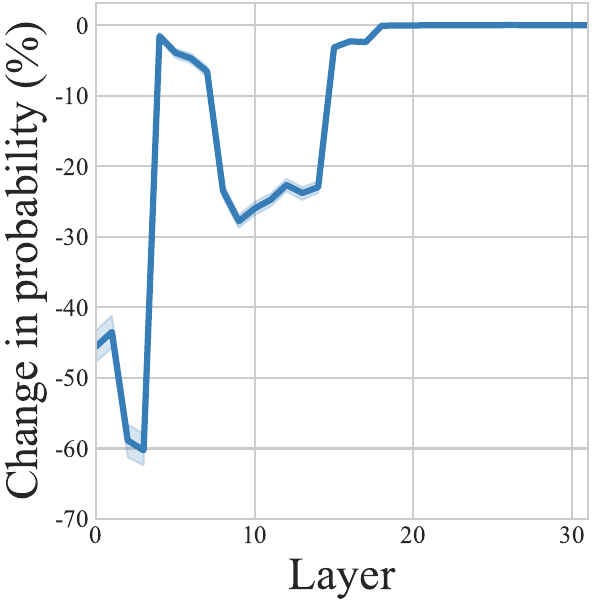}
}
\caption{
The relative changes in prediction probability when blocking attention edges from the \textit{question} positions to the \textit{image} positions on \textit{LLaVA-v1.6-Vicuna-7b} with six \textit{VQA} tasks.
}
\label{fig:imageToquestion_LLaVA-v1.6-Vicuna-7b}
\end{figure}

\textit{LLaVA-v1.6-Vicuna-7b} has the similar architecture with \textit{LLaVA-1.5-13b} in the main body of the paper. The difference between them includes the layer number and the the way processing image patch features. The \textit{LLaVA-v1.6-Vicuna-7b} has 32 layers versus 40 layers in \textit{LLaVA-1.5-13b}. \textit{LLaVA-1.5-13b} directly feeds the original fixed-length image patch features from the image encoder into the LLM as input tokens. In contrast, \textit{LLaVA-v1.6-Vicuna-7b} employs a dynamic high-resolution technique, which dynamically adjusts image resolution, resulting in variable-length image patch features with higher resolution.  Specifically, the higher resolution is implemented by splitting the image into grids and encoding them independently. 

\begin{figure}[t]
\centering
\includegraphics[width=1.0\linewidth]{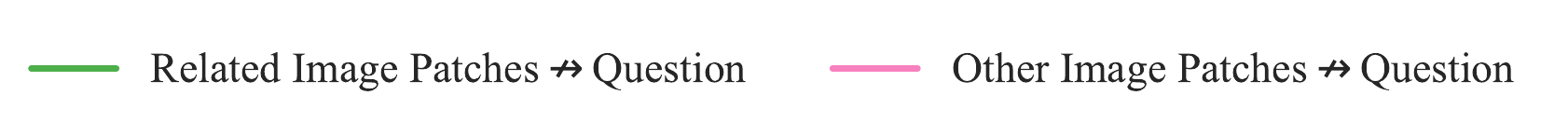}
\\
\centering
\subfloat[ChooseAttr]{
    \includegraphics[width=0.32\linewidth]{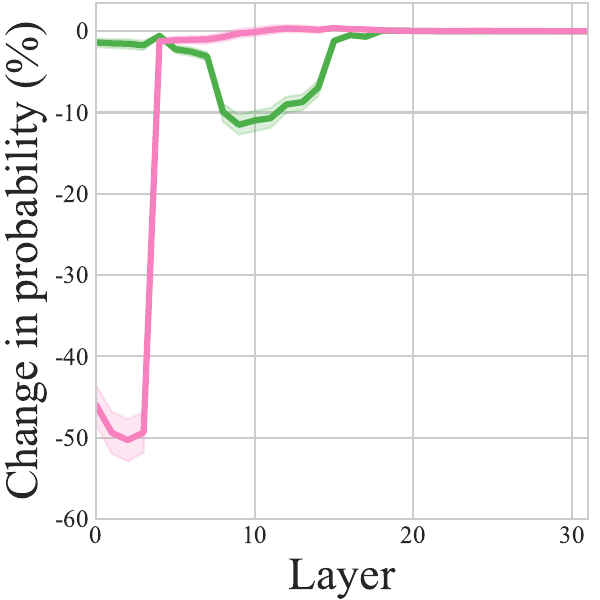}
}
\subfloat[ChooseCat]{
    \includegraphics[width=0.32\linewidth]{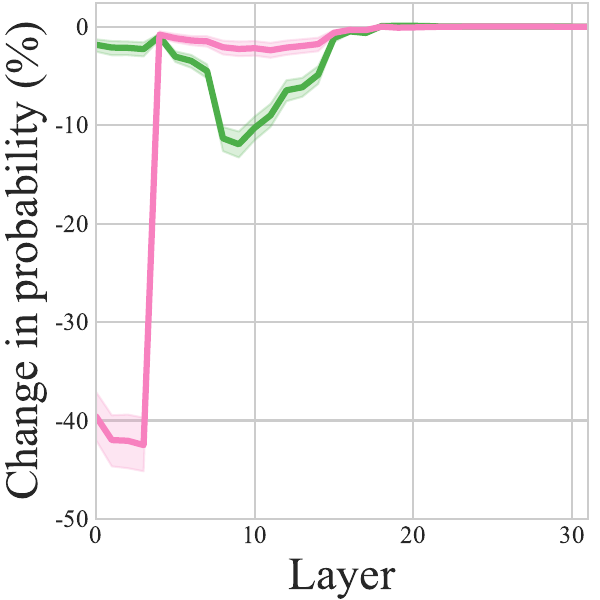}
}
\subfloat[ChooseRel]{
    \includegraphics[width=0.32\linewidth]{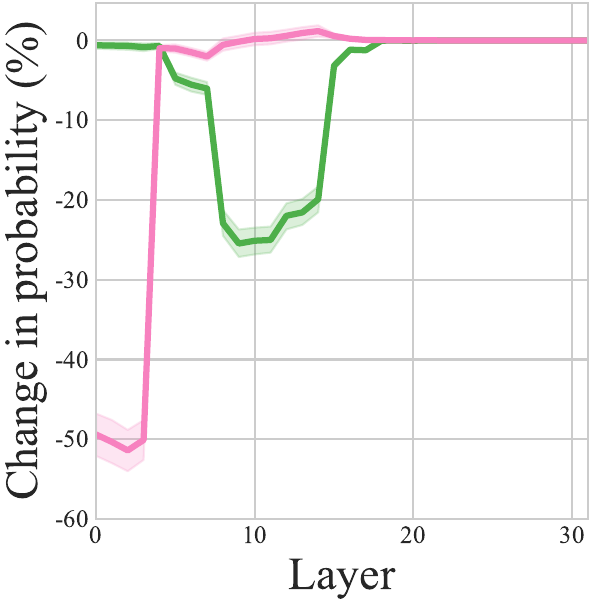}
}\\
\subfloat[CompareAttr]{
    \includegraphics[width=0.32\linewidth]{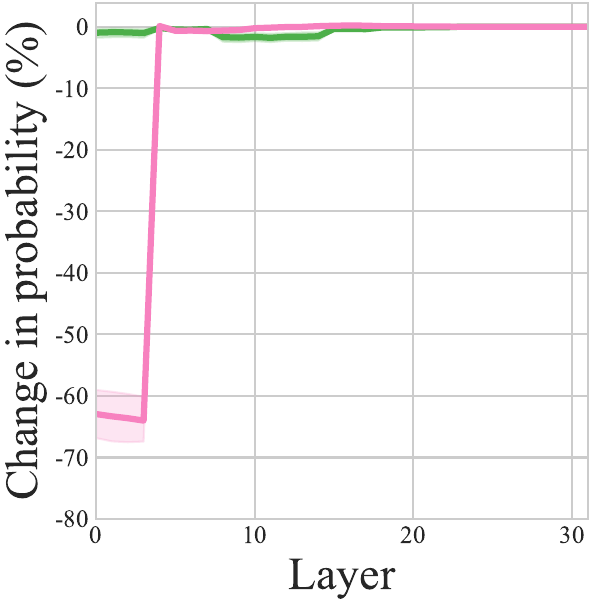}
}
\subfloat[LogicalObj]{
    \includegraphics[width=0.32\linewidth]{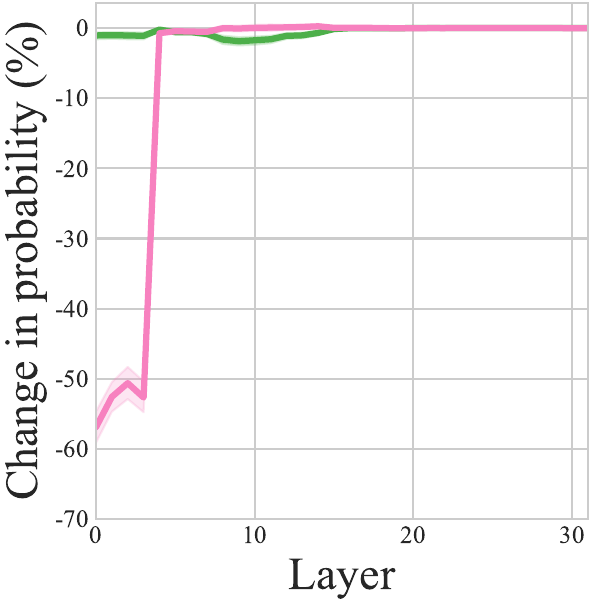}
}
\subfloat[QueryAttr]{
    \includegraphics[width=0.32\linewidth]{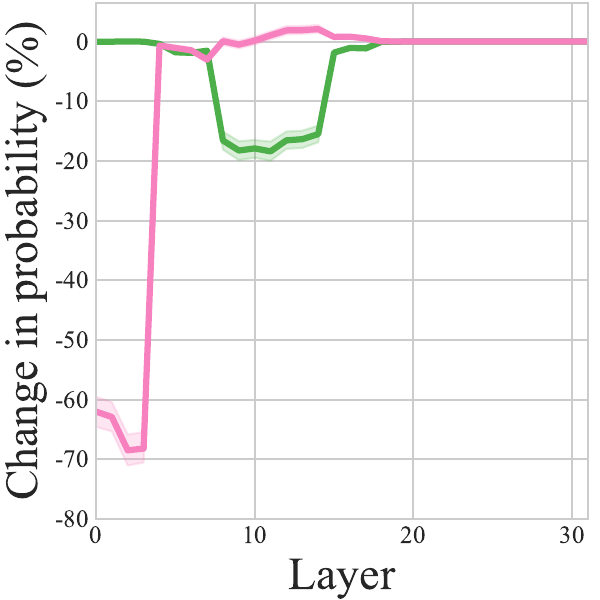}
}
\caption{The relative changes in prediction probability on \textit{LLaVA-v1.6-Vicuna-7b} with six \textit{VQA} tasks.
\textit{Related Image Patches}$\nrightarrow$\textit{question} and \textit{Other Image Patches}$\nrightarrow$\textit{question} represent blocking the position of \textit{question} from attending to that of different image patches, region of interest and remainder, respectively.}
\label{fig:imagepatchesToquestion_LLaVA-v1.6-Vicuna-7b}
\end{figure}

\begin{figure}[t]
\centering
\includegraphics[width=0.8\linewidth]{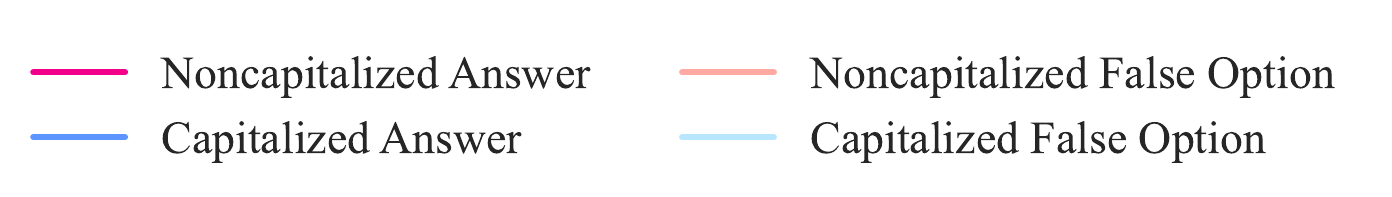}

\centering
\subfloat[ChooseAttr]{
    \includegraphics[width=0.32\linewidth]{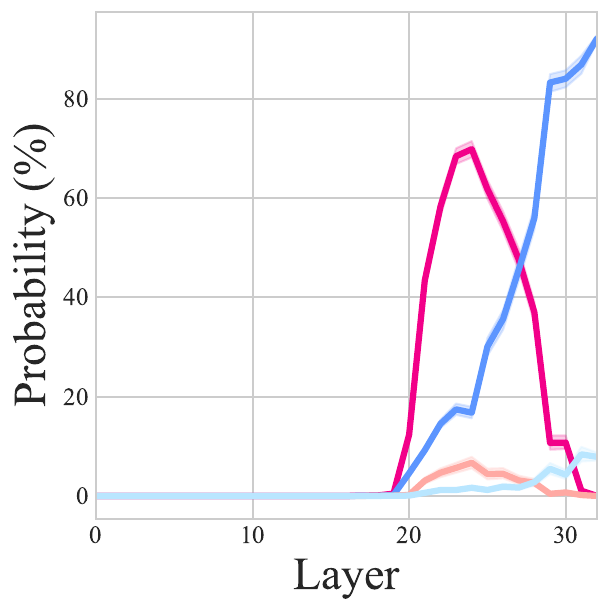}
}
\subfloat[ChooseCat]{
    \includegraphics[width=0.32\linewidth]{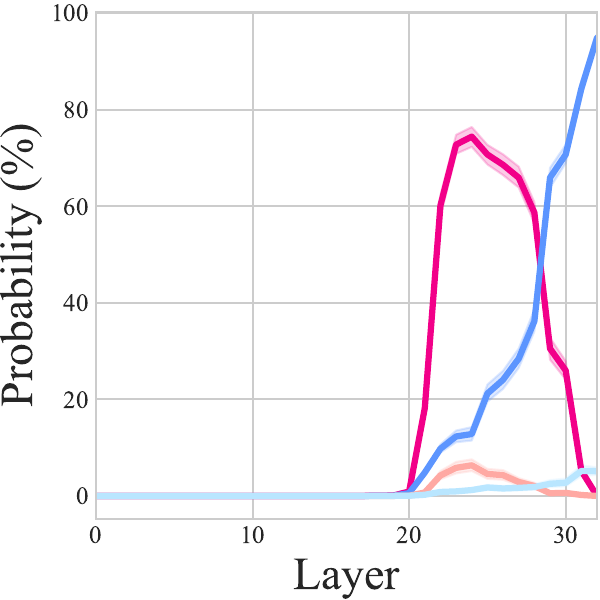}
}
\subfloat[ChooseRel]{
    \includegraphics[width=0.32\linewidth]{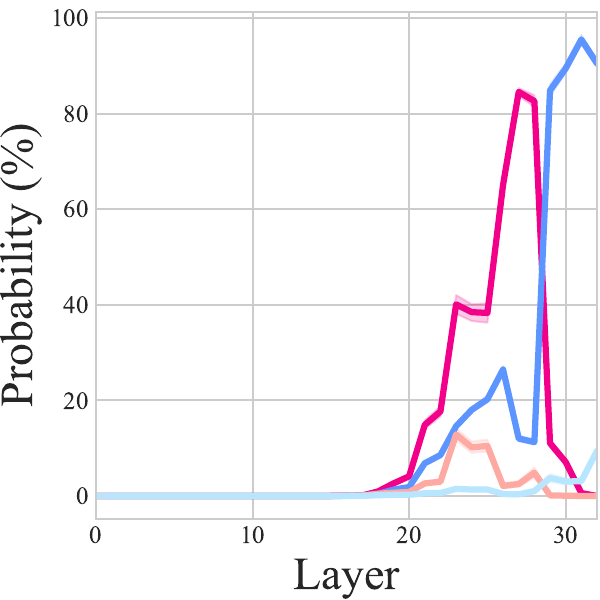}
}\\
\subfloat[CompareAttr]{
    \includegraphics[width=0.32\linewidth]{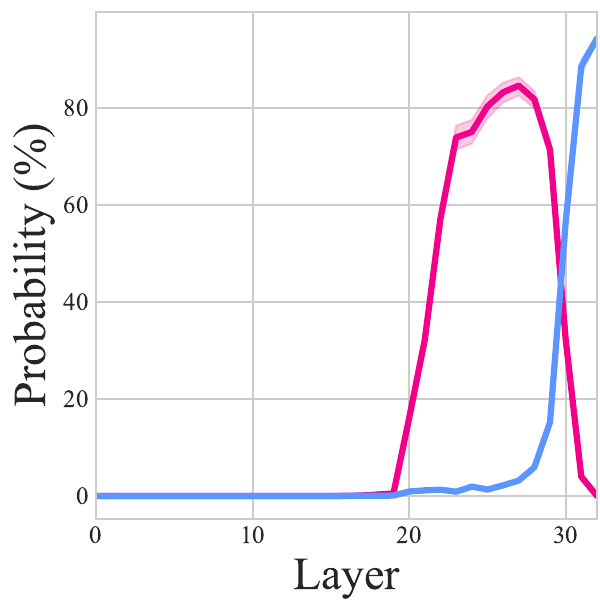}
}
\subfloat[LogicalObj]{
    \includegraphics[width=0.32\linewidth]{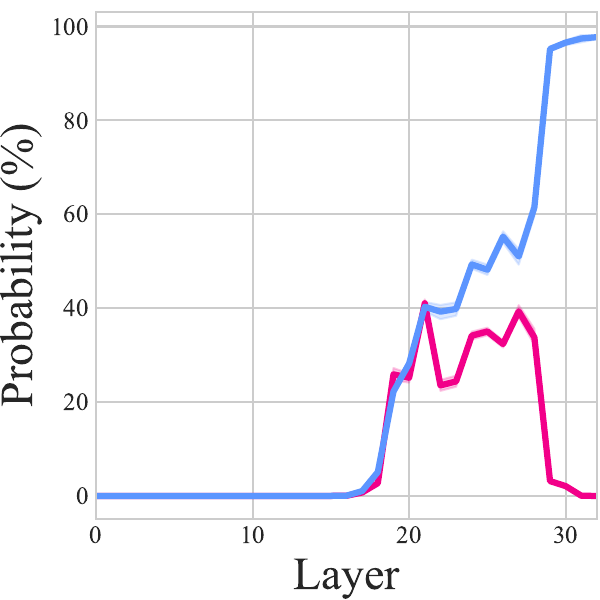}
}
\subfloat[QueryAttr]{
    \includegraphics[width=0.32\linewidth]{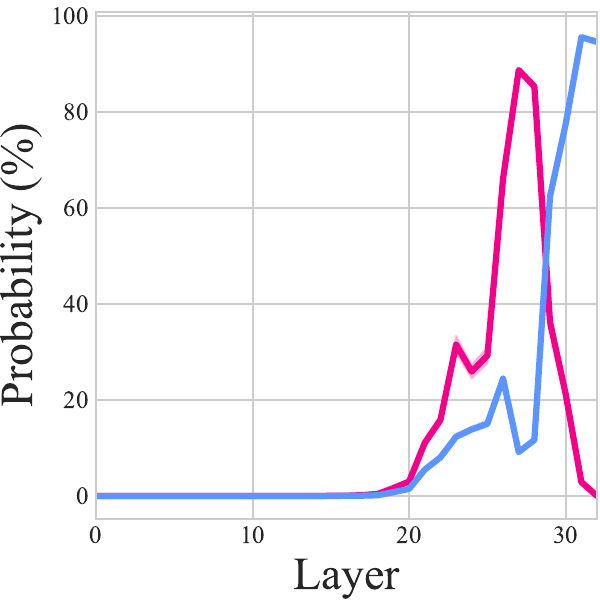}
}
\caption{The probability of the answer word at the \textit{last position} across all layers in LLaVA-v1.6-Vicuna-7b with six VQA tasks. \textit{Capitalized Answer} and \textit{Noncapitalized Answer} represent the answer word with or without the uppercase of the initial letter, respectively. As the tasks of \textit{ChooseAttr}, \textit{ChooseCat} and \textit{ChooseRel} contain \textit{false option}, we also provide the probability of it.}
\label{fig:last_position_prob_LLaVA-v1.6-Vicuna-7b}
\end{figure}

The information flow from different parts of the input sequence (\textit{image} and \textit{question}) to \textit{last position}, from \textit{image} to \textit{question} and from different image patches (\textit{related image patches} and \textit{other image patches}) to \textit{question}, as shown in \Cref{fig:tolast_LLaVA-v1.6-Vicuna-7b}, \Cref{fig:imageToquestion_LLaVA-v1.6-Vicuna-7b} and \Cref{fig:imagepatchesToquestion_LLaVA-v1.6-Vicuna-7b} respectively, are consistent with the observations for the \textit{LLaVA-1.5-13b} model, as shown in \Cref{fig:tolast_13b}, \Cref{fig:imageToquestion_13b} and \Cref{fig:imagepatchesToquestion_13b} respectively, in the main body of the paper. Specifically, the model first propagates critical information twice from the \textit{image} positions to the \textit{question} positions in the lower-to-middle layers of the MLLM. For the dual-stage multimodal information integration, the first stage emphasizes generating holistic representations of the entire image, while the second stage focuses on constructing representations that are specifically aligned with the given question.
Subsequently, in the middle layers, the critical multimodal information flows from the \textit{question} positions to the \textit{last position} for the final prediction. Moreover, the probability change of the answer word across all layers as shown in \Cref{fig:last_position_prob_LLaVA-v1.6-Vicuna-7b} is also consistent with the result in \Cref{fig:last_position_prob_13b} in the main body of the paper. Specifically, the model first generates the answer semantically in the middle layers and then refines the syntactic correctness of the answer in the higher layers.

\subsection{\textit{Llama3-LLaVA-NEXT-8b}}
\begin{figure}[t]
\centering
\includegraphics[width=1\linewidth]{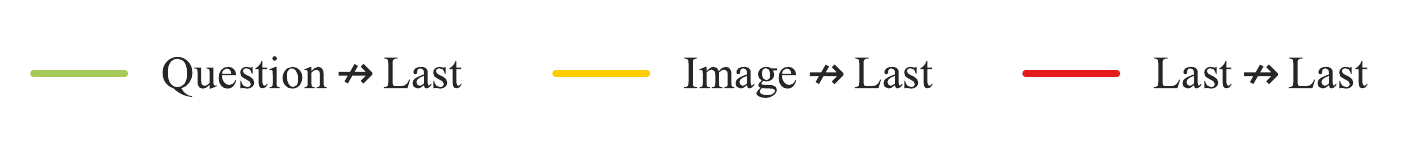}

\centering
\subfloat[ChooseAttr]{
    \includegraphics[width=0.32\linewidth]{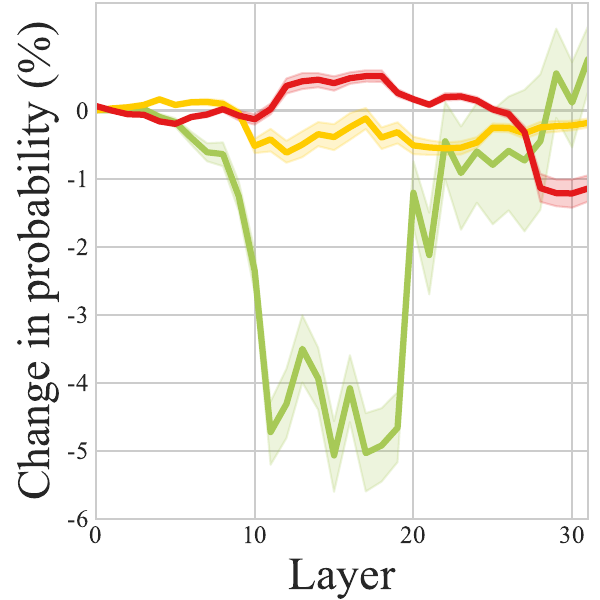}
}
\subfloat[ChooseCat]{
    \includegraphics[width=0.32\linewidth]{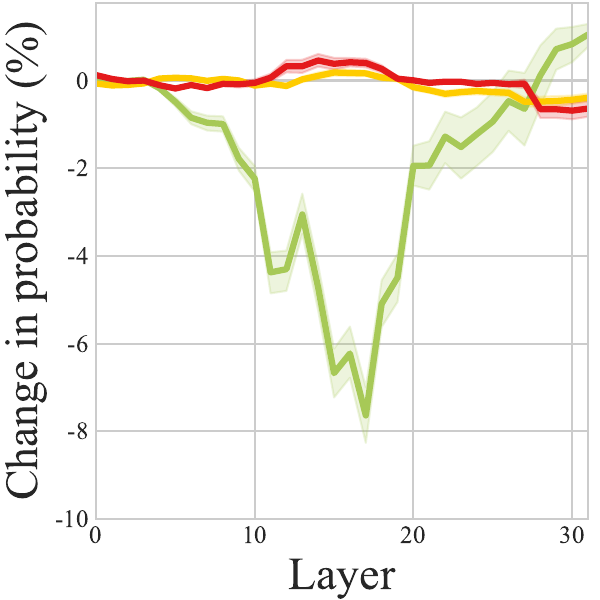}
}
\subfloat[ChooseRel]{
    \includegraphics[width=0.32\linewidth]{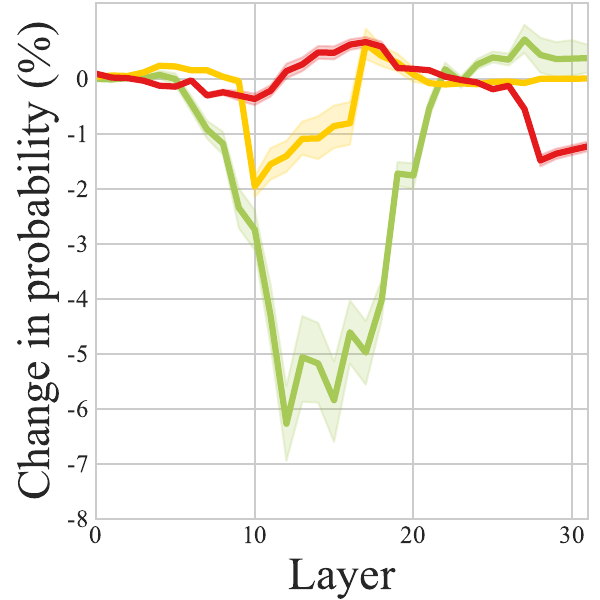}
}\\
\subfloat[CompareAttr]{
    \includegraphics[width=0.32\linewidth]{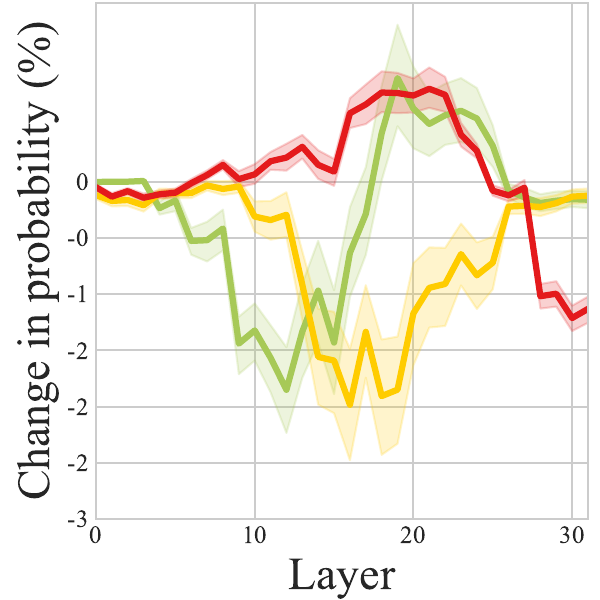}
}
\subfloat[LogicalObj]{
    \includegraphics[width=0.32\linewidth]{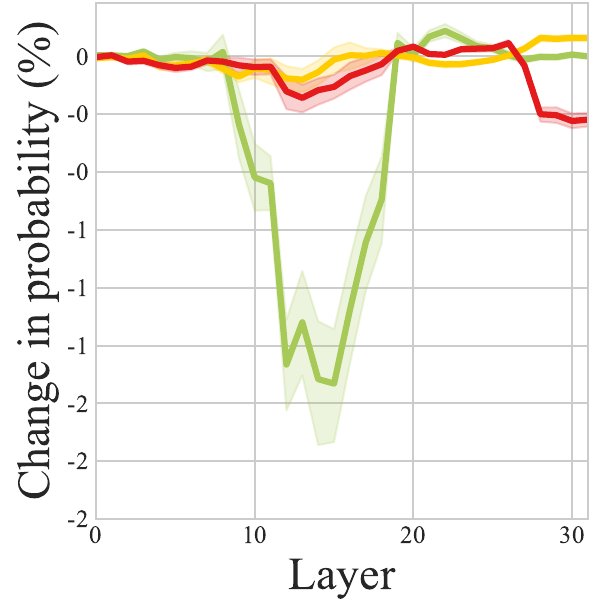}
}
\subfloat[QueryAttr]{
    \includegraphics[width=0.32\linewidth]{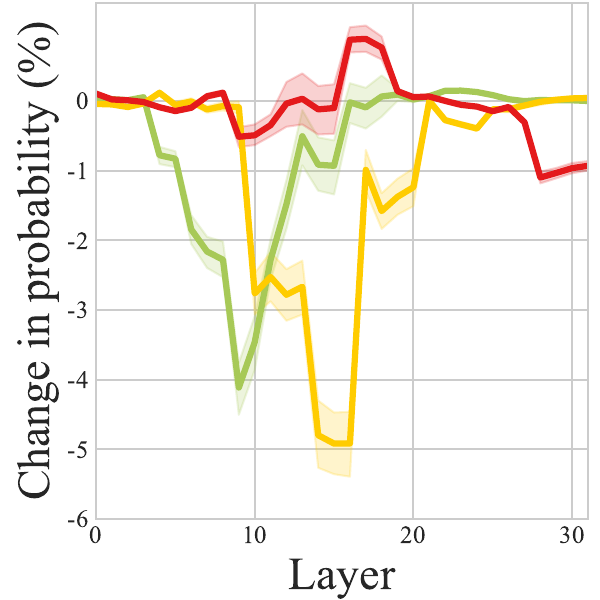}
}
\caption{The relative changes in prediction probability on \textit{llama3-llava-next-8b} with six \textit{VQA} tasks. The \textit{Question}$\nrightarrow$ \textit{Last}, \textit{Image}$\nrightarrow$ \textit{Last} and  \textit{Last}$\nrightarrow$ \textit{Last} represent preventing \textit{last position} from attending to \textit{Question}, \textit{Image} and itself respectively.}
\label{fig:tolast_llama3-llava-next-8b}
\end{figure}

\begin{figure}[t]
\centering
\includegraphics[width=0.33\linewidth]{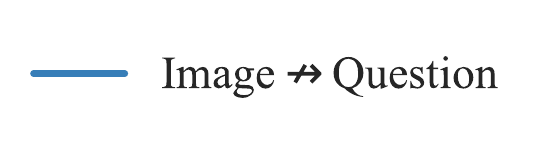}
\centering
\subfloat[ChooseAttr]{
    \includegraphics[width=0.32\linewidth]{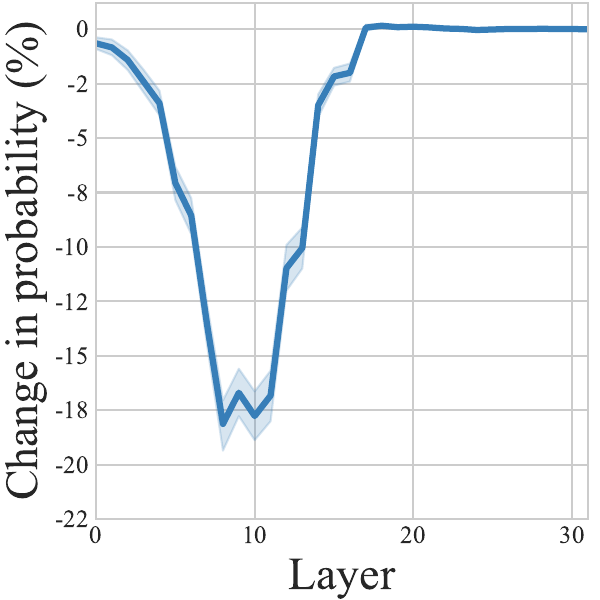}
}
\subfloat[ChooseCat]{
    \includegraphics[width=0.32\linewidth]{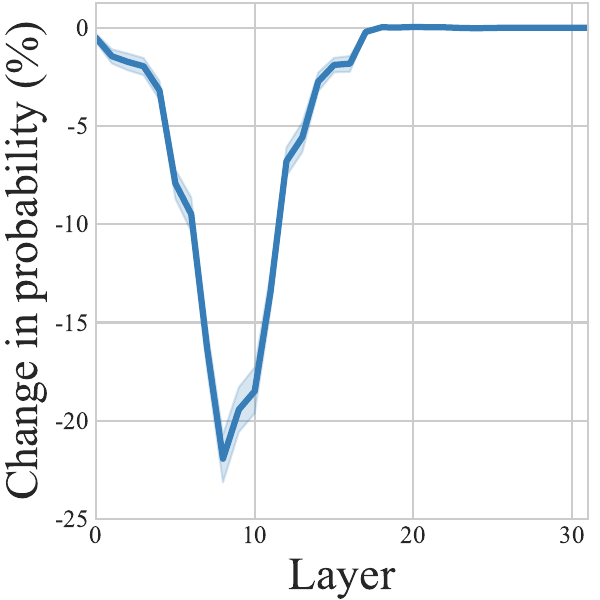}
}
\subfloat[ChooseRel]{
    \includegraphics[width=0.32\linewidth]{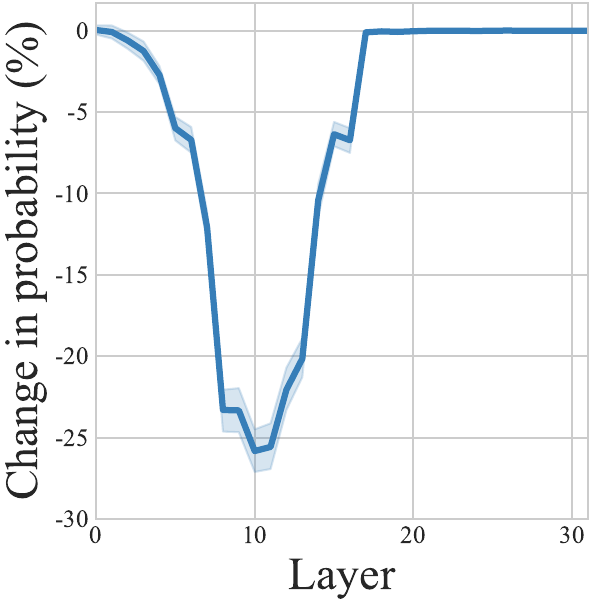}
}\\
\subfloat[CompareAttr]{
    \includegraphics[width=0.32\linewidth]{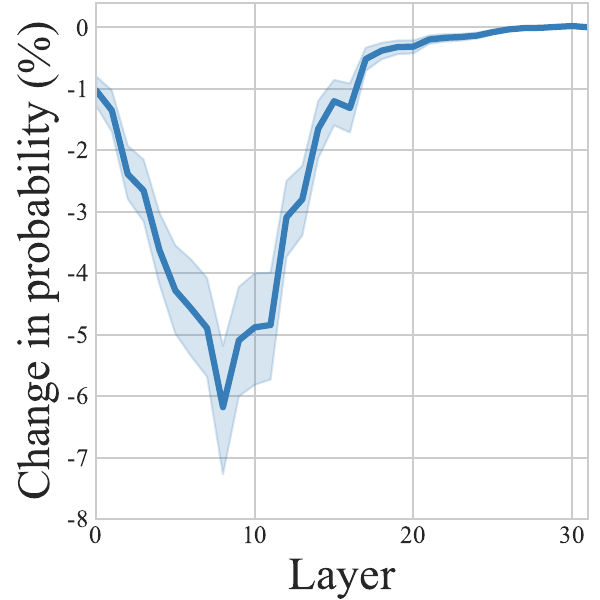}
}
\subfloat[LogicalObj]{
    \includegraphics[width=0.32\linewidth]{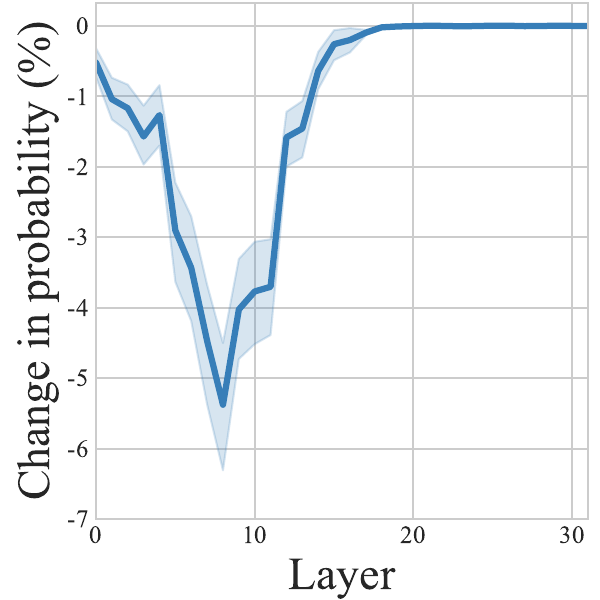}
}
\subfloat[QueryAttr]{
    \includegraphics[width=0.32\linewidth]{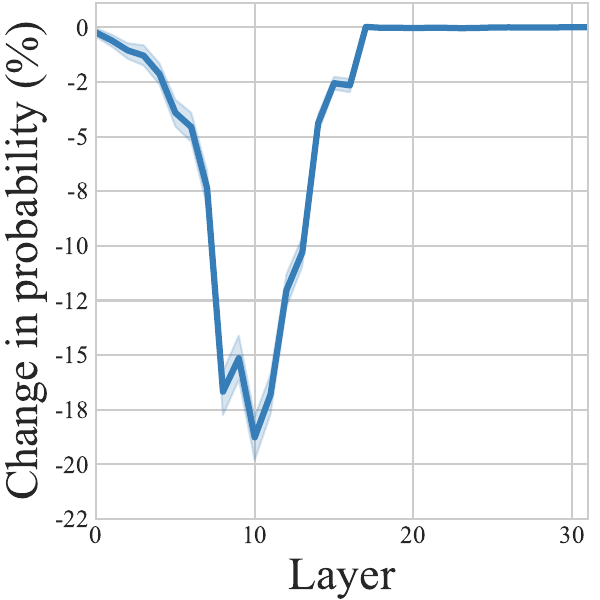}
}
\caption{
The relative changes in prediction probability when blocking attention edges from the \textit{question} positions to the \textit{image} positions on \textit{llama3-llava-next-8b} with six \textit{VQA} tasks.
}
\label{fig:imageToquestion_llama3-llava-next-8b}
\end{figure}

\begin{figure}[t]
\centering
\includegraphics[width=1.0\linewidth]{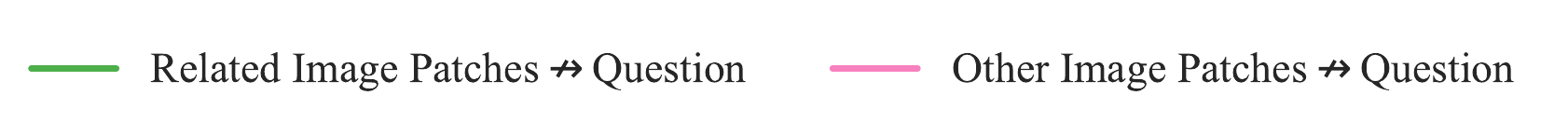}
\\
\centering
\subfloat[ChooseAttr]{
    \includegraphics[width=0.32\linewidth]{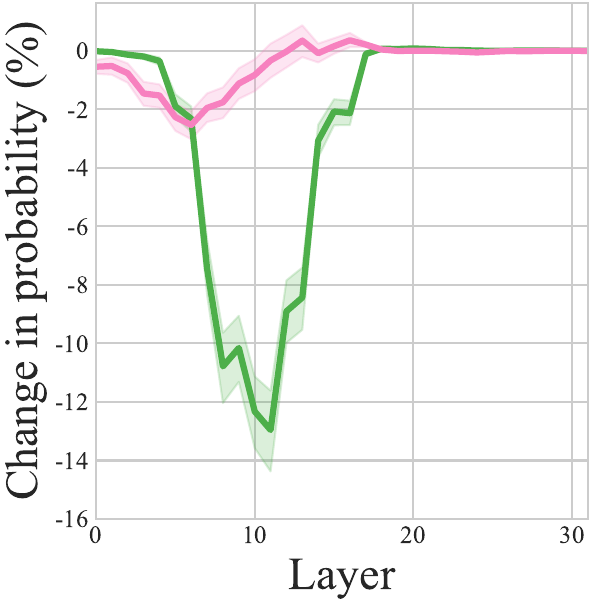}
}
\subfloat[ChooseCat]{
    \includegraphics[width=0.32\linewidth]{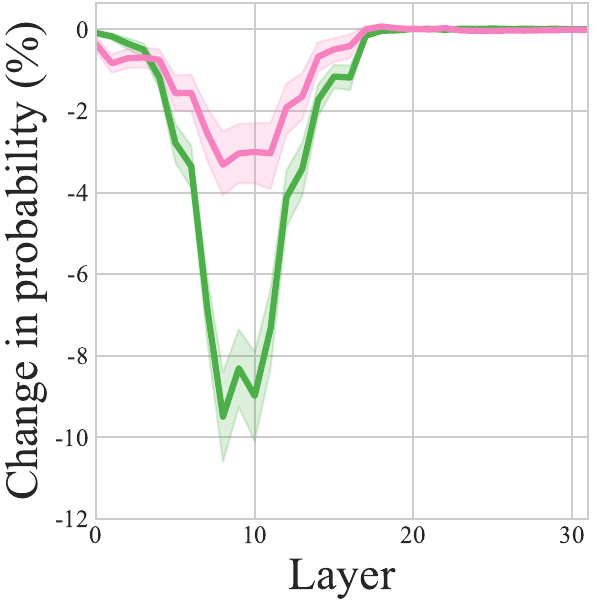}
}
\subfloat[ChooseRel]{
    \includegraphics[width=0.32\linewidth]{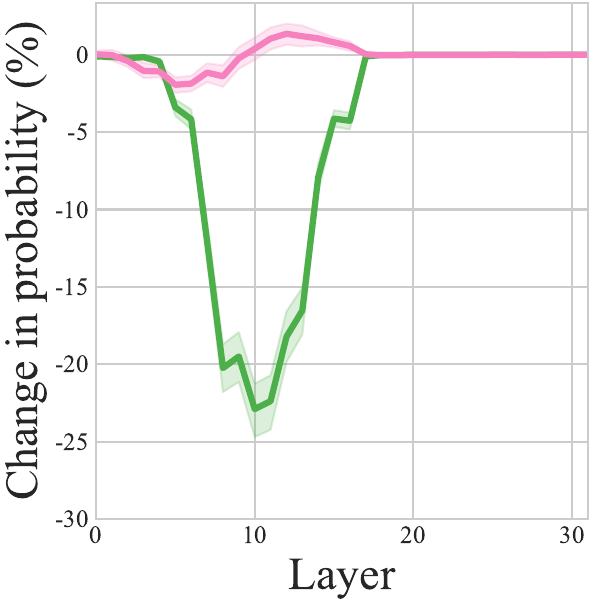}
}\\
\subfloat[CompareAttr]{
    \includegraphics[width=0.32\linewidth]{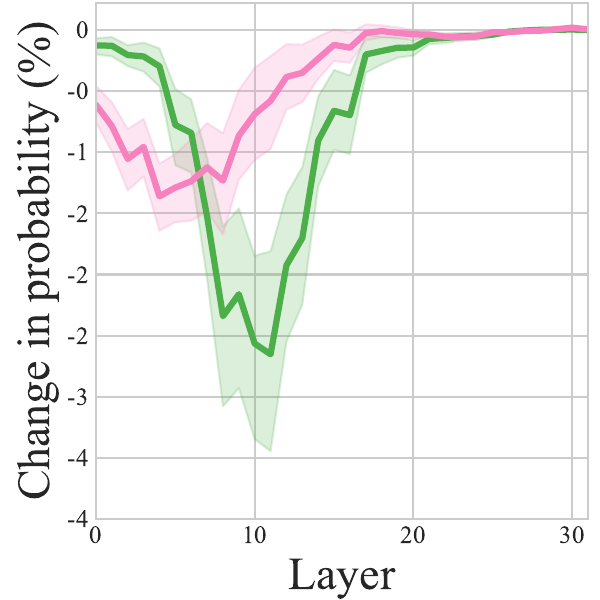}
}
\subfloat[LogicalObj]{
    \includegraphics[width=0.32\linewidth]{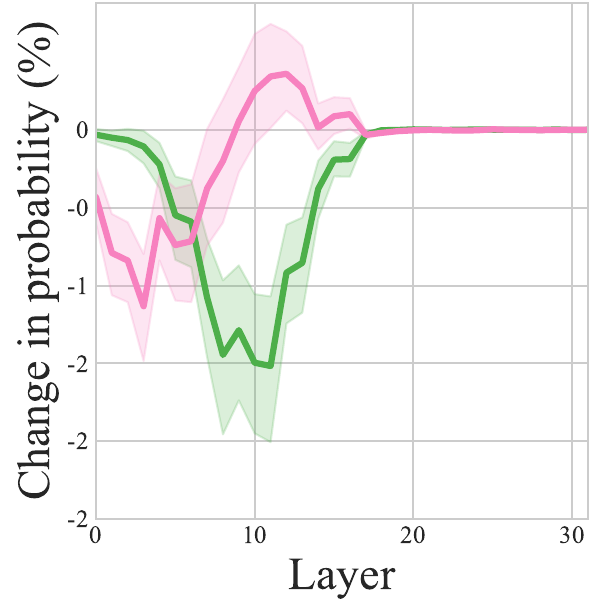}
}
\subfloat[QueryAttr]{
    \includegraphics[width=0.32\linewidth]{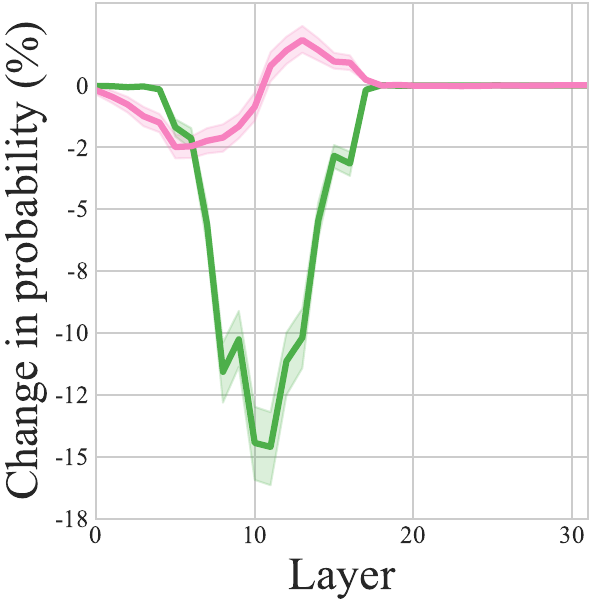}
}
\caption{The relative changes in prediction probability on \textit{llama3-llava-next-8b} with six \textit{VQA} tasks.
\textit{Related Image Patches}$\nrightarrow$\textit{question} and \textit{Other Image Patches}$\nrightarrow$\textit{question} represent blocking the position of \textit{question} from attending to that of different image patches, region of interest and remainder, respectively.}
\label{fig:imagepatchesToquestion_llama3-llava-next-8b}
\end{figure}

\begin{figure}[t]
\centering
\includegraphics[width=0.8\linewidth]{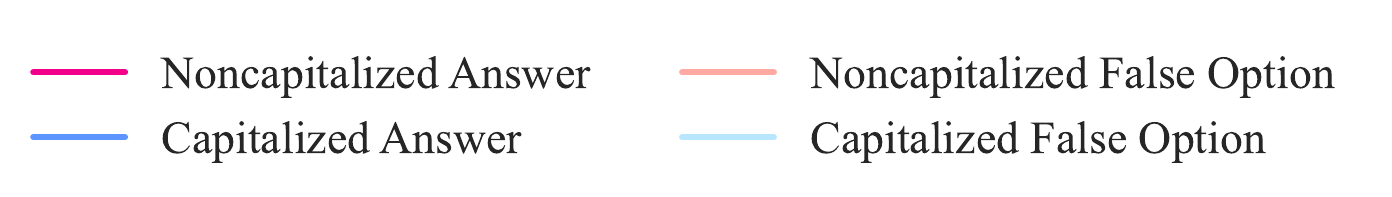}

\centering
\subfloat[ChooseAttr]{
    \includegraphics[width=0.32\linewidth]{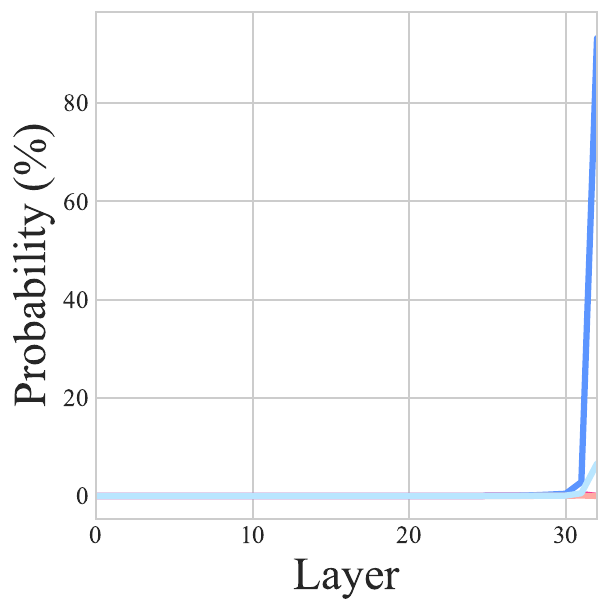}
}
\subfloat[ChooseCat]{
    \includegraphics[width=0.32\linewidth]{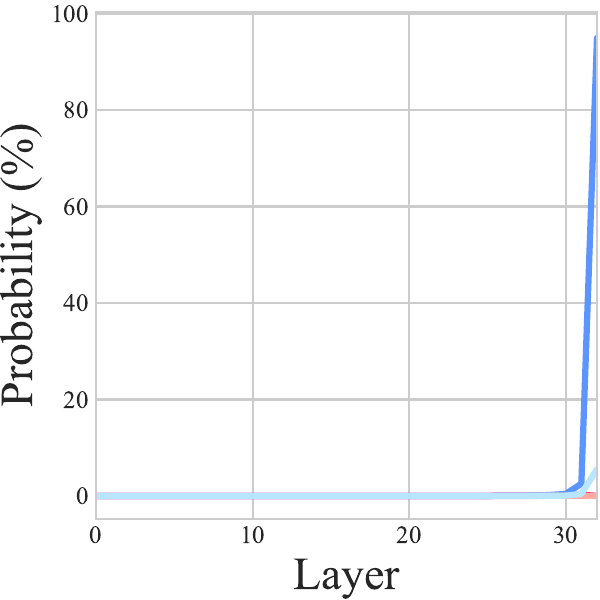}
}
\subfloat[ChooseRel]{
    \includegraphics[width=0.32\linewidth]{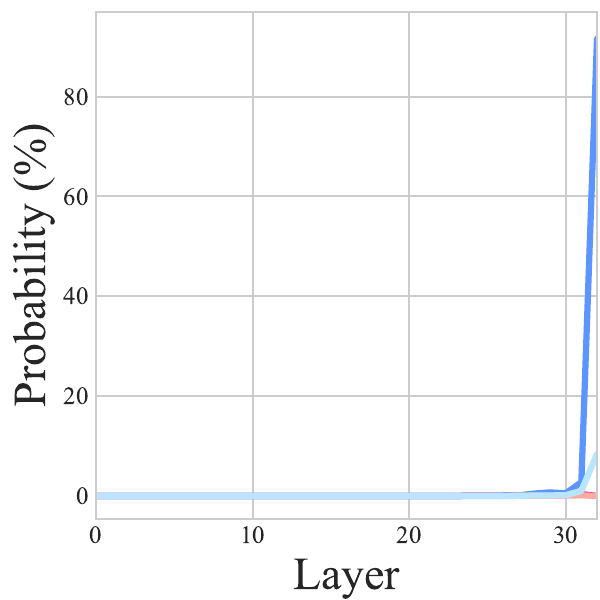}
}\\
\subfloat[CompareAttr]{
    \includegraphics[width=0.32\linewidth]{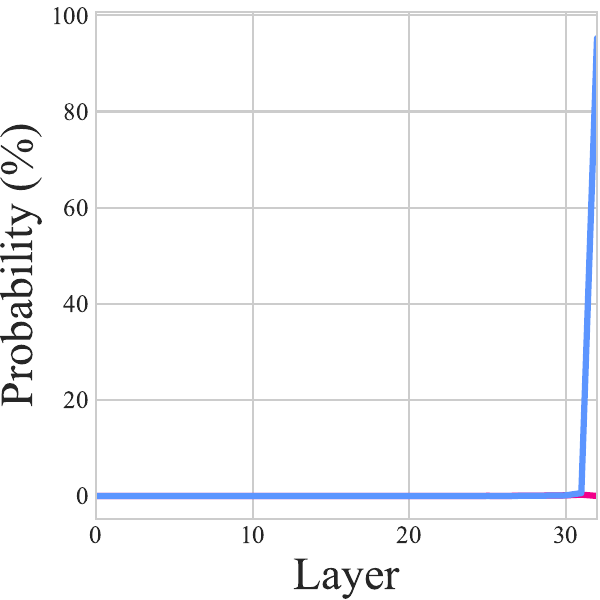}
}
\subfloat[LogicalObj]{
    \includegraphics[width=0.32\linewidth]{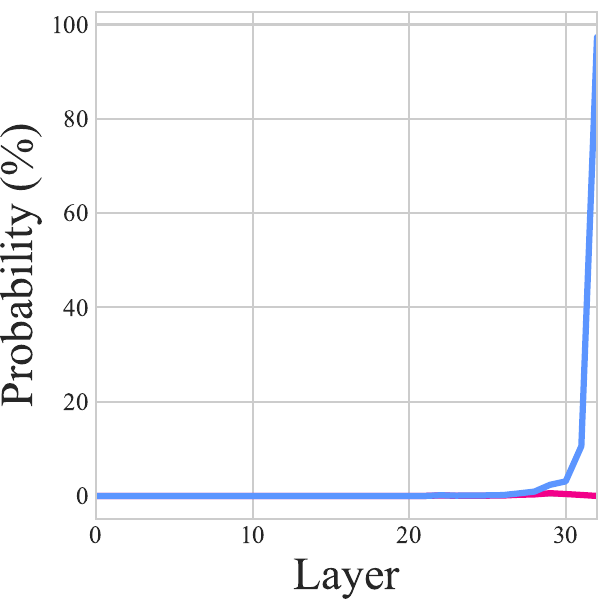}
}
\subfloat[QueryAttr]{
    \includegraphics[width=0.32\linewidth]{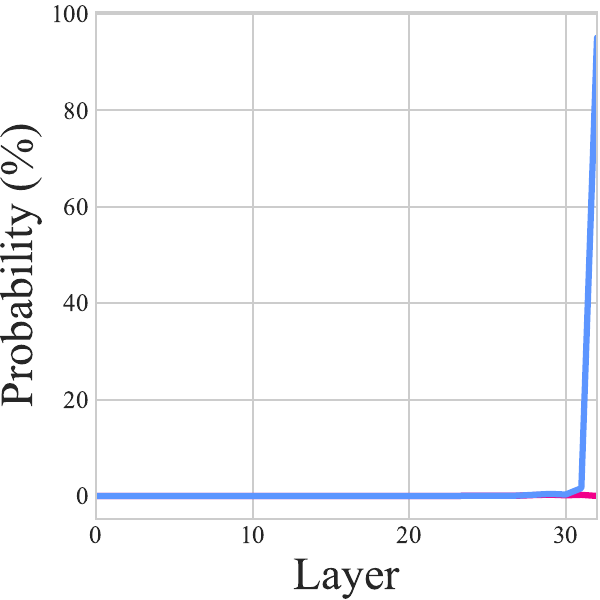}
}
\caption{The probability of the answer word at the \textit{last position} across all layers in llama3-llava-next-8b with six VQA tasks. \textit{Capitalized Answer} and \textit{Noncapitalized Answer} represent the answer word with or without the uppercase of the initial letter, respectively. As the tasks of \textit{ChooseAttr}, \textit{ChooseCat} and \textit{ChooseRel} contain \textit{false option}, we also provide the probability of it.}
\label{fig:last_position_prob_llama3-llava-next-8b}
\end{figure}

\textit{Llama3-LLaVA-NEXT-8b} has quiet different architecture with \textit{LLaVA-1.5-13b} in the main body of the paper. The difference between them includes the layer number, the way of processing image patch features, and the attention mechanism. The \textit{Llama3-LLaVA-NEXT-8b} has 32 layers verse 40 layers in \textit{LLaVA-1.5-13b}. \textit{LLaVA-1.5-13b} directly feeds the original fixed-length image patch features from the image encoder into the LLM as input tokens. In contrast, \textit{Llama3-LLaVA-NEXT-8b} employs a dynamic high-resolution technique, which dynamically adjusts image resolution, resulting in variable-length image patch features with higher resolution.  Specifically, the higher resolution is implemented by splitting the image into grids and encoding them independently. As for the attention mechanism, \textit{LLaVA-1.5-13b} use a standard and dense transformer architecture \cite{vaswani_attention_2017} while \textit{Llama3-LLaVA-NEXT-8b} adopts grouped query attention \cite{ainslie2023gqa} where the queries are grouped and the queries in the same group has shared key and value.

The information flow from different parts of the input sequence (\textit{image} and \textit{question}) to \textit{last position}, from \textit{image} to \textit{question} and from different image patches (\textit{related image patches} and \textit{other image patches}) to \textit{question}, as shown in \Cref{fig:tolast_llama3-llava-next-8b}, \Cref{fig:imageToquestion_llama3-llava-next-8b} and \Cref{fig:imagepatchesToquestion_llama3-llava-next-8b} respectively, are consistent with the observations for the \textit{LLaVA-1.5-13b} model, as shown in \Cref{fig:tolast_13b}, \Cref{fig:imageToquestion_13b} and \Cref{fig:imagepatchesToquestion_13b} respectively, in the main body of the paper. 
Although the information flow from \textit{image} to \textit{question} in \Cref{fig:imageToquestion_llama3-llava-next-8b} appears to exhibit only a single drop, the \Cref{fig:imagepatchesToquestion_llama3-llava-next-8b} reveals that, in lower layers, the information flow from \textit{Other Image Patches} to the \textit{question} play a dominant role compared to that from \textit{Related Image Patches}  to \textit{question} and in following layers, information flow from \textit{Related Image Patches} to \textit{question} are more notable than that form \textit{Other Image Patches} to \textit{question}.
This observation indicates that the model still has a two-stage multimodal information integration process. Specifically, in the first stage, the model focuses on generating holistic representations of the entire image. In the second stage, it refines these representations to align them more closely with the specific given question.
Subsequently, in the middle layers, the critical multimodal information flows from the \textit{question} positions to the \textit{last position} for the final prediction. 
Moreover, the probability changes for the \textit{Capitalized Answer} across all layers, as illustrated in \Cref{fig:last_position_prob_llama3-llava-next-8b}, align closely with the results in the main body of the paper while no such pattern is observed for the \textit{Noncapitalized Answer}. This suggests that the model generates the syntactically correct answer directly, without a distinct intermediate step of semantic generation followed by syntactic correction. A potential explanation for this behavior is that when \textit{Llama3} generates an answer to a given question, it first outputs a “\textbackslash n” token, which may act as a cue to produce an answer word starting with an uppercase letter.

\section{The fine-grain analysis for information flow}
\label{app: fine-grain analysis}
In the main body of the paper, we primarily focus on analyzing the information flow between one specific combination of (\textit{image}, \textit{question}, and \textit{last position}) for analyzing the multimodal information integration. In this section, we will further investigate the information flow between fine-grain parts of the input sequence, including the \textit{question without options}, \textit{true option}, \textit{false option}, \textit{objects} in the question, \textit{question without objects}, \textit{related image patches} and \textit{other image patches}. We also use the same \textit{attention knockout} method to block the attention edge between them to investigate the information flow between them.

\subsection{Different parts of the \textit{\textbf{question}} to the \textit{\textbf{last position}}}

\begin{figure}[t]
\centering
\subfloat[ChooseAttr]{
    \includegraphics[width=0.32\linewidth]{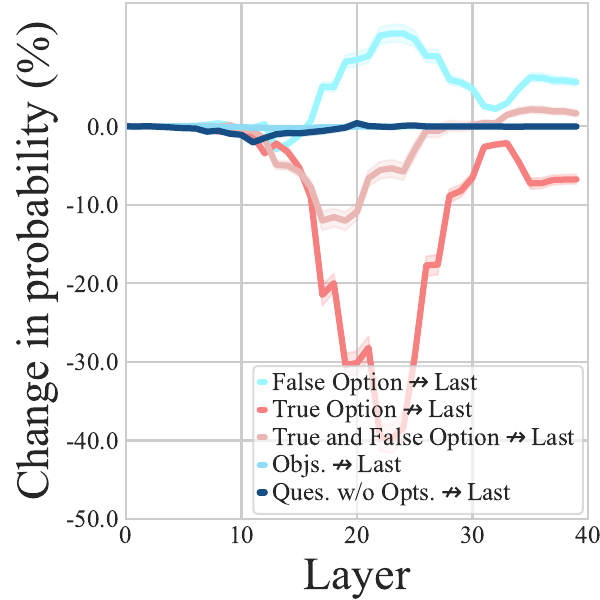}
}
\subfloat[ChooseCat]{
    \includegraphics[width=0.32\linewidth]{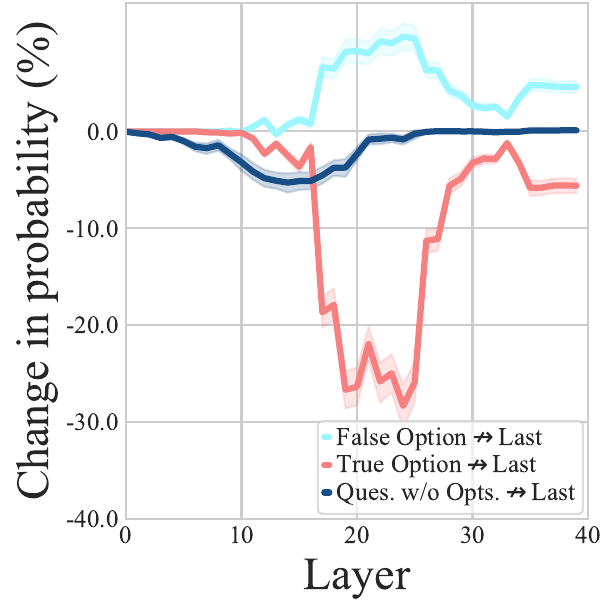}
}
\subfloat[ChooseRel]{
    \includegraphics[width=0.32\linewidth]{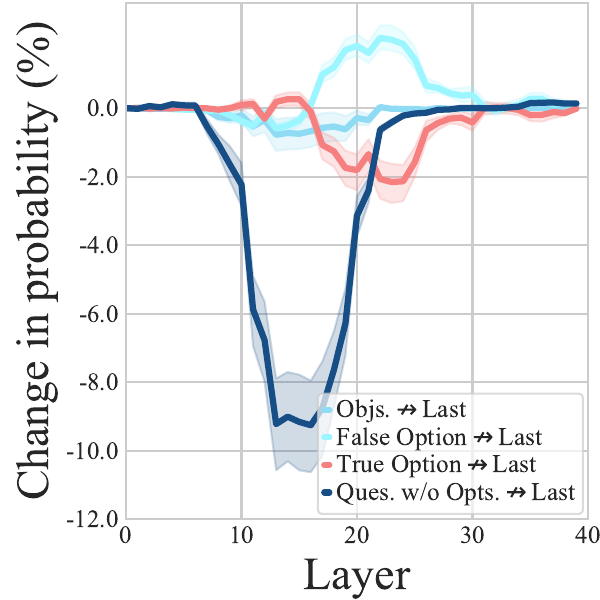}
}\\
\subfloat[CompareAttr]{
    \includegraphics[width=0.32\linewidth]{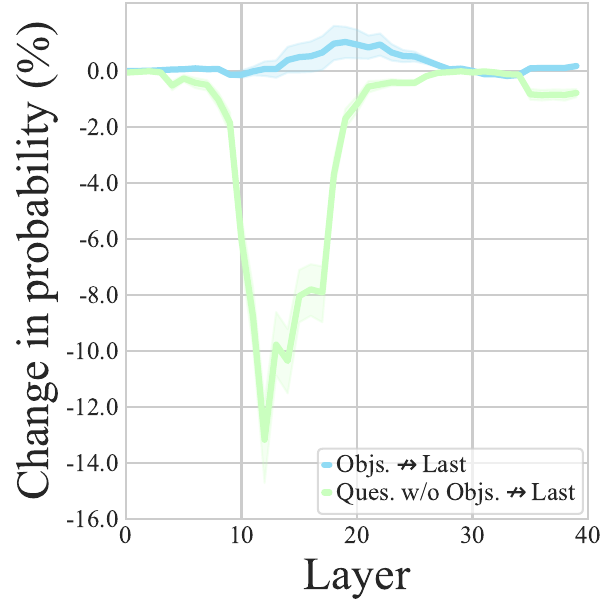}
}
\subfloat[LogicalObj]{
    \includegraphics[width=0.32\linewidth]{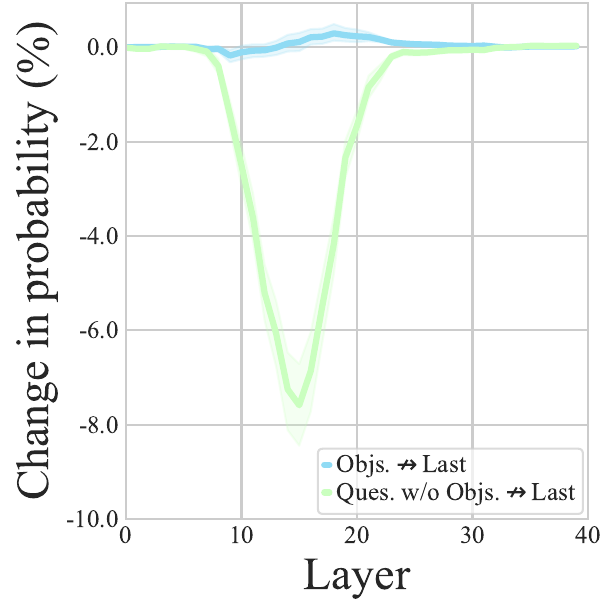}
}
\subfloat[QueryAttr]{
    \includegraphics[width=0.32\linewidth]{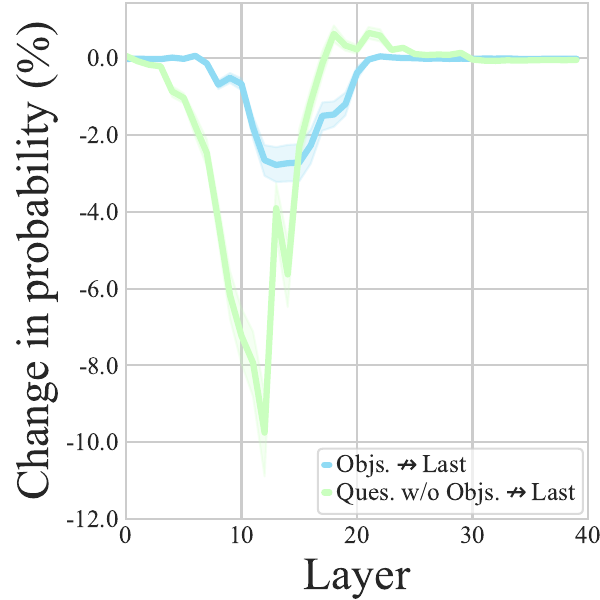}
}
\caption{The relative changes in prediction probability on \textit{LLaVA-v1.5-13b} with six \textit{VQA} tasks. Preventing \textit{Last Position} from attending to different parts of \textit{Question}, such as \textit{True Option}, \textit{False Option}, \textit{Objects} in question, \textit{Question without Options}, \textit{Question without Objects}, both \textit N{True Option} and \textit{False Option} together.}
\label{fig:diffpartofquestiontolast_LLaVA-v1.5-13b}
\end{figure}

In the tasks of \textit{ChooseAttr}, \textit{ChooseCat} and \textit{ChooseRel}, for each layer $\ell$, we block \textit{last position} from attending to different parts of \textit{question}, including \textit{question without options}, \textit{true option}, \textit{false option}, with the same window size ($k=9$) around the $\ell$-th layer and observe the change in the probability of the answer word at the \textit{last position}. In the tasks of \textit{CompareAttr} \textit{LogicalObj} and \textit{QueryAttr}, we conduct the same operations with the above tasks except for blocking \textit{last position} from attending to \textit{objects} or \textit{question without objects} as these tasks do not contain \textit{options} in the question.

As shown in \Cref{fig:diffpartofquestiontolast_LLaVA-v1.5-13b} (a), (b) and (c), for the tasks of \textit{ChooseAttr}, \textit{ChooseCat} and \textit{ChooseRel}, the \textit{true option} and \textit{false option} flowing the information to the \textit{last position} occur in similar layers (higher layers) in the model. When blocking \textit{last position} from attending \textit{true option}, the probability obtain a reduction, while blocking \textit{last position} from attending \textit{false option} increases the probability of the correct answer word. The increase is reasonable because the question without the false option becomes easy for the modal. For the tasks of \textit{ChooseAttr} and \textit{ChooseCat}, in the information flowing to \textit{last position}, the \textit{options} play a dominant role while \textit{question without options} only results in a small reduction for the probability fo the correct answer word. 
In contrast, for the \textit{ChooseRel} task, the \textit{true option} does not significantly reduce the probability of the correct answer word. This may stem from the format of the \textit{ChooseRel} questions, where the options are positioned in the middle of the question, rather than at the end as in the \textit{ChooseAttr} and \textit{ChooseCat} tasks. As a result, the options in \textit{ChooseRel} are less effective at aggregating the complete contextual information of the question within an auto-regressive transformer decoder. Consequently, the flow of information from the \textit{option} to the \textit{final position} becomes less critical in determining the correct answer.

As the questions in our dataset target one or more specific objects in the image, we also conduct experiments on blocking \textit{last position} from attending to \textit{objects} or \textit{question without objects}. As shown in \Cref{fig:diffpartofquestiontolast_LLaVA-v1.5-13b} (d), (e) and (f), the critical information from the \textit{objects} does not directly transfer into the \textit{last position} compared to that form \textit{question without objects} to \textit{last position}. This implies that the \textit{objects} might affect the final prediction in an indirect way.

\subsection{Different parts of the \textit{\textbf{question}} to different parts of the \textit{\textbf{question}}}
\begin{figure}[t]
\subfloat[ChooseAttr]{
    \includegraphics[width=0.32\linewidth]{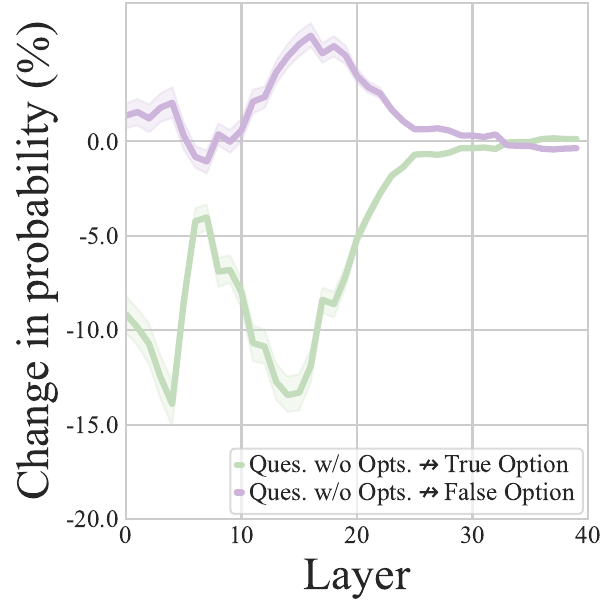}
}
\subfloat[ChooseCat]{
    \includegraphics[width=0.32\linewidth]{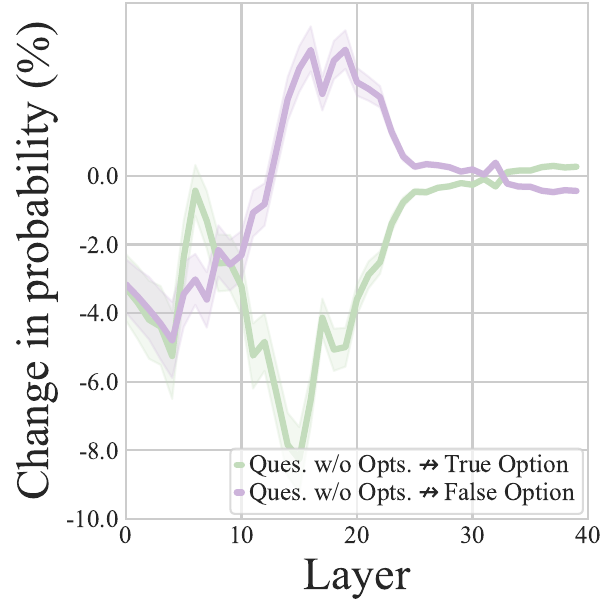}
}
\subfloat[ChooseRel]{
    \includegraphics[width=0.32\linewidth]{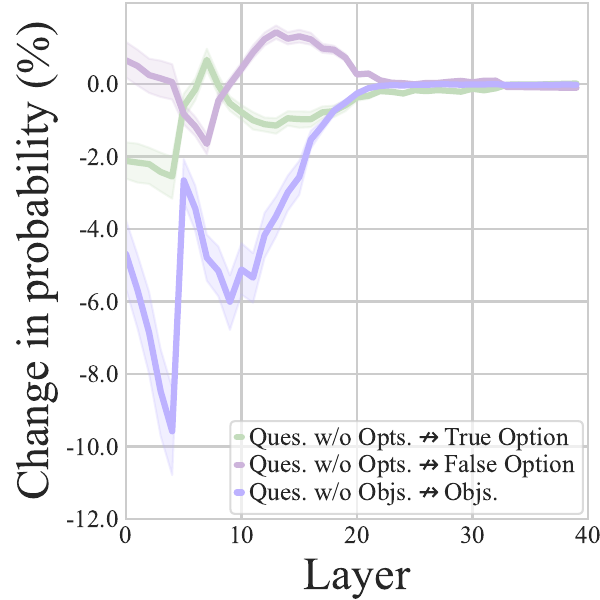}
}\\
\subfloat[CompareAttr]{
    \includegraphics[width=0.32\linewidth]{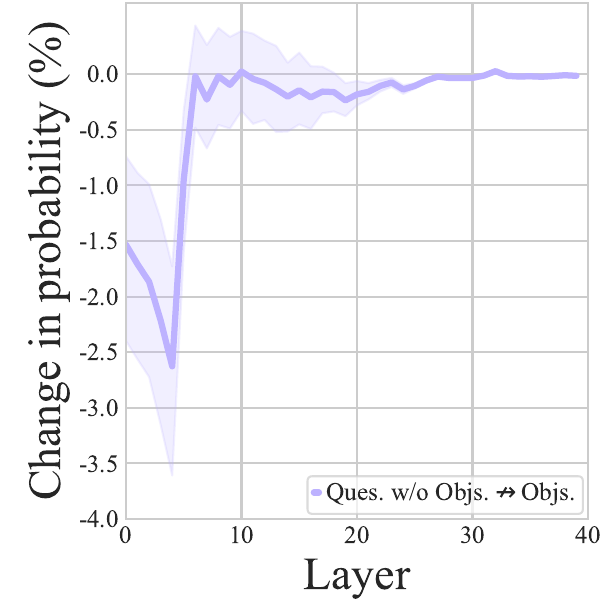}
}
\subfloat[LogicalObj]{
    \includegraphics[width=0.32\linewidth]{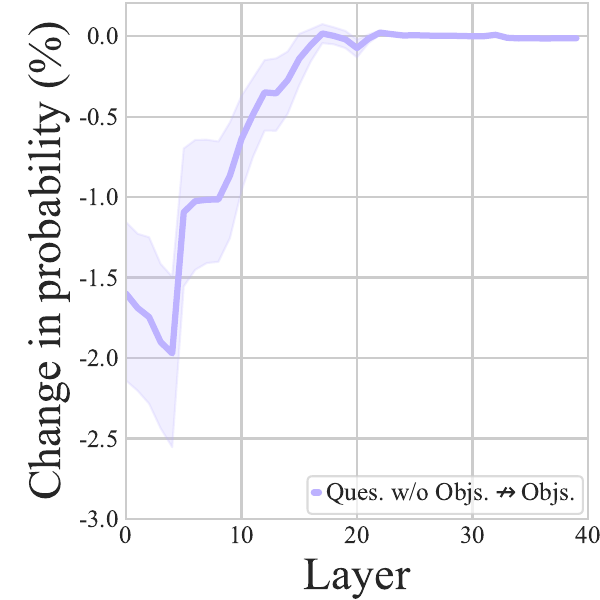}
}
\caption{The relative changes in prediction probability on \textit{LLaVA-v1.5-13b} with six \textit{VQA} tasks. Preventing information flow from \textit{Question without Option} to \textit{Options} and from \textit{Question without Objects} to \textit{Objects}.}
\label{fig:diffpartofquestiontodiffpartofquestion_LLaVA-v1.5-13b}
\end{figure}

In the tasks of \textit{ChooseAttr}, \textit{ChooseCat} and \textit{ChooseRel}, for each layer $\ell$, we block \textit{options} from attending to \textit{question without options} with the same window size ($k=9$) around the $\ell$-th layer and observe the change in the probability of the answer word. In the tasks of \textit{CompareAttr} and \textit{LogicalObj}, we conduct the same operations with the above tasks except for blocking \textit{objects} from attending to \textit{question without objects}.

As shown in \Cref{fig:diffpartofquestiontodiffpartofquestion_LLaVA-v1.5-13b} (a), (b) and (c), for the tasks of \textit{ChooseAttr}, \textit{ChooseCat} and \textit{ChooseRel}, the information flow from \textit{question without options} to \textit{true option} occurs in similar transformer layers with that from \textit{question without options} to \textit{false option}. 
We also observe that these indirect information flows from \textit{question without options} to \textit{false option} occur before the information flow from \textit{options} to \textit{last position} as shown in \Cref{fig:diffpartofquestiontolast_LLaVA-v1.5-13b}. This indicates that the information of the question is aggregated into the \textit{options} in lower layers and then the information in \textit{options} is transferred to the \textit{last position} for the prediction of the final answer in higher layers.
For the tasks of \textit{CompareAttr} and \textit{LogicalObj}, we observe that the information flow from \textit{question without objects} to \textit{objects} occurs in lower layers. 

\subsection{\textit{\textbf{Image}} to different parts of \textit{\textbf{question}}}
\begin{figure}[t]
\centering
\subfloat[ChooseAttr]{
    \includegraphics[width=0.32\linewidth]{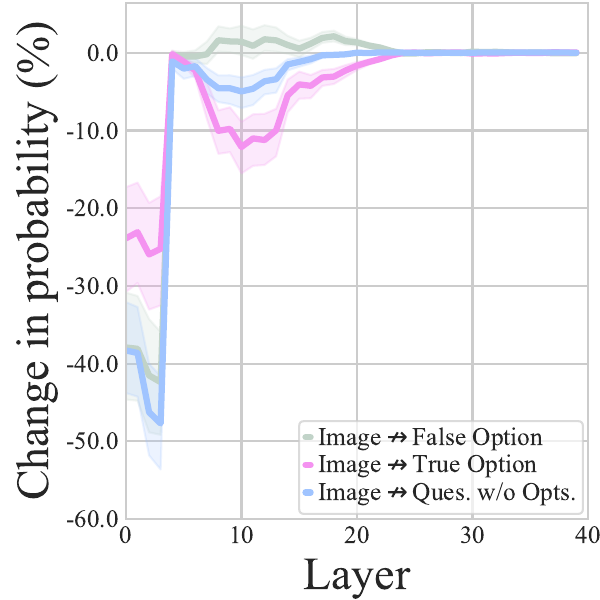}
}
\subfloat[ChooseCat]{
    \includegraphics[width=0.32\linewidth]{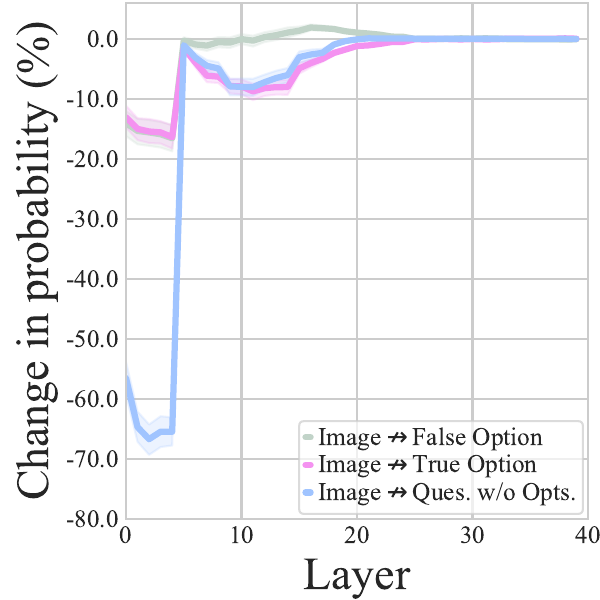}
}
\subfloat[ChooseRel]{
    \includegraphics[width=0.32\linewidth]{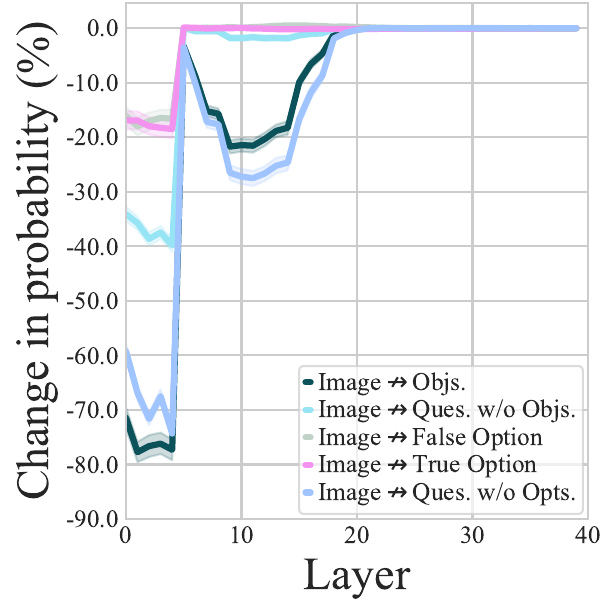}
}\\
\subfloat[CompareAttr]{
    \includegraphics[width=0.32\linewidth]{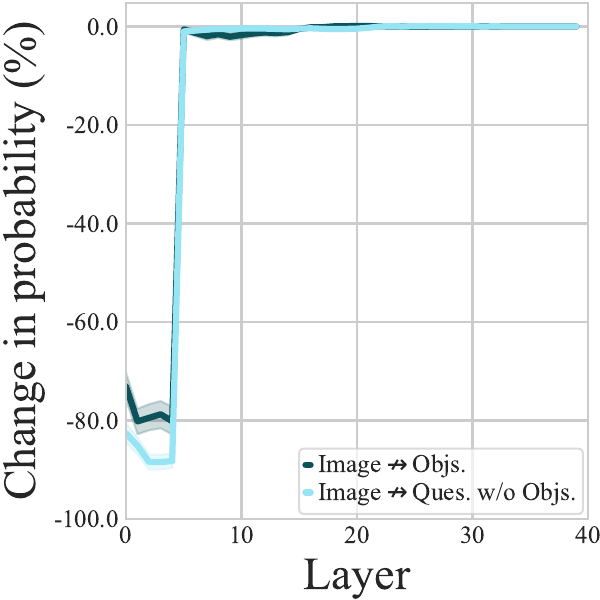}
}
\subfloat[LogicalObj]{
    \includegraphics[width=0.32\linewidth]{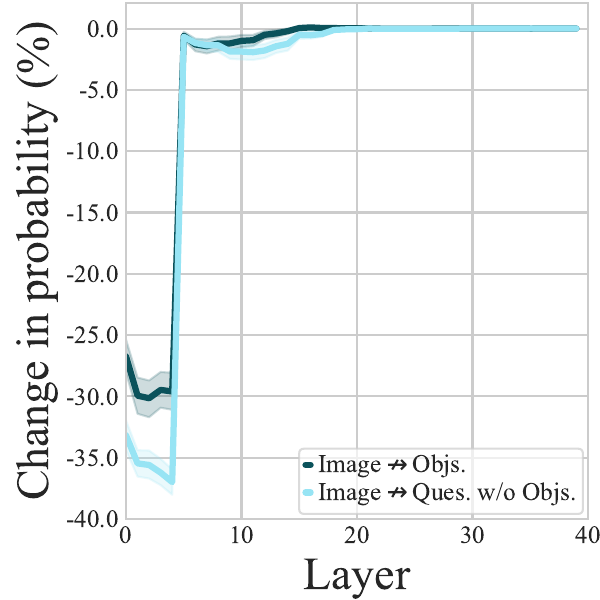}
}
\subfloat[QueryAttr]{
    \includegraphics[width=0.32\linewidth]{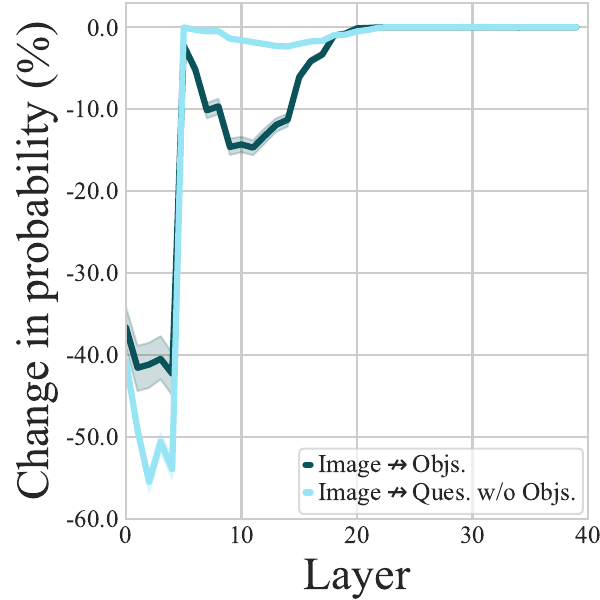}
}
\caption{The relative changes in prediction probability on \textit{LLaVA-v1.5-13b} with six \textit{VQA} tasks. Blocking the information flow from \textit{Image} to different parts of the \textit{question}, including \textit{True Option}, \textit{False Option}, \textit{Objects} in question, \textit{Question without Objects}, \textit{Question without Options}.}
\label{fig:Image2diffpartQues_LLaVA-v1.5-13b}
\end{figure}

 In the tasks of \textit{ChooseAttr}, \textit{ChooseCat} and \textit{ChooseRel}, for each layer $\ell$, we block attention edge between \textit{image} and different parts of \textit{question}, including \textit{question without options}, \textit{true option} and \textit{false option}, with the same window size ($k=9$) around the $\ell$-th layer and observe the change in the probability of the answer word. In the tasks of \textit{CompareAttr}, \textit{LogicalObj} and \textit{QueryAttr}, we carry out the same operations as in the above tasks except for blocking the edge of the attention between \textit{ image} and \textit{question without objects} or \textit{objects} respectively.

As illustrated in \Cref{fig:Image2diffpartQues_LLaVA-v1.5-13b}, the overall information flow from \textit{image} to different parts of the \textit{question} aligns consistently with the information flow from the \textit{image} to the entire \textit{question}, as depicted in \Cref{fig:imageToquestion_13b} in the main body of the paper. Specifically, two distinct flows are from the \textit{image} to the \textit{question}. Notably, however, different parts of the \textit{question} exhibit varying magnitudes of probability change, especially in the second-time drop in probability, which may be because different kinds of questions have different attention patterns to the image. For example, during the second-time drop in probability, in the tasks of \textit{ChooseAttr} and \textit{ChooseCat}, the \textit{image} information does not transfer to \textit{false option} while it transfers much more information to \textit{true option}. However, this pattern isn't observed in the task of \textit{ChooseRel}, where most \textit{image} information is transferred into \textit{question without options} and \textit{objects}.

\subsection{\textit{\textbf{Other image patches}} to different parts of \textit{\textbf{question}}}
\begin{figure}[t]
\centering
\subfloat[ChooseAttr]{
    \includegraphics[width=0.32\linewidth]{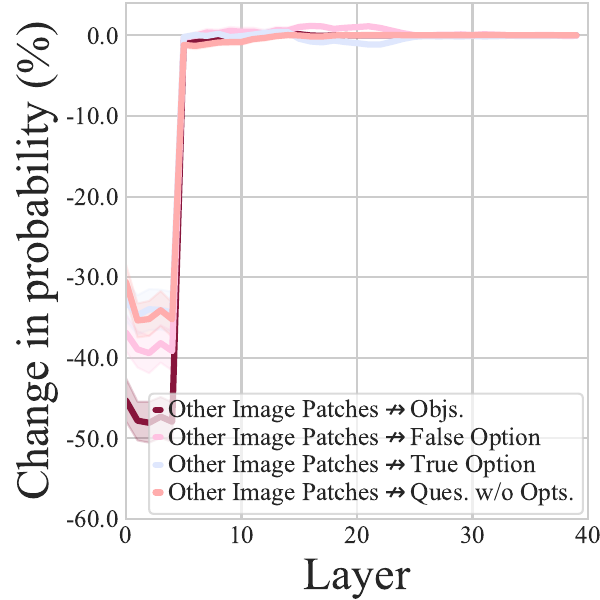}
}
\subfloat[ChooseCat]{
    \includegraphics[width=0.32\linewidth]{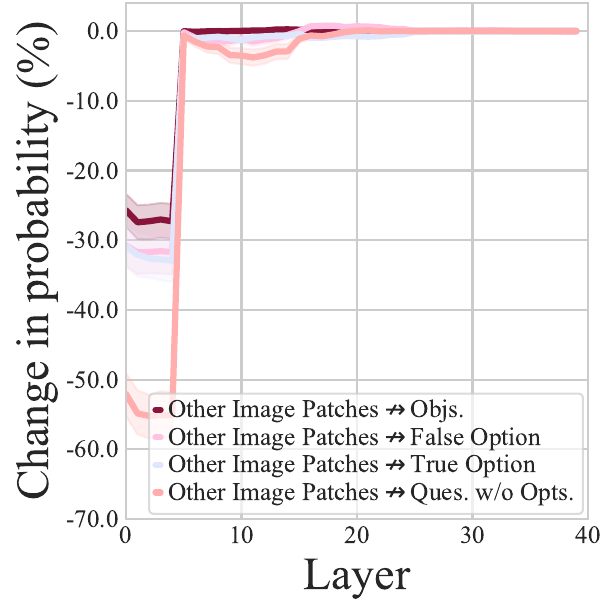}
}
\subfloat[ChooseRel]{
    \includegraphics[width=0.32\linewidth]{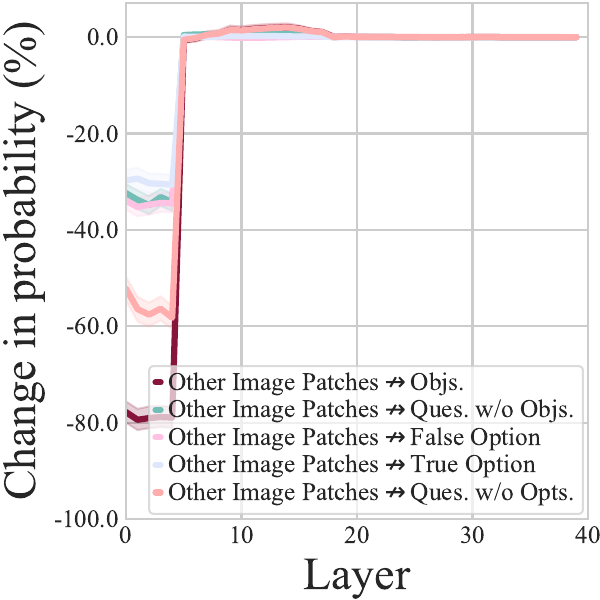}
}\\
\subfloat[CompareAttr]{
    \includegraphics[width=0.32\linewidth]{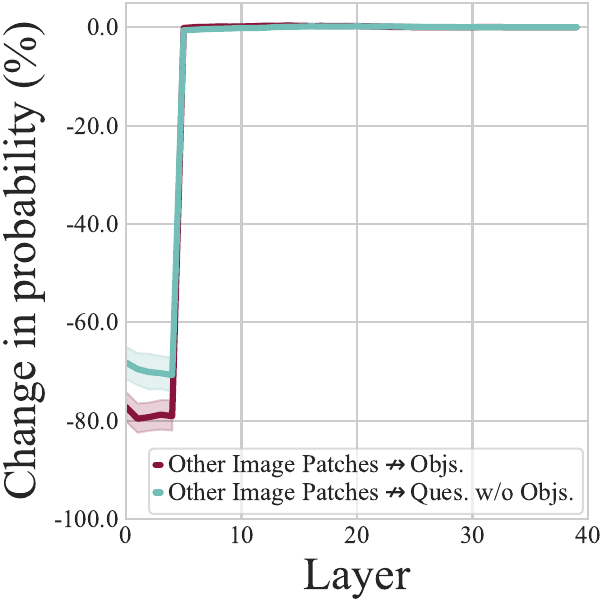}
}
\subfloat[LogicalObj]{
    \includegraphics[width=0.32\linewidth]{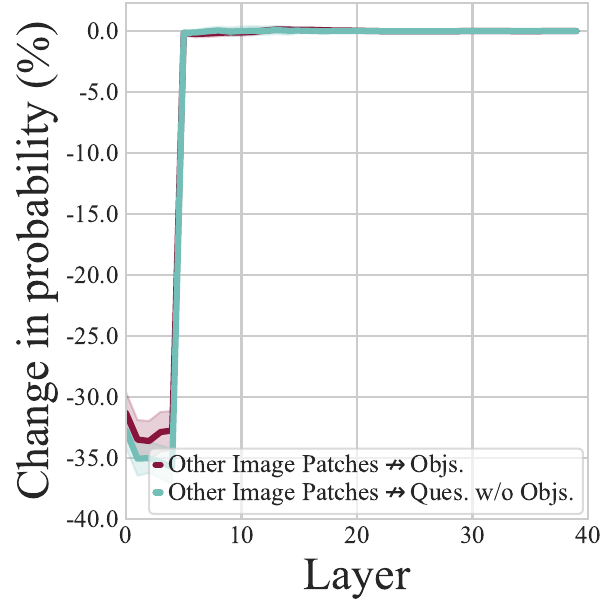}
}
\subfloat[QueryAttr]{
    \includegraphics[width=0.32\linewidth]{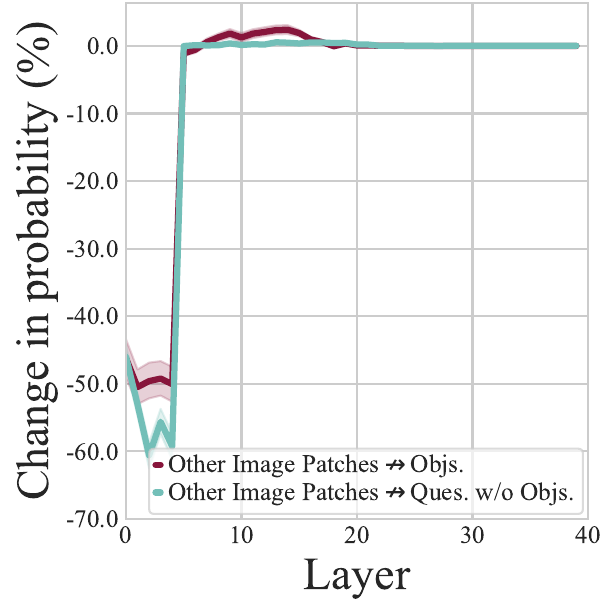}
}
\caption{The relative changes in prediction probability on \textit{LLaVA-v1.5-13b} with six \textit{VQA} tasks. Blocking the information flow from \textit{Other Image Patches} to different parts of the question, including \textit{True Option}, \textit{False Option}, \textit{Objects} in question, \textit{Question without Objects}, \textit{Question without Options}.}
\label{fig:otherImagePatches2diffpartQues_LLaVA-v1.5-13b}
\end{figure}

In the tasks of \textit{ChooseAttr}, \textit{ChooseCat} and \textit{ChooseRel}, for each layer $\ell$, we block attention edge between \textit{other image patches} and different parts of \textit{question}, including \textit{question without options}, \textit{true option}, \textit{false option}, \textit{objects} and \textit{question without objects}, with the same window size ($k=9$) around the $\ell$-th layer and observe the change in the probability of the answer word. In the tasks of \textit{CompareAttr}, \textit{LogicalObj} and \textit{QueryAttr}, we conduct the same operations with the above tasks except for blocking attention edge between \textit{other image patches} and \textit{question without objects} or \textit{objects} respectively.

As shown in \Cref{fig:otherImagePatches2diffpartQues_LLaVA-v1.5-13b}, the information flow from \textit{other image patches} to different parts of the \textit{question} for all six tasks consistently aligns the flow observed from \textit{other image patches} to the entire \textit{question}, as illustrated in \Cref{fig:imagepatchesToquestion_13b} in the main body of the paper. Specifically, the information flow dominantly occurs in the first-time drop in the probability in the lower layers, regardless of which part of the \textit{question} is being blocked.

\subsection{\textit{\textbf{Related image patches}} to different parts of \textit{\textbf{question}}}
\begin{figure}[t]
\centering
\subfloat[ChooseAttr]{
    \includegraphics[width=0.32\linewidth]{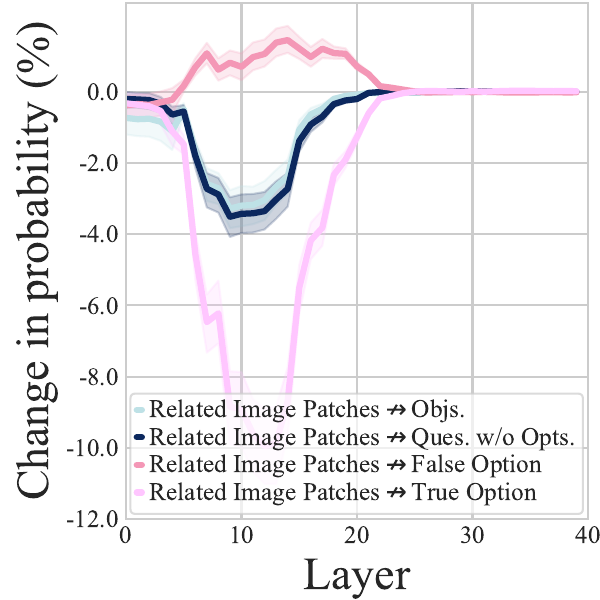}
}
\subfloat[ChooseCat]{
    \includegraphics[width=0.32\linewidth]{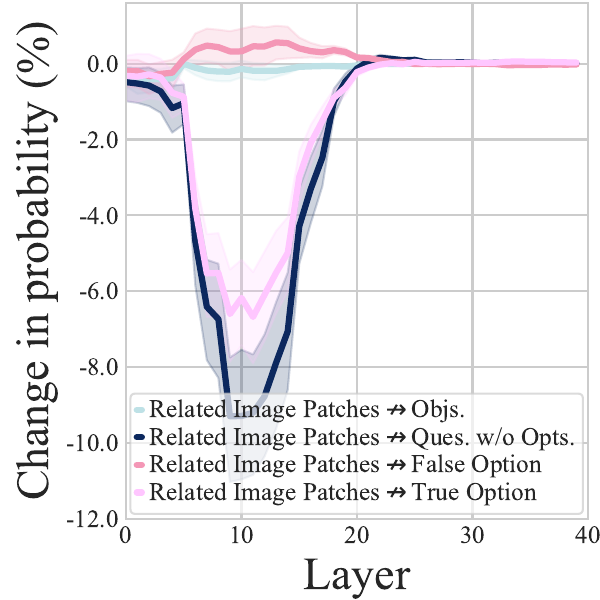}
}
\subfloat[ChooseRel]{
    \includegraphics[width=0.32\linewidth]{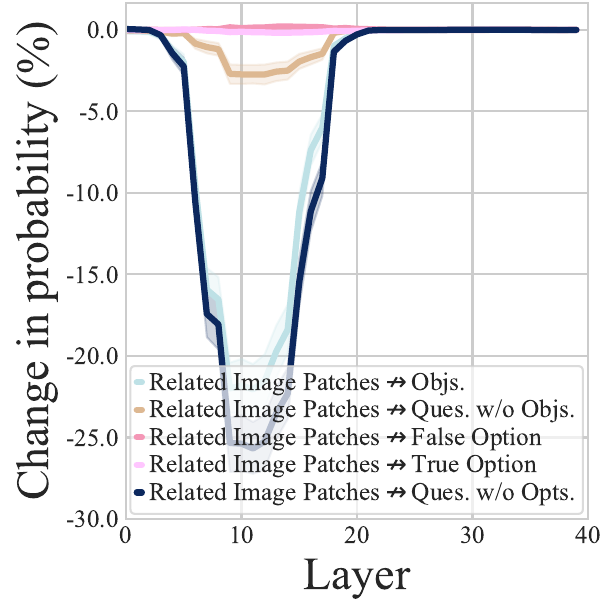}
}\\
\subfloat[CompareAttr]{
    \includegraphics[width=0.32\linewidth]{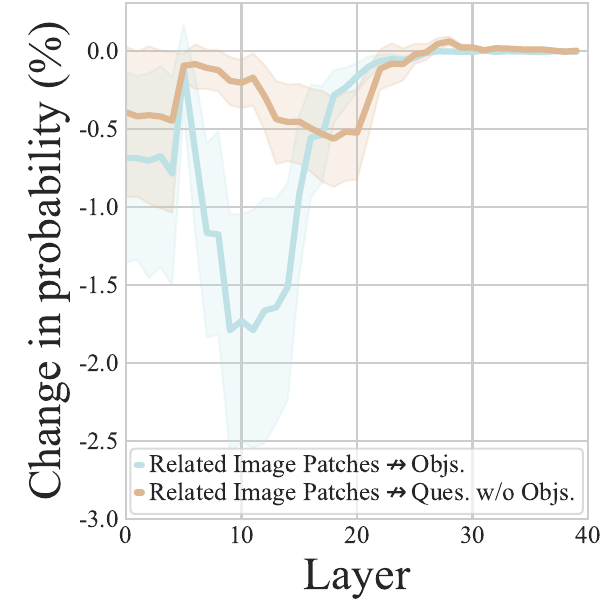}
}
\subfloat[LogicalObj]{
    \includegraphics[width=0.32\linewidth]{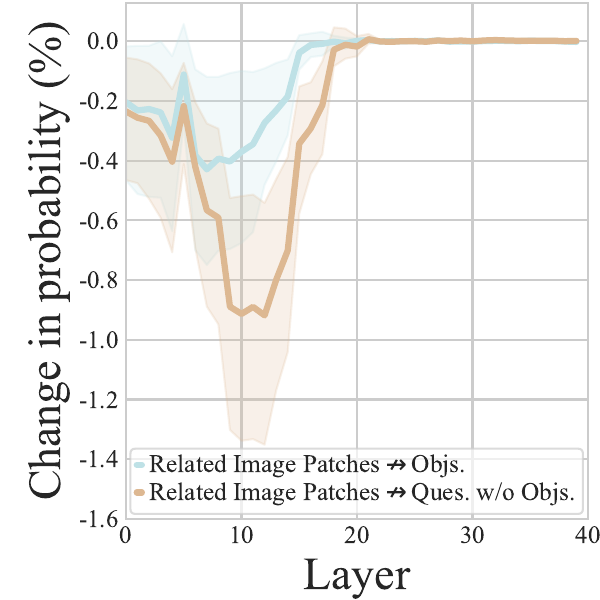}
}
\subfloat[QueryAttr]{
    \includegraphics[width=0.32\linewidth]{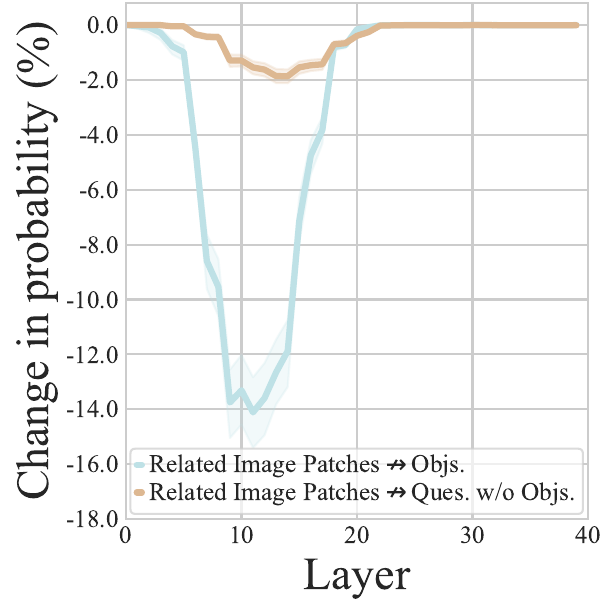}
}
\caption{The relative changes in prediction probability on \textit{LLaVA-v1.5-13b} with six \textit{VQA} tasks. Blocking the information flow from \textit{Related Image Patches} to different parts of the question, including \textit{True Option}, \textit{False Option}, \textit{Objects} in question, \textit{Question without Objects}, \textit{Question without Options}.}
\label{fig:relatedImagePatches2diffpartQues_LLaVA-v1.5-13b}
\end{figure}

In the tasks of \textit{ChooseAttr}, \textit{ChooseCat} and \textit{ChooseRel}, for each layer $\ell$, we block the attention edge between \textit{Related image patches} and different parts of \textit{question}, including \textit{question without options}, \textit{true option}, \textit{false option}, \textit{objects} and \textit{question without objects}, with the same window size ($k=9$) around the $\ell$-th layer and observe the change in the probability of the answer word. In the tasks of \textit{CompareAttr}, \textit{LogicalObj} and \textit{QueryAttr}, we conduct the same operations with the above tasks except for blocking the attention edge between \textit{Related image patches} and \textit{question without objects} or \textit{objects} respectively. 

The observations of the overall information flow from \textit{related image patches} to different parts of the \textit{question} for all six tasks shown in \Cref{fig:relatedImagePatches2diffpartQues_LLaVA-v1.5-13b}
consistently align the flow observed from \textit{related image patches} to the entire \textit{question}, as illustrated in \Cref{fig:imagepatchesToquestion_13b} in the main body of the paper. 
Specifically, the information flow dominantly occurs in the second-time drop in the probability in the lower-to-middle layers (around 10th). However, there are some parts of \textit{question} that don't obtain the information followed from the \textit{related image patches}. For example, the \textit{objects} in the task of \textit{ChooseCat}, or  \textit{false option} and \textit{true option} in the task of \textit{ChooseRel}.

\section{The influence of \textit{\textbf{images}} on the semantics of \textit{\textbf{Questions}}}

\begin{figure}[t]
\centering
\includegraphics[width=0.5\linewidth]{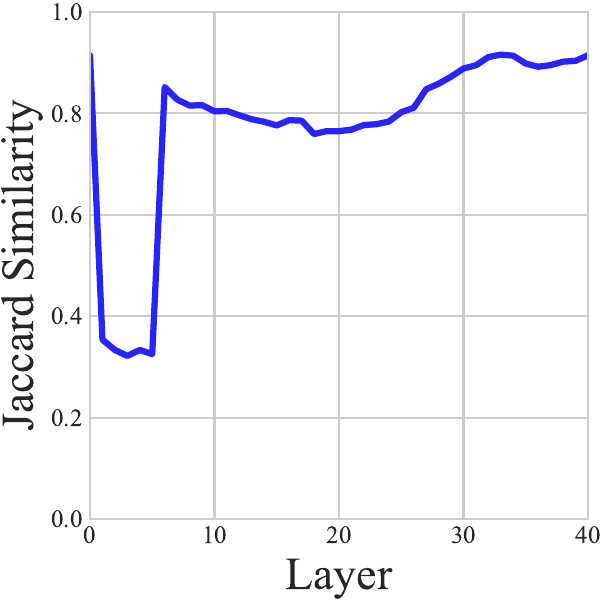}
\caption{The Jaccard similarity between the predicted words
of the original model \textit{LLaVA-1.5-13b} and those of the intervened model blocking \textit{question} from attending to \textit{image} on the task of \textit{ChooseAttr}.}
\label{fig:js_withoutImage_chooseattr}
\end{figure}

We already know that the \textit{image} information is integrated into the representation corresponding to the position of \textit{question}.
To investigate whether the \textit{image} affects the final semantics of the \textit{question}, for each layer $\ell$, we prevent the \textit{question} from attending to the \textit{question}, with the same window size ($k=9$) around the $\ell$-th layer and observe the change of semantics of the \textit{questtion} in the final layer. The semantics of the \textit{question} is evaluated by the \textit{Jaccard Similarity} as in \Cref{app: constructing multimodal semantic representations}.

As illustrated in \Cref{fig:js_withoutImage_chooseattr}, the \textit{Jaccard Similarity} demonstrates a significant decline in the lower layers, resembling the behaviour observed in layers where information flows from the \textit{image} to the \textit{question}. This pattern highlights the critical role of \textit{image} information in constructing the final multimodal semantic representation.

%% file: main.bbl
\begin{thebibliography}{47}
\providecommand{\natexlab}[1]{#1}
\providecommand{\url}[1]{\texttt{#1}}
\expandafter\ifx\csname urlstyle\endcsname\relax
  \providecommand{\doi}[1]{doi: #1}\else
  \providecommand{\doi}{doi: \begingroup \urlstyle{rm}\Url}\fi

\bibitem[les()]{lesswrongInterpretingGPT}
interpreting {G}{P}{T}: the logit lens — {L}ess{W}rong --- lesswrong.com.
\newblock \url{https://www.lesswrong.com/posts/AcKRB8wDpdaN6v6ru/interpreting-gpt-the-logit-lens}.
\newblock [Accessed 14-11-2024].

\bibitem[lmm(2024)]{lmms-lab/llama3-llava-next-8b-2024-11-13}
lmms-lab/llama3-llava-next-8b · hugging face.
\newblock \url{https://huggingface.co/lmms-lab/llama3-llava-next-8b}, 2024.
\newblock Accessed: 2024-11-13.

\bibitem[Aflalo et~al.(2022)Aflalo, Du, Tseng, Liu, Wu, Duan, and Lal]{aflalo2022vl}
Estelle Aflalo, Meng Du, Shao-Yen Tseng, Yongfei Liu, Chenfei Wu, Nan Duan, and Vasudev Lal.
\newblock Vl-interpret: An interactive visualization tool for interpreting vision-language transformers.
\newblock In \emph{Proceedings of the IEEE/CVF Conference on computer vision and pattern recognition}, pages 21406--21415, 2022.

\bibitem[Ainslie et~al.(2023)Ainslie, Lee-Thorp, de~Jong, Zemlyanskiy, Lebr{\'o}n, and Sanghai]{ainslie2023gqa}
Joshua Ainslie, James Lee-Thorp, Michiel de Jong, Yury Zemlyanskiy, Federico Lebr{\'o}n, and Sumit Sanghai.
\newblock Gqa: Training generalized multi-query transformer models from multi-head checkpoints.
\newblock \emph{arXiv preprint arXiv:2305.13245}, 2023.

\bibitem[Bai et~al.(2023)Bai, Bai, Yang, Wang, Tan, Wang, Lin, Zhou, and Zhou]{bai_qwen-vl_2023}
Jinze Bai, Shuai Bai, Shusheng Yang, Shijie Wang, Sinan Tan, Peng Wang, Junyang Lin, Chang Zhou, and Jingren Zhou.
\newblock Qwen-{VL}: {A} {Versatile} {Vision}-{Language} {Model} for {Understanding}, {Localization}, {Text} {Reading}, and {Beyond}, 2023.
\newblock arXiv:2308.12966 [cs].

\bibitem[Basu et~al.(2024)Basu, Grayson, Morrison, Nushi, Feizi, and Massiceti]{basu2024understanding}
Samyadeep Basu, Martin Grayson, Cecily Morrison, Besmira Nushi, Soheil Feizi, and Daniela Massiceti.
\newblock Understanding information storage and transfer in multi-modal large language models.
\newblock \emph{arXiv preprint arXiv:2406.04236}, 2024.

\bibitem[Cao et~al.(2020)Cao, Gan, Cheng, Yu, Chen, and Liu]{cao_behind_2020}
Jize Cao, Zhe Gan, Yu Cheng, Licheng Yu, Yen~Chun Chen, and Jingjing Liu.
\newblock Behind the {Scene}: {Revealing} the {Secrets} of {Pre}-trained {Vision}-and-{Language} {Models}.
\newblock \emph{Lecture Notes in Computer Science (including subseries Lecture Notes in Artificial Intelligence and Lecture Notes in Bioinformatics)}, 12351 LNCS:\penalty0 565--580, 2020.
\newblock arXiv: 2005.07310 ISBN: 9783030585389.

\bibitem[Chefer et~al.(2021)Chefer, Gur, and Wolf]{chefer_generic_2021}
Hila Chefer, Shir Gur, and Lior Wolf.
\newblock Generic {Attention}-model {Explainability} for {Interpreting} {Bi}-{Modal} and {Encoder}-{Decoder} {Transformers}.
\newblock pages 397--406, 2021.
\newblock arXiv: 2103.15679.

\bibitem[Dahlgren~Lindstr{\"o}m et~al.(2020)Dahlgren~Lindstr{\"o}m, Bj{\"o}rklund, Bensch, and Drewes]{dahlgren-lindstrom-etal-2020-probing}
Adam Dahlgren~Lindstr{\"o}m, Johanna Bj{\"o}rklund, Suna Bensch, and Frank Drewes.
\newblock Probing multimodal embeddings for linguistic properties: the visual-semantic case.
\newblock In \emph{Proceedings of the 28th International Conference on Computational Linguistics}, pages 730--744, Barcelona, Spain (Online), 2020. International Committee on Computational Linguistics.

\bibitem[Dai et~al.(2022)Dai, Dong, Hao, Sui, Chang, and Wei]{dai_knowledge_2022}
Damai Dai, Li Dong, Yaru Hao, Zhifang Sui, Baobao Chang, and Furu Wei.
\newblock Knowledge {Neurons} in {Pretrained} {Transformers}, 2022.
\newblock arXiv:2104.08696 [cs].

\bibitem[Dai et~al.(2023)Dai, Li, Li, Tiong, Zhao, Wang, Li, Fung, and Hoi]{dai_instructblip_2023}
Wenliang Dai, Junnan Li, Dongxu Li, Anthony Meng~Huat Tiong, Junqi Zhao, Weisheng Wang, Boyang Li, Pascale Fung, and Steven Hoi.
\newblock {InstructBLIP}: {Towards} {General}-purpose {Vision}-{Language} {Models} with {Instruction} {Tuning}, 2023.
\newblock arXiv:2305.06500 [cs].

\bibitem[Dar et~al.(2022)Dar, Geva, Gupta, and Berant]{dar2022analyzing}
Guy Dar, Mor Geva, Ankit Gupta, and Jonathan Berant.
\newblock Analyzing transformers in embedding space.
\newblock \emph{arXiv preprint arXiv:2209.02535}, 2022.

\bibitem[Dosovitskiy et~al.(2020)Dosovitskiy, Beyer, Kolesnikov, Weissenborn, Zhai, Unterthiner, Dehghani, Minderer, Heigold, Gelly, Uszkoreit, and Houlsby]{dosovitskiy_image_2020}
Alexey Dosovitskiy, Lucas Beyer, Alexander Kolesnikov, Dirk Weissenborn, Xiaohua Zhai, Thomas Unterthiner, Mostafa Dehghani, Matthias Minderer, Georg Heigold, Sylvain Gelly, Jakob Uszkoreit, and Neil Houlsby.
\newblock An {Image} is {Worth} 16x16 {Words}: {Transformers} for {Image} {Recognition} at {Scale}.
\newblock 2020.
\newblock arXiv: 2010.11929.

\bibitem[Dubey et~al.(2024)Dubey, Jauhri, Pandey, Kadian, Al-Dahle, Letman, Mathur, Schelten, Yang, Fan, et~al.]{dubey2024llama}
Abhimanyu Dubey, Abhinav Jauhri, Abhinav Pandey, Abhishek Kadian, Ahmad Al-Dahle, Aiesha Letman, Akhil Mathur, Alan Schelten, Amy Yang, Angela Fan, et~al.
\newblock The llama 3 herd of models.
\newblock \emph{arXiv preprint arXiv:2407.21783}, 2024.

\bibitem[Elhage et~al.(2021)Elhage, Nanda, Olsson, Henighan, Joseph, Mann, Askell, Bai, Chen, Conerly, DasSarma, Drain, Ganguli, Hatfield-Dodds, Hernandez, Jones, Kernion, Lovitt, Ndousse, Amodei, Brown, Clark, Kaplan, McCandlish, and Olah]{elhage2021mathematical}
Nelson Elhage, Neel Nanda, Catherine Olsson, Tom Henighan, Nicholas Joseph, Ben Mann, Amanda Askell, Yuntao Bai, Anna Chen, Tom Conerly, Nova DasSarma, Dawn Drain, Deep Ganguli, Zac Hatfield-Dodds, Danny Hernandez, Andy Jones, Jackson Kernion, Liane Lovitt, Kamal Ndousse, Dario Amodei, Tom Brown, Jack Clark, Jared Kaplan, Sam McCandlish, and Chris Olah.
\newblock A mathematical framework for transformer circuits.
\newblock \emph{Transformer Circuits Thread}, 2021.
\newblock https://transformer-circuits.pub/2021/framework/index.html.

\bibitem[Fang et~al.(2023)Fang, Wang, Xie, Sun, Wu, Wang, Huang, Wang, and Cao]{fang2023eva}
Yuxin Fang, Wen Wang, Binhui Xie, Quan Sun, Ledell Wu, Xinggang Wang, Tiejun Huang, Xinlong Wang, and Yue Cao.
\newblock Eva: Exploring the limits of masked visual representation learning at scale.
\newblock In \emph{Proceedings of the IEEE/CVF Conference on Computer Vision and Pattern Recognition}, pages 19358--19369, 2023.

\bibitem[Frank et~al.(2021)Frank, Bugliarello, and Elliott]{frank_vision-and-language_2021}
Stella Frank, Emanuele Bugliarello, and Desmond Elliott.
\newblock Vision-and-{Language} or {Vision}-for-{Language}? {On} {Cross}-{Modal} {Influence} in {Multimodal} {Transformers}.
\newblock pages 9847--9857, 2021.
\newblock arXiv: 2109.04448 ISBN: 9781955917094.

\bibitem[Geva et~al.(2021)Geva, Schuster, Berant, and Levy]{Geva_2021}
Mor Geva, Roei Schuster, Jonathan Berant, and Omer Levy.
\newblock Transformer feed-forward layers are key-value memories.
\newblock In \emph{Proceedings of the 2021 Conference on Empirical Methods in Natural Language Processing}. Association for Computational Linguistics, 2021.

\bibitem[Geva et~al.(2023)Geva, Bastings, Filippova, and Globerson]{geva_dissecting_2023}
Mor Geva, Jasmijn Bastings, Katja Filippova, and Amir Globerson.
\newblock Dissecting {Recall} of {Factual} {Associations} in {Auto}-{Regressive} {Language} {Models}, 2023.
\newblock arXiv:2304.14767 [cs].

\bibitem[Hendricks and Nematzadeh(2021)]{hendricks-nematzadeh-2021-probing}
Lisa~Anne Hendricks and Aida Nematzadeh.
\newblock Probing image-language transformers for verb understanding.
\newblock In \emph{Findings of the Association for Computational Linguistics: ACL-IJCNLP 2021}, pages 3635--3644, Online, 2021. Association for Computational Linguistics.

\bibitem[Hudson and Manning(2019)]{hudson2019gqa}
Drew~A Hudson and Christopher~D Manning.
\newblock Gqa: A new dataset for real-world visual reasoning and compositional question answering.
\newblock In \emph{Proceedings of the IEEE/CVF conference on computer vision and pattern recognition}, pages 6700--6709, 2019.

\bibitem[Krishna et~al.(2017)Krishna, Zhu, Groth, Johnson, Hata, Kravitz, Chen, Kalantidis, Li, Shamma, et~al.]{krishna2017visual}
Ranjay Krishna, Yuke Zhu, Oliver Groth, Justin Johnson, Kenji Hata, Joshua Kravitz, Stephanie Chen, Yannis Kalantidis, Li-Jia Li, David~A Shamma, et~al.
\newblock Visual genome: Connecting language and vision using crowdsourced dense image annotations.
\newblock \emph{International journal of computer vision}, 123:\penalty0 32--73, 2017.

\bibitem[Li et~al.()Li, Li, Xiong, and Hoi]{li_blip_nodate}
Junnan Li, Dongxu Li, Caiming Xiong, and Steven Hoi.
\newblock {BLIP}: {Bootstrapping} {Language}-{Image} {Pre}-training for {Unified} {Vision}-{Language} {Understanding} and {Generation}.
\newblock Technical report.

\bibitem[Li et~al.(2023)Li, Li, Savarese, and Hoi]{li_blip-2_2023}
Junnan Li, Dongxu Li, Silvio Savarese, and Steven Hoi.
\newblock {BLIP}-2: {Bootstrapping} {Language}-{Image} {Pre}-training with {Frozen} {Image} {Encoders} and {Large} {Language} {Models}.
\newblock 2023.
\newblock arXiv: 2301.12597.

\bibitem[Li et~al.(2024)Li, Zhang, Wang, Zhong, Chen, Chu, Liu, and Jia]{li2024minigeminiminingpotentialmultimodality}
Yanwei Li, Yuechen Zhang, Chengyao Wang, Zhisheng Zhong, Yixin Chen, Ruihang Chu, Shaoteng Liu, and Jiaya Jia.
\newblock Mini-gemini: Mining the potential of multi-modality vision language models, 2024.

\bibitem[Liu et~al.(2023)Liu, Li, Wu, and Lee]{liu_visual_2023}
Haotian Liu, Chunyuan Li, Qingyang Wu, and Yong~Jae Lee.
\newblock Visual {Instruction} {Tuning}, 2023.
\newblock arXiv:2304.08485 [cs].

\bibitem[Liu et~al.(2024{\natexlab{a}})Liu, Li, Li, and Lee]{Liu2023llava1_5}
Haotian Liu, Chunyuan Li, Yuheng Li, and Yong~Jae Lee.
\newblock Improved baselines with visual instruction tuning.
\newblock In \emph{Proceedings of the IEEE/CVF Conference on Computer Vision and Pattern Recognition (CVPR)}, pages 26296--26306, 2024{\natexlab{a}}.

\bibitem[Liu et~al.(2024{\natexlab{b}})Liu, Li, Li, Li, Zhang, Shen, and Lee]{liu2024llavanext}
Haotian Liu, Chunyuan Li, Yuheng Li, Bo Li, Yuanhan Zhang, Sheng Shen, and Yong~Jae Lee.
\newblock Llava-next: Improved reasoning, ocr, and world knowledge, 2024{\natexlab{b}}.

\bibitem[Lyu et~al.(2022)Lyu, Liang, Deng, Salakhutdinov, and Morency]{lyu2022dime}
Yiwei Lyu, Paul~Pu Liang, Zihao Deng, Ruslan Salakhutdinov, and Louis-Philippe Morency.
\newblock Dime: Fine-grained interpretations of multimodal models via disentangled local explanations.
\newblock In \emph{Proceedings of the 2022 AAAI/ACM Conference on AI, Ethics, and Society}, pages 455--467, 2022.

\bibitem[Meng et~al.(2022)Meng, Bau, Andonian, and Belinkov]{meng2022locating}
Kevin Meng, David Bau, Alex~J Andonian, and Yonatan Belinkov.
\newblock Locating and editing factual associations in {GPT}.
\newblock In \emph{Advances in Neural Information Processing Systems}, 2022.

\bibitem[Nanda et~al.(2023)Nanda, Chan, Lieberum, Smith, and Steinhardt]{nanda2023progress}
Neel Nanda, Lawrence Chan, Tom Lieberum, Jess Smith, and Jacob Steinhardt.
\newblock Progress measures for grokking via mechanistic interpretability.
\newblock \emph{arXiv preprint arXiv:2301.05217}, 2023.

\bibitem[Neo et~al.(2024)Neo, Ong, Torr, Geva, Krueger, and Barez]{neo2024towards}
Clement Neo, Luke Ong, Philip Torr, Mor Geva, David Krueger, and Fazl Barez.
\newblock Towards interpreting visual information processing in vision-language models.
\newblock \emph{arXiv preprint arXiv:2410.07149}, 2024.

\bibitem[Olah(2024)]{Mechanistic-2024-10-20}
Chris Olah.
\newblock Mechanistic interpretability, variables, and the importance of interpretable bases.
\newblock \url{https://www.transformer-circuits.pub/2022/mech-interp-essay}, 2024.
\newblock Accessed: 2024-10-20.

\bibitem[Palit et~al.()Palit, Pandey, Arora, and Liang]{palit_towards_nodate}
Vedant Palit, Rohan Pandey, Aryaman Arora, and Paul~Pu Liang.
\newblock Towards {Vision}-{Language} {Mechanistic} {Interpretability}: {A} {Causal} {Tracing} {Tool} for {BLIP}.

\bibitem[Radford et~al.(2021)Radford, Kim, Hallacy, Ramesh, Goh, Agarwal, Sastry, Askell, Mishkin, Clark, et~al.]{radford2021learning}
Alec Radford, Jong~Wook Kim, Chris Hallacy, Aditya Ramesh, Gabriel Goh, Sandhini Agarwal, Girish Sastry, Amanda Askell, Pamela Mishkin, Jack Clark, et~al.
\newblock Learning transferable visual models from natural language supervision.
\newblock In \emph{International conference on machine learning}, pages 8748--8763. PMLR, 2021.

\bibitem[Salin et~al.(2022)Salin, Farah, Ayache, Favre, Salin, Farah, Ayache, Trans, and Perspective]{salin_are_2022}
Emmanuelle Salin, Badreddine Farah, Stéphane Ayache, Benoit Favre, Emmanuelle Salin, Badreddine Farah, Stéphane Ayache, Benoit Favre Are Vision-language Trans, and Probing Perspective.
\newblock Are {Vision}-{Language} {Transformers} {Learning} {Multimodal} {Representations}? {A} probing perspective.
\newblock \emph{Proceedings of the 36th AAAI Conference on Artificial Intelligence}, 2022.

\bibitem[Schwettmann et~al.(2023)Schwettmann, Chowdhury, Klein, Bau, and Torralba]{schwettmann_multimodal_2023}
Sarah Schwettmann, Neil Chowdhury, Samuel Klein, David Bau, and Antonio Torralba.
\newblock Multimodal {Neurons} in {Pretrained} {Text}-{Only} {Transformers}.
\newblock In \emph{2023 {IEEE}/{CVF} {International} {Conference} on {Computer} {Vision} {Workshops} ({ICCVW})}, pages 2854--2859, Paris, France, 2023. IEEE.

\bibitem[Stan et~al.(2024)Stan, Rohekar, Gurwicz, Olson, Bhiwandiwalla, Aflalo, Wu, Duan, Tseng, and Lal]{stan2024lvlm}
Gabriela Ben~Melech Stan, Raanan~Yehezkel Rohekar, Yaniv Gurwicz, Matthew~Lyle Olson, Anahita Bhiwandiwalla, Estelle Aflalo, Chenfei Wu, Nan Duan, Shao-Yen Tseng, and Vasudev Lal.
\newblock Lvlm-intrepret: An interpretability tool for large vision-language models.
\newblock \emph{arXiv preprint arXiv:2404.03118}, 2024.

\bibitem[Touvron et~al.()Touvron, Lavril, Izacard, Martinet, Lachaux, Lacroix, Rozière, Goyal, Hambro, Azhar, Rodriguez, Joulin, Grave, and Lample]{touvron_llama_nodate}
Hugo Touvron, Thibaut Lavril, Gautier Izacard, Xavier Martinet, Marie-Anne Lachaux, Timothee Lacroix, Baptiste Rozière, Naman Goyal, Eric Hambro, Faisal Azhar, Aurelien Rodriguez, Armand Joulin, Edouard Grave, and Guillaume Lample.
\newblock {LLaMA}: {Open} and {Efficient} {Foundation} {Language} {Models}.

\bibitem[Touvron et~al.(2023)Touvron, Martin, Stone, Albert, Almahairi, Babaei, Bashlykov, Batra, Bhargava, Bhosale, Bikel, Blecher, Ferrer, Chen, Cucurull, Esiobu, Fernandes, Fu, Fu, Fuller, Gao, Goswami, Goyal, Hartshorn, Hosseini, Hou, Inan, Kardas, Kerkez, Khabsa, Kloumann, Korenev, Koura, Lachaux, Lavril, Lee, Liskovich, Lu, Mao, Martinet, Mihaylov, Mishra, Molybog, Nie, Poulton, Reizenstein, Rungta, Saladi, Schelten, Silva, Smith, Subramanian, Tan, Tang, Taylor, Williams, Kuan, Xu, Yan, Zarov, Zhang, Fan, Kambadur, Narang, Rodriguez, Stojnic, Edunov, and Scialom]{touvron_llama_2023}
Hugo Touvron, Louis Martin, Kevin Stone, Peter Albert, Amjad Almahairi, Yasmine Babaei, Nikolay Bashlykov, Soumya Batra, Prajjwal Bhargava, Shruti Bhosale, Dan Bikel, Lukas Blecher, Cristian~Canton Ferrer, Moya Chen, Guillem Cucurull, David Esiobu, Jude Fernandes, Jeremy Fu, Wenyin Fu, Brian Fuller, Cynthia Gao, Vedanuj Goswami, Naman Goyal, Anthony Hartshorn, Saghar Hosseini, Rui Hou, Hakan Inan, Marcin Kardas, Viktor Kerkez, Madian Khabsa, Isabel Kloumann, Artem Korenev, Punit~Singh Koura, Marie-Anne Lachaux, Thibaut Lavril, Jenya Lee, Diana Liskovich, Yinghai Lu, Yuning Mao, Xavier Martinet, Todor Mihaylov, Pushkar Mishra, Igor Molybog, Yixin Nie, Andrew Poulton, Jeremy Reizenstein, Rashi Rungta, Kalyan Saladi, Alan Schelten, Ruan Silva, Eric~Michael Smith, Ranjan Subramanian, Xiaoqing~Ellen Tan, Binh Tang, Ross Taylor, Adina Williams, Jian~Xiang Kuan, Puxin Xu, Zheng Yan, Iliyan Zarov, Yuchen Zhang, Angela Fan, Melanie Kambadur, Sharan Narang, Aurelien Rodriguez, Robert Stojnic, Sergey Edunov, and Thomas
  Scialom.
\newblock Llama 2: {Open} {Foundation} and {Fine}-{Tuned} {Chat} {Models}, 2023.
\newblock arXiv:2307.09288 [cs].

\bibitem[Vaswani et~al.(2017)Vaswani, Shazeer, Parmar, Uszkoreit, Jones, Gomez, Kaiser, and Polosukhin]{vaswani_attention_2017}
Ashish Vaswani, Noam Shazeer, Niki Parmar, Jakob Uszkoreit, Llion Jones, Aidan~N Gomez, Łukasz Kaiser, and Illia Polosukhin.
\newblock Attention is {All} you {Need}.
\newblock In \emph{Advances in {Neural} {Information} {Processing} {Systems}}. Curran Associates, Inc., 2017.

\bibitem[Wang et~al.(2023)Wang, Li, Dai, Chen, Zhou, Meng, Zhou, and Sun]{wang_label_2023}
Lean Wang, Lei Li, Damai Dai, Deli Chen, Hao Zhou, Fandong Meng, Jie Zhou, and Xu Sun.
\newblock Label {Words} are {Anchors}: {An} {Information} {Flow} {Perspective} for {Understanding} {In}-{Context} {Learning}, 2023.
\newblock arXiv:2305.14160 [cs].

\bibitem[Xu et~al.(2024)Xu, Pang, Zhu, Shen, and Cheng]{xu2024cross}
Shicheng Xu, Liang Pang, Yunchang Zhu, Huawei Shen, and Xueqi Cheng.
\newblock Cross-modal safety mechanism transfer in large vision-language models.
\newblock \emph{arXiv preprint arXiv:2410.12662}, 2024.

\bibitem[Zhang et~al.(2022)Zhang, Roller, Goyal, Artetxe, Chen, Chen, Dewan, Diab, Li, Lin, Mihaylov, Ott, Shleifer, Shuster, Simig, Koura, Sridhar, Wang, and Zettlemoyer]{zhang_opt_2022}
Susan Zhang, Stephen Roller, Naman Goyal, Mikel Artetxe, Moya Chen, Shuohui Chen, Christopher Dewan, Mona Diab, Xian Li, Xi~Victoria Lin, Todor Mihaylov, Myle Ott, Sam Shleifer, Kurt Shuster, Daniel Simig, Punit~Singh Koura, Anjali Sridhar, Tianlu Wang, and Luke Zettlemoyer.
\newblock {OPT}: {Open} {Pre}-trained {Transformer} {Language} {Models}, 2022.
\newblock arXiv:2205.01068 [cs].

\bibitem[Zhang et~al.(2024)Zhang, Shen, Yuan, Yan, Xie, Wang, Gu, Tang, and Ye]{zhang_redundancy_2024}
Xiaofeng Zhang, Chen Shen, Xiaosong Yuan, Shaotian Yan, Liang Xie, Wenxiao Wang, Chaochen Gu, Hao Tang, and Jieping Ye.
\newblock From {Redundancy} to {Relevance}: {Enhancing} {Explainability} in {Multimodal} {Large} {Language} {Models}, 2024.
\newblock arXiv:2406.06579 [cs].

\bibitem[Zhao et~al.(2024)Zhao, Xu, Gupta, Asthana, Zheng, and Gould]{zhao2024first}
Qinyu Zhao, Ming Xu, Kartik Gupta, Akshay Asthana, Liang Zheng, and Stephen Gould.
\newblock The first to know: How token distributions reveal hidden knowledge in large vision-language models?
\newblock \emph{arXiv preprint arXiv:2403.09037}, 2024.

\bibitem[Zheng et~al.(2023)Zheng, Chiang, Sheng, Zhuang, Wu, Zhuang, Lin, Li, Li, and OTHERS]{Zheng2023Vicuna}
Lianmin Zheng, Wei-Lin Chiang, Ying Sheng, Siyuan Zhuang, Zhanghao Wu, Yonghao Zhuang, Zi Lin, Zhuohan Li, Li, and OTHERS.
\newblock Judging llm-as-a-judge with mt-bench and chatbot arena.
\newblock 2023.

\end{thebibliography}
